\documentclass[lettersize,journal]{IEEEtran}
\usepackage{array}
\usepackage[caption=false,font=normalsize,labelfont=sf,textfont=sf]{subfig}
\usepackage{textcomp}
\usepackage{stfloats}
\usepackage{url}
\usepackage{verbatim}

\usepackage{graphics} 
\usepackage{epsfig} 
\usepackage{times} 
\usepackage{amsmath} 
\usepackage{amssymb}  
\usepackage{soul}
\usepackage{mathtools}
\usepackage{graphicx}
\usepackage{tabularx}
\usepackage{algorithm}
\usepackage{algorithmic}
\usepackage{color}
\usepackage{amsmath,amssymb,amsfonts}
\usepackage{graphicx}
\usepackage{textcomp}
\usepackage{xcolor}
\usepackage{algorithm}
\usepackage{verbatim} 
\usepackage{mathrsfs}
\usepackage{empheq}
\usepackage{lipsum}
\usepackage{physics}
\usepackage{comment}
\usepackage{mathtools, cuted}
\usepackage{bm}
\usepackage{amsthm}
\usepackage[utf8]{inputenc}
\usepackage[english]{babel}

\usepackage{hyperref}
\usepackage{soul}
\graphicspath{{./figures/}}
\usepackage{listings}
\makeatletter
\AtBeginDocument{%
  \let\c@figure\c@lstlisting
  
  \let\ftype@lstlisting\ftype@figure 
}
\makeatother

\lstdefinelanguage{PDDL}
{
  sensitive=false,    
  morecomment=[l]{;}, 
  alsoletter={:,-},   
  morekeywords={
    define,domain,problem,not,and,or,when,forall,exists,either,
    :domain,:requirements,:types,:objects,:constants,
    :predicates,:action,:parameters,:precondition,:effect,
    :fluents,:primary-effect,:side-effect,:init,:goal,
    :strips,:adl,:equality,:typing,:conditional-effects,
    :negative-preconditions,:disjunctive-preconditions,
    :existential-preconditions,:universal-preconditions,:quantified-preconditions,
    :functions,assign,increase,decrease,scale-up,scale-down,
    :metric,minimize,maximize,
    :durative-actions,:duration-inequalities,:continuous-effects,
    :durative-action,:duration,:condition,:cost,:param,:pre,:eff
  }
}
\newcommand\captionof[1]{\def\@captype{#1}\caption}
\usepackage{tcolorbox}
\newcommand{\PlanCode}[1]{{#1}}
\definecolor{lightblue}{RGB}{164, 194, 244}
\newcolumntype{L}{>{\centering\arraybackslash}m{3cm}}

\newcolumntype{P}[1]{>{\centering\arraybackslash}p{#1}}
\newcolumntype{M}[1]{>{\centering\arraybackslash}m{#1}}

\begin{document}
\bstctlcite{bstctl:nodash}
\title{A Survey of Optimization-based Task and Motion Planning: From Classical To Learning Approaches}

\author{Zhigen Zhao$^{1}$,
Shuo Cheng$^{2}$,
Yan Ding$^{3}$,
Ziyi Zhou$^{1}$,
Shiqi Zhang$^{3}$,
Danfei Xu$^{2}$,
and Ye Zhao$^{1,*}$
\thanks{$^{1}$The Laboratory for Intelligent Decision and Autonomous Robots (LIDAR), Georgia Institute of Technology, Atlanta, GA 30318, USA. {\tt\small \{zhigen.zhao, zhouziyi, yzhao301\}@gatech.edu}}
\thanks{$^{2}$Robot Learning and Reasoning Lab (RL2), Georgia Institute of Technology, Atlanta, GA 30318, USA. {\tt\small \{shuocheng, danfei\}@gatech.edu}}
\thanks{$^{3}$Autonomous Intelligent Robotics (AIR) Group, Binghamton University, Binghamton, NY 13902, USA. {\tt\small \{yding25, zhangs\}@binghamton.edu}}
\thanks{This work was partially funded by the Office of Naval Research (ONR) Award \char"0023 N000142312223, National Science Foundation (NSF) grants \char"0023 IIS-1924978, \char"0023 CMMI-2144309, \char"0023 FRR-2328254, and USDA \char"0023 2023-67021-41397.} 
\thanks{$^{*}$Corresponding Author.}
}
\IEEEaftertitletext{\vspace{-2\baselineskip}}
\markboth{IEEE/ASME Transactions on Mechatronics}%
{Shell \MakeLowercase{\textit{et al.}}: A Sample Article Using IEEEtran.cls for IEEE Journals}


\maketitle

\begin{abstract}
Task and Motion Planning (TAMP) integrates high-level task planning and low-level motion planning to equip robots with the autonomy to effectively reason over long-horizon, dynamic tasks. Optimization-based TAMP focuses on hybrid optimization approaches that define goal conditions via objective functions and are capable of handling open-ended goals, robotic dynamics, and physical interaction between the robot and the environment. Therefore, optimization-based TAMP is particularly suited to solve highly complex, contact-rich locomotion and manipulation problems.
This survey provides a comprehensive review on optimization-based TAMP, covering 
(i) planning domain representations, including action description languages and temporal logic, (ii) individual solution strategies for components of TAMP, including AI planning and trajectory optimization (TO), and (iii) the dynamic interplay between logic-based task planning and model-based TO. A particular focus of this survey is to highlight the algorithm structures to efficiently solve TAMP, especially hierarchical and distributed approaches. 
Additionally, the survey emphasizes the synergy between the classical methods and contemporary learning-based innovations such as large language models. 
Furthermore, the future research directions for TAMP is discussed in this survey, highlighting both algorithmic and application-specific challenges.

\end{abstract}
\begin{IEEEkeywords}
Task and Motion Planning, Trajectory Optimization, Robot Learning, Temporal Logic, AI Planning, Large Language Models
\end{IEEEkeywords}

\section{Introduction}
\begin{figure}\centering
\includegraphics[width=0.45\textwidth]{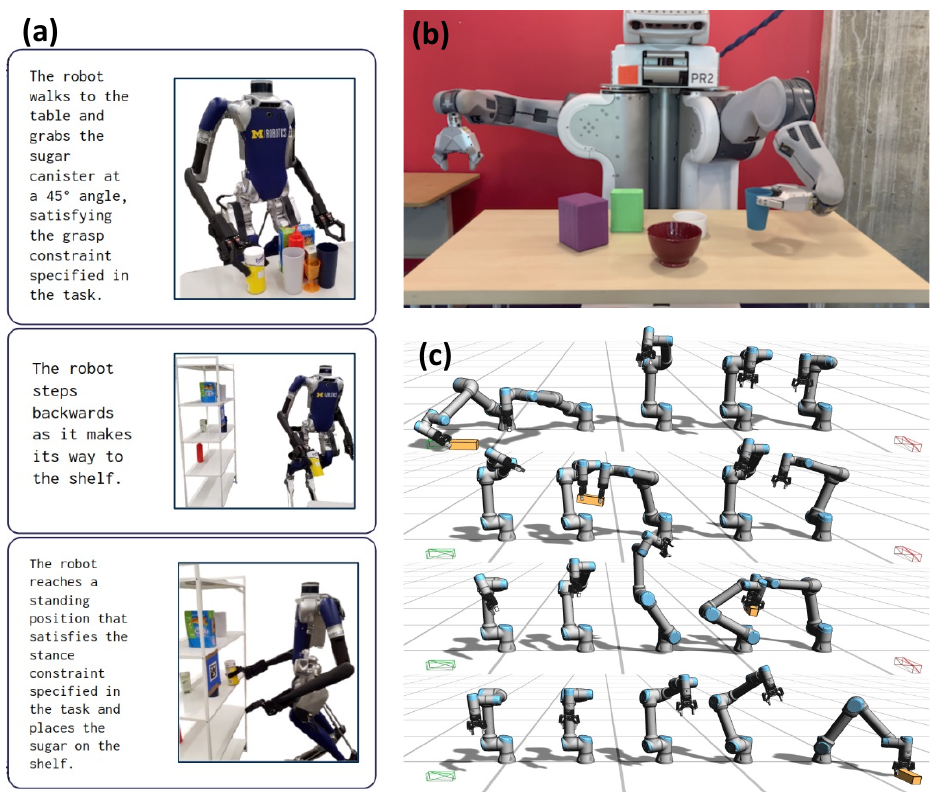}
\caption{Optimization-based TAMP enables dynamic locomotion and manipulation behaviors in complex environments: (a) bipedal robot loco-manipulation~\cite{adu2022optimal}; (b) mobile robot table-top manipulation~\cite{garrett2021integrated}; (c) long-horizon multi-agent collaboration~\cite{envall2023differentiable}.}
\label{fig:intro_figure}
\vspace{-0.25in}
\end{figure}

In recent years, robotic systems are rapidly transitioning from structured factory floors to unstructured human-centric environments. To this end, the demand continues to grow for a planning system that enables robots to efficiently perform complex, long-horizon tasks, as exemplified in Fig.~\ref{fig:intro_figure}. To achieve this level of autonomy, robots must be capable of generating and executing feasible and efficient motion plans that allow them to interact with their environment and complete assigned tasks. This complex problem is often framed as robot task and motion planning (TAMP), which breaks a complex, often intractable planning problem into a hybrid symbolic search and a set of local motion planning problems, where each sub-problem is tractable to solve.

The main research focus in TAMP is to develop appropriate problem representations and algorithms that efficiently synthesize both symbolic and continuous components of the planning problem~\cite{garrett2021integrated}. In the existing literature, there are three mainstream classes of TAMP methods: (i) constraint-based TAMP~\cite{dantam2018incremental, lozano2014constraint}, (ii) sampling-based TAMP~\cite{garrett2020pddlstream, krontiris2016efficiently}, and (iii) optimization-based TAMP~\cite{toussaint2015logic, takano2021continuous}. Constraint- and sampling-based TAMP characterizes the problem as a set of goal conditions. 
The solutions are typically found via constraint satisfaction or sampling-based approaches~\cite{guo2023recent}, which satisfy the defined goal conditions, but often cannot evaluate or compare the quality of the generated plan or the final state due to the lack of objective functions.
In many robotics problems, goals are often expressed as an objective function rather than an explicitly defined set of states.
For example, ``Given a number of rectangular blocks on the table, build a stable structure that is as tall as possible with minimal robot control effort.''
This is challenging for traditional sampling-based methods, which often require explicit goal definition and do not have mechanisms to compare plan qualities.
As an exception, a specific class of sampling-based motion planning~\cite{orthey2023sampling} have been proposed to address optimal planning using $\text{RRT}^*$ and $\text{PRM}^*$~\cite{karaman2011sampling, schmitt2017optimal}. 
{However, the complexity and expressiveness of the objective functions are often limited to simple costs such as path length, time, and energy consumption~\cite{gammell2021asymptotically}. A comparison between optimization and sampling-based TAMP methods is presented in Table~\ref{tab:compare_opt_sam}.}
For clarification, the scope of this survey focuses on optimization-based TAMP, which naturally defines an objective function for representing the plan quality, in addition to task- and motion-level constraints. This framework enables us to represent and solve a broad range of tasks with complex objective functions. 
\begin{figure*}\centering
\includegraphics[width=0.9\textwidth]{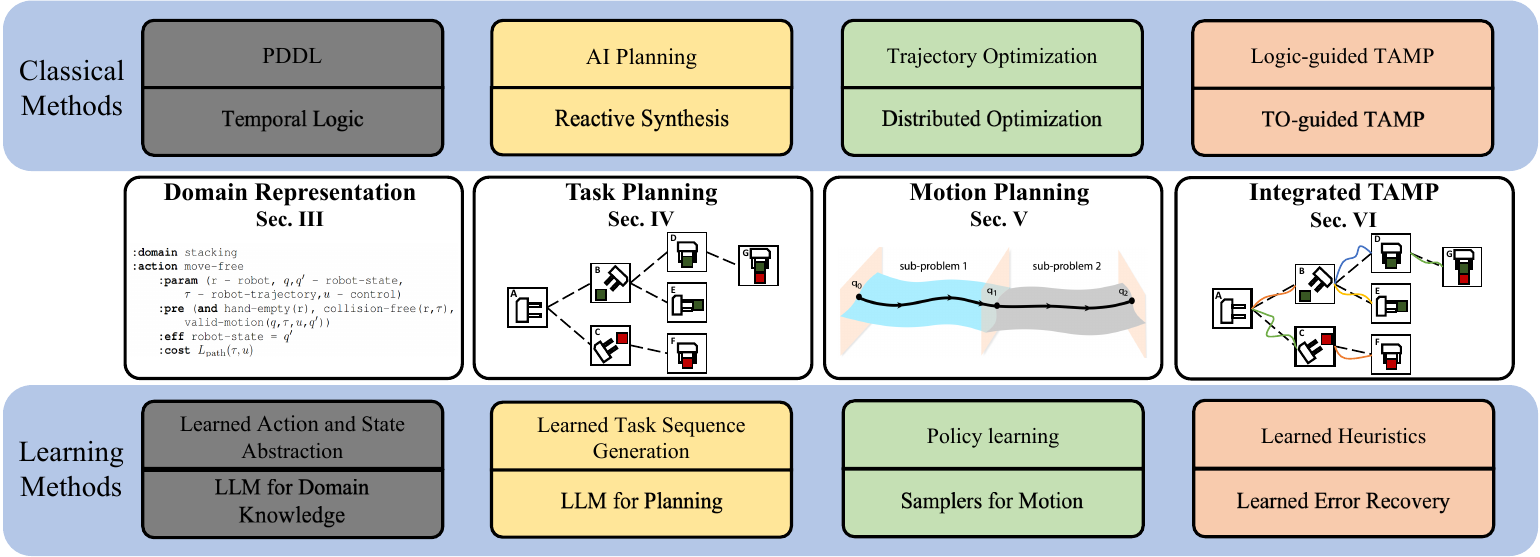}
\caption{Overview of the problem structures and related algorithms in optimization-based TAMP discussed in this survey paper.}
\label{fig:overview}
\vspace{-0.15in}
\end{figure*}

Optimization-based TAMP optimizes the objective function while adhering to constraints imposed by the robot kinematics and dynamics at the motion planning level and the discrete logic at the task planning level. This motivates the formulation of optimization-based TAMP as a hybrid optimization problem.
Optimization-based TAMP naturally incorporates model-based trajectory optimization (TO) methods in motion planning, which allow the planning framework to encode complex robot dynamics, leading to not only feasible but also natural,  efficient, and dynamic robot motions. This is especially important for contact-rich applications such as long-horizon robot manipulation~\cite{toussaint2018differentiable} of objects with complex geometry and frictional properties \cite{migimatsu2020object, stouraitis2020online} and dynamic locomotion over uneven terrains~\cite{aceituno2017simultaneous, sleiman2023versatile, zhao2022reactive, asselmeier2024hierarchical}. Additionally, optimization-based TAMP allows the inclusion of more complex objective functions and constraints (e.g., nonlinear and non-convex ones), enabling the robot to achieve various robot behaviors, thereby enhancing the applicability of robotic systems in real-world deployments. 

However, the hybrid optimization problems formed by optimization-based TAMP are often computationally intractable. A successful planning algorithm needs to simultaneously overcome the combinatorial complexity at the task planning level, and the numerical complexity at the motion planning level. As such, a common theme in optimization-based TAMP is to trade off between the complexity of the optimization and comprehensiveness of the information included in the planning problem. Either extreme of this trade-off tends to degrade either the quality or the computational efficiency of the resulting robot plans.
{Additionally, optimization-based TAMP faces several limitations comparing to sampling-based methods: (i) it is sensitive to the initial and goal conditions of the problem setup, which can lead to failures in complex environments such as complex obstacle geometry or difficult terrain, where certain initial and goal conditions can make it particularly challenging to find the optimal solution; (ii) the optimization results can be dependent on the initialization of decision variables, which might cause the planner to get stuck in local optima; (iii) optimization-based methods are not complete, meaning they cannot discover infeasible problems.} Therefore, the challenge remains to improve the robustness of optimization-based TAMP and bridge the gap between planning for long-horizon tasks~\cite{ghallab2004automated, meli2023logic} and generating highly dynamic robot behaviors, showcased in model-based optimal control strategies~\cite{wensing2023optimization, posa2014direct, tassa2014control}. 
\begin{table}[h!]
\centering
\begin{tabular}{M{0.25\linewidth}|M{0.3\linewidth}|M{0.3\linewidth}}
    \textbf{Aspect} & \textbf{Optimization-Based TAMP} & \textbf{Sampling-Based TAMP} \\
    \hline
     Optimality & Capable of converging to optimal or near-optimal solutions & Asymptotically optimal but no guarantee on convergence speed\\
     \hline
     Task Definition & Expressive definitions via objective functions \& constraints & Less expressive due to requirement for explicit goal definition, the objective function is limited to simple forms  \\ 
     \hline
     Robot Dynamics & Naturally incorporates dynamics in model-based TO & Costly to handle dynamics via steering or forward propagation \\
     \hline
     Constraint Handling & Handle complex constraints explicitly in model-based TO & Rely on relaxation or simplification to handle manifold constraints\\
     \hline
     Robustness & Sensitive to initialization and problem setup, potentially trapped by local optima & Robust to variations in initial conditions due to random sampling\\
     \hline
     Completeness & No mechanism to determine infeasibility & Typically probabilistically complete \\
     \hline
\end{tabular}
\vspace{0.025in}
\caption{Comparison Between Optimization-Based vs Sampling-Based TAMP Methods}
\label{tab:compare_opt_sam}
\end{table}
The integration of learning-based approaches in TAMP has become a significant research trend, as learning-based approaches offer considerable promise for enhancing the scalability and generalizability of classical TAMP methods. Leveraging learning as heuristics improves the efficiency of classical methods. 
For example, action feasibility checks during the task sequence search process can be accelerated by a neural feasibility classifier~\cite{xu2022accelerating,  yang2023sequence}. 
As an alternative method, generative models offer promising avenues to effectively replace certain components within the classical methods, as illustrated by learned task sequence generation from visual input~\cite{driess2020deep} and the use of large language models (LLMs) for domain knowledge representation and planning~\cite{lin2023text2motion, silver2024generalized, huang2022language}. 
Along another line of research, reinforcement learning (RL)-based skill learning has been studied in conjunction with the symbolic interface of a task planner, resulting in reusable skill learning that is generalizable across long-horizon tasks~\cite{10227514, mcdonald2022guided}.

\vspace{-0.1in}
\subsection{Survey Goals and Roadmap}
\begin{figure}
\centering
\includegraphics[width=0.5\textwidth]{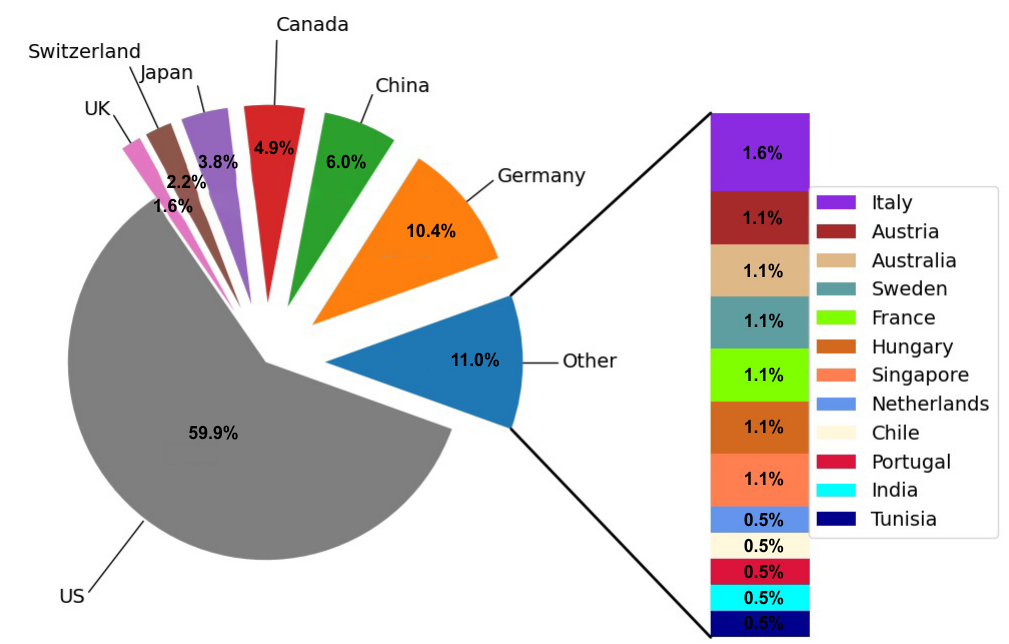}
\caption{{Percentage statistics sorted by countries of origin. 182 references that directly address TAMP are included.}}~\label{fig:countries}
\vspace{-0.15in}
\end{figure}

This work is inspired by and builds upon previous surveys in TAMP~\cite{garrett2021integrated, guo2023recent, mansouri2021combining, antonyshyn2023multiple}, but carries the unique overarching goal of reviewing the historical background and state-of-the-art optimization-based TAMP, and illustrating the connection between classical methods and the recent development in learning methods. Portions of this work are inspired by recent surveys in other relevant areas such as logic programming~\cite{meli2023logic}, formal methods~\cite{belta2019formal}, distributed optimization~\cite{shorinwa2023distributed}, and TO for legged locomotion~\cite{wensing2023optimization}. {Additionally, research contributions originated from 19 countries are highlighted to provide a global research landscape on TAMP innovations (Fig.~\ref{fig:countries}).}

We aim to provide discussion, promising solutions, and future trends in the following questions:
\begin{itemize}
    \item Q1: Why are optimization-based methods important for TAMP? What are the benefits?
    \item Q2: How will solutions for each individual component of optimization-based methods inform the strategies to solve integrated TAMP?  
    \item Q3: What common structures are observed in optimization-based TAMP problems, and which tools and strategies can exploit these structures to efficiently generate long-horizon, dynamic robot plans?
    \item Q4: How to leverage machine learning algorithms to enable robust, and generalizable TAMP frameworks?    
\end{itemize}

Q1 and Q2 motivate us to explore the key components and critical features of optimization-based TAMP, including problem formulation (Sec.~\ref{sec:preliminaries}), domain representation (Sec.~\ref{sec:domain_representation}), task planning (Sec.~\ref{sec:task_planning}), and motion planning (Sec.~\ref{sec:motion_planning}). 
Q3 seeks to present the current strategies and remaining challenges of optimization-based TAMP (Sec.~\ref{sec:tamp}), and inspire improved TAMP frameworks that efficiently manage complex task structures and robot dynamics. A particular focus of the discussion is on the interaction between task planning and motion planning layers.
Q4 addresses advancements in learning-based methods with the intent of synthesizing these elements for the enhancement of TAMP frameworks. The discussion on Q4 is interleaved with the classical methods to place the learning approaches into proper context. An overview of the structure of the survey is illustrated in Fig.~\ref{fig:overview}.

Q1-Q3 serve as an effective introduction for early-stage researchers new to the TAMP field, but also provide background information for experts in one or more of the TAMP components looking to explore an integrated optimization-based TAMP framework. 
Q4 provides important context for machine learning experts on how to combine learning techniques with classical TAMP. For the research groups currently exploring classical TAMP, this survey provides a systematic overview of the recent works in learning methods.

Finally, we offer our outlook on the potential future research directions in TAMP (Sec.~\ref{sec:challenges}), including the challenges in incorporating LLMs and skill learning in TAMP, as well as under-explored application areas such as loco-manipulation and human-robot collaboration.

\section{Preliminaries}
\label{sec:preliminaries}
In this section, we present the assumptions and definitions involved in optimization-based TAMP, as well as a motivating example that will be used throughout the survey. 
\vspace{-0.1in}
\subsection{Assumptions}
The following assumptions, adapted from~\cite{ghallab2004automated}, are made to formulate a basic definition for optimization-based TAMP:
\begin{itemize}
    \item \textbf{A1 Deterministic transitions:} If a symbolic action is applicable to a symbolic state, applying the action brings the deterministic symbolic transition system to a \textit{single} other symbolic state, similarly for the application of continuous control; 
    \item \textbf{A2 Known models and objectives:} The planner has complete knowledge about the continuous state transition system and the continuous dynamic system, as well as the objective function \textit{before} the planning process begins.
    \item \textbf{A3 Fully observable environments:} The planner has \textit{complete knowledge} about the symbolic and continuous states;
    \item \textbf{A4 Sequential plans:} the solutions to a TAMP problem are two \textit{linearly ordered} finite sequences of symbolic actions and continuous controls, respectively; 
\end{itemize}
In some extensions of the optimization-based TAMP problems, certain assumptions may not be satisfied. For example, in planning problem with probabilistic operators~\cite{silver2021learning}, the symbolic transitions are not deterministic, relaxing A1; in RL-based TAMP, a prior world model is unavailable, relaxing A2; in a TAMP framework incorporating visual input~\cite{driess2020deep}, the mapping function between observation and state is often learned implicitly or explicitly, relaxing A3. These variants present unique additional challenges due to the relaxation of certain assumptions. Nevertheless, the principles and patterns underscored throughout this survey retain their relevance and applicability, even in these more complex scenarios.

\vspace{-0.1in}
\subsection{Task and Motion Planning as Joint Optimization}\label{sec:preliminaries_tamp}
\vspace{-0.05in}
The optimization-based TAMP problem can be viewed as a joint optimization between task planning and motion planning. The optimization at two different levels are interconnected by constraints in both decision variables and cost functions.

The task planning domain is defined as $\mathcal{D}^t$, with a set of symbolic states $\mathcal{S}$, and a set of actions $\mathcal{A}$. Each symbolic state $s\in\mathcal{S}$ is defined by the values of a fixed set of discrete variables; each action $a\in\mathcal{A}$ specifies a state transition $s_{k+1}\in\gamma(s_{k}, a_k)$, where $k = 1, \cdots, K$ is the index of the discrete mode of the task planner. 
A task planning problem is represented by a task planning domain $\mathcal{D}^t$ an initial state $s^{\rm init}\in\mathcal{S}$, and a set of goal states $\mathcal{S}^{\rm goal}\subseteq\mathcal{S}$. A task plan consists of a symbolic state-action sequence of length $K$: $\langle \mathbf{S},\mathbf{A} \rangle = \langle s_0, a_0, s_1, a_1, \cdots, a_{K-1}, s_K \rangle$, where $s_0 = s^{\rm init}$, $s_K \in \mathcal{S}^{\rm goal}$.

The motion planning domain is defined as $\mathcal{D}^{m}$. The continuous robot state at the $t^{\rm th}$ knot point of the trajectory is represented by $\mathbf{x}_t=[\mathbf{q}_t, \dot{\mathbf{q}}_t]\in \mathbb{R}^{2n}$, where $\mathbf{q}_t, \dot{\mathbf{q}}_t\in\mathbb{R}^{n}$ represent the generalized configuration and velocity of the robot. The control input is $\mathbf{u}_t\in\mathbb{R}^m$. The discretized dynamics of the robot is denoted as $\mathbf{x}_{t+1}=f(\mathbf{x}_t, \mathbf{u}_t)$. The cost function at time $t$ is $L(\mathbf{x}_t, \mathbf{u}_t)\rightarrow\mathbb{R}$, which maps the state-control pair $\langle \mathbf{x}_t, \mathbf{u}_t \rangle$ to a real number. Additionally, $\langle \mathbf{x}_t, \mathbf{u}_t \rangle$ is constrained by various factors such as joint limits, torque limits, and robot collision. These constraints are denoted as $g(\mathbf{x}_t, \mathbf{u}_t) \leq \mathbf{0}$. 

The planning domain of the TAMP problem is jointly defined by the task planning domain $\mathcal{D}^t$ and the motion planning domain $\mathcal{D}^m$. Each symbolic state $s\in\mathcal{S}$ represents a manifold $\mathcal{X}^{s}$ in the continuous state space, which is specified by the state mapping function $M: \mathcal{X}^{s} = M(s)$. A symbolic state transition $\langle s_k, a, s_{k+1} \rangle$ corresponds to a continuous trajectory representing the robot motion: $\langle {\mathbf{X}_k},{\mathbf{U}_k} \rangle = \langle \mathbf{x}_{k,0}, \mathbf{u}_{k,0}, \mathbf{x}_{k,1}, \mathbf{u}_{k,1}, \cdots, \mathbf{u}_{k, T_k-1}, \mathbf{x}_{k, T_k} \rangle$. To achieve the symbolic state transition, the entire trajectory must lie within the manifold indexed by $s_k$: $\mathbf{x}_t\in\mathcal{X}^{s_k}\ , \forall\ t\in[0, T_k]$, while the final state of the $k^{\rm th}$ trajectory should lie on the intersection between the manifolds indexed by $s_k$ and $s_{k+1}$: $\mathbf{x}_{T_k}\in\mathcal{X}^{s_k}\cap\mathcal{X}^{s_{k+1}}$ and trigger the mode transition.

Given the planning domains $\langle \mathcal{D}^t, \mathcal{D}^m \rangle$, the initial states $\langle s^{\rm init}, \mathbf{x}_0^{\rm init} \rangle$ and goal states $\langle \mathcal{S}^{\rm goal}, \mathbf{x}_K^{\rm goal} \rangle$, the optimization-based TAMP problem is formulated as a joint optimization of the task-level decisions and the motion-level trajectory segments:
 \begin{subequations}\label{eq:joint_opt_tamp}
    \begin{align}\nonumber
      &\underset{\langle \mathbf{S},\mathbf{A}, \mathbf{X}_{1:k}, \mathbf{U}_{1:k} \rangle}{\min}  &&\sum_{k=0}^{K-1}\sum_{t=0}^{T_k-1} L_{\rm path}(\mathbf{x}_{k, t}, \mathbf{u}_{k,t}) + L_{\rm goal}(\mathbf{x}_{k,T_k})  \\
     & \quad\quad\;\; \text{s.t.}  && s_0 = s^{\rm init}, \ s_K\in \mathcal{S}^{\rm goal}, \;\\ 
     &&& \hspace{-0.1in} \forall k \in \{1,\cdots,K-1\}, \forall t \in \{0,\cdots,T_{k}-1\}\nonumber\\
      \label{eq:switch_constraint} &&&a_k \in \mathcal{A},  s_{k+1}=\gamma(s_k, a_k),\\
        &&& \mathbf{x}_{k,t+1}=f(\mathbf{x}_{k,t}, \mathbf{u}_{k,t}),\\
        &&&\mathbf{x}_{k,0} = \mathbf{x}_{k}^{\rm init}, \;\;\mathbf{x}_{k,T_k} = \mathbf{x}_{k}^{\rm goal}, \\
        &&&\mathbf{x}_{k, t} \in \mathcal{X}^{s_k}, \;\;\mathbf{x}_{k, T_k} \in \mathcal{X}^{s_k}\cap\mathcal{X}^{s_{k+1}},\\
        &&& g_k(\mathbf{x}_{k,t}, \mathbf{u}_{k,t}) \leq \mathbf{0},h_k(\mathbf{x}_{k,t}, \mathbf{u}_{k,t}) = \mathbf{0}.
    \end{align}
\end{subequations}

In this formulation, task planning and motion planning inform each other as they contain different subsets of the TAMP problem.
Symbolic states in TAMP correspond to manifold constraints in the continuous domain, while symbolic actions define transitions and constraints for motion planning. The sequence of actions, or the plan skeleton, guides the trajectory planning process by defining the sequence of mode transitions to be achieved. Conversely, motion planning informs task planning by providing geometric information, action feasibility, and cost evaluations, ensuring that task decisions are realizable at the motion level.

\subsection{Motivating Example}
\begin{figure}
\centering
\includegraphics[width=0.4\textwidth]{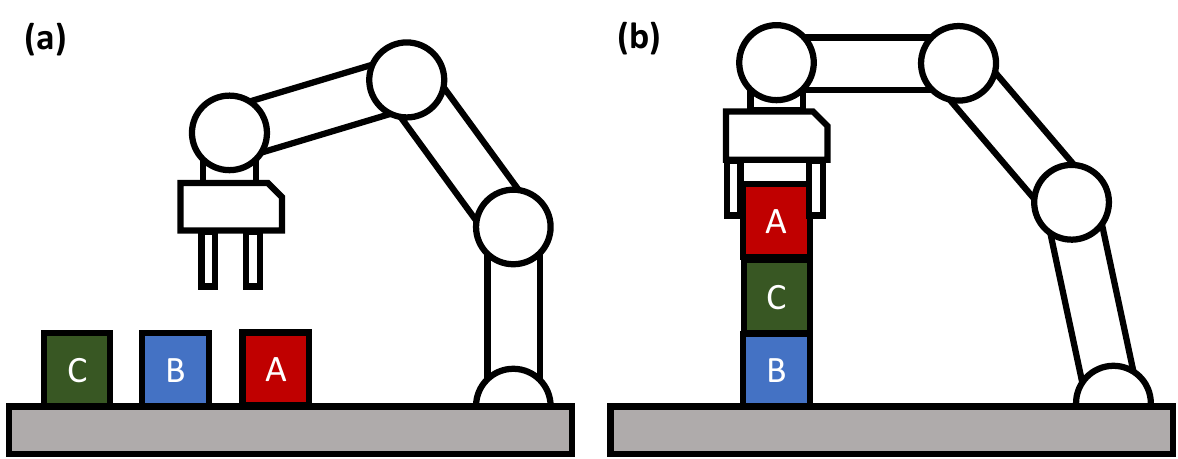}
\caption{An illustration for the table top manipulation example: (a) initial state (b) one possible final state.}~\label{fig:example}
\vspace{-0.15in}
\end{figure}

We introduce a tabletop manipulation task as a representative example to illustrate the formulations and algorithms discussed in this survey. As illustrated in Fig~\ref{fig:example}, the task involves a robot manipulator, denoted as $R$, and three distinct movable objects labeled as $A, B, C$. The primary objective of this task is to stack the objects such that the final height of object $A$ is maximized. Additionally, the robot should exert a minimal amount of control effort to achieve this task.

To quantitatively evaluate the performance of the manipulator in executing the task, we define an objective function. This function encompasses two distinct components: the path cost, which quantifies the control effort exerted by the robot throughout the task execution; the terminal cost, which measures the final elevation achieved by object $A$ at the conclusion of the task. The objective function is then constructed as a weighted sum of these individual costs, providing a holistic measure of the task's efficiency and effectiveness.

Throughout the survey, we will enrich the initial tabletop manipulation task with various extensions to demonstrate the practical considerations of the discussed algorithms and formulations. 
\section{Planning Domain Representation}\label{sec:domain_representation}
In real-world scenarios, planning domain representation 
demands formulating declarative knowledge about environments, robots, objects, their inter-relationships, and task goals, alongside integrating continuous-domain knowledge, such as robot configurations and object placement. 
Converting this knowledge into an optimization-based formulation requires a standardized interface, which ensures seamless integration by transforming varied input knowledge into an encoded form that can be effectively utilized by optimization algorithms. Therefore, this interface bridges the gap between real-world complexities and optimization-based TAMP.

Traditional methods for planning domain representation have been adopted from both the AI planning and temporal logic communities~\cite{meli2023logic}. These methods generally involve the use of logic. Sec.~\ref{sec:domain_representation_AI_planning} presents AI planning techniques that often employ domain-independent action description languages like the Planning Domain Definition Language (PDDL), which are widely interfaced with state-of-the-art task planners. Temporal logic approaches (Sec.~\ref{sec:domain_representation_temporal_logic}), utilizing formalisms such as linear temporal logic (LTL)~\cite{pnueli1977temporal, de2013linear}, signal temporal logic (STL)~\cite{maler2004monitoring}, and metric temporal logic (MTL)~\cite{koymans1990specifying}, have been extensively used to express time-dependent behaviors and constraints. 

One drawback of these traditional logic-based formalisms is that the domain representations are generally hand-specified by expert users. Therefore, a recent trend is to use learning-based methods to automatically encode domain knowledge for TAMP (Sec.~\ref{sec:domain_representation_learned_operators}-\ref{sec:domain_representation_llm}). These methods include the learning of symbolic operators, which can model the preconditions and effects of actions based on previous experiences. Furthermore, LLMs have been explored to process and interpret natural language inputs, providing a novel method for encoding the planning domain in a more intuitive and accessible format.

\subsection{AI Planning}\label{sec:domain_representation_AI_planning}
The task planning problem with discrete planning domains has long been the focus of the AI planning community.~\cite{ghallab2004automated} provides a comprehensive discussion of task planning representations and algorithms in the AI planning perspective. 

PDDL~\cite{aeronautiques1998pddl, haslum2019introduction} is a standard language extensively used in the AI planning community for encoding a task planning problem. It offers a compact and domain-independent syntax that aids in the clear delineation and representation of the task planning problem. An action $a \in \mathcal{A}(\mathcal{S})$ in PDDL is expressed as a tuple consisting of five components: $\langle\texttt{name}(a), \texttt{param}(a), \texttt{pre}(a), \texttt{eff}(a), \texttt{cost}(a)\rangle$.

\begin{itemize}
    \item \texttt{name}: name of the action.
    \item \texttt{param}: discrete and continuous parameters involved to evaluate $\texttt{pre}(a)$ and $\texttt{eff}(a)$.
    \item \texttt{pre}: a set of predicates that represent a set of facts that must be satisfied before the action can be applied.
    \item \texttt{eff}: a set of predicates that represent a set of facts that must be satisfied after the action is applied.
    \item \texttt{cost}: cost of the action represented by a positive scalar.
\end{itemize}

Classically, PDDL only supports a deterministic, discrete, and non-temporal world model~\cite{haslum2019introduction}. Historically speaking, multiple versions and extensions of PDDL have been developed to improve its expressiveness. Numerical expressions, plan metrics, and temporal planning are introduced in PDDL2.1~\cite{fox2003pddl2}. The latest official version is PDDL3.1~\cite{kovacs2011bnf}, which includes more elements of modern planning problems such as state-trajectory constraints, soft constraints, and object-fluents.
Among the PDDL extensions, hybrid system planning is handled in PDDL+~\cite{fox2006modelling}; probabilistic operators are introduced in PPDDL~\cite{younes2004ppddl1}; and multi-agent planning is included in MA-PDDL~\cite{kovacs2012multi}.

Within the context of TAMP, PDDL undergoes certain modifications to accommodate the inherent complexity of the domain. For robotics problems, additional continuous parameters are often introduced into the planning domain. The values of predicates in both the precondition, $\texttt{pre}(a)$, and the effect, $\texttt{eff}(a)$, can be functions of these continuous variables, such as robot poses or the continuous trajectory taken by the robot. Furthermore, the cost of an action may be defined as a function of the continuous trajectory.
This expanded use of PDDL allows for a more detailed and nuanced representation of planning problems, enabling the bridging of symbolic task planning and continuous motion planning.

\emph{Example: }The block stacking example can be represented by a hybrid AI planning \texttt{stacking} domain with five distinct actions, as seen in Fig.~\ref{lst:pddl_code}. The \texttt{move-holding} and \texttt{move-free} actions allows the manipulator to move along a collision free trajectory with or without holding an object. The \texttt{pick} action allows the robot to pick up an object that is on the table. The \texttt{place} action allows that robot to place the object it is currently holding onto the table. The \texttt{stack} action allows the robot to stack one object on top of another. Note that the actions are represented in a templated manner, which provides a compact representation of the planning problem. At the planning time, the actions are instantiated into grounded representations associated with specific robots and objects.

\begin{lstlisting}[language=PDDL,
                   basicstyle=\ttfamily\footnotesize,
                   frame=single,
                   breaklines=true,
                   escapeinside={(*}{*)},
                   caption={Hybrid block stacking problem expressed in a PDDL-style action description language.},
                   captionpos=b,
                   label={lst:pddl_code}]
:domain stacking
:action move-free
    :param (r - robot, (*$q$*),(*$q'$*) - robot-state,
        (*$\tau$*) - robot-trajectory,(*$u$*) - control)
    :pre (and hand-empty(r), collision-free(r,(*$\tau$*)), 
        valid-motion((*$q$*),(*$\tau$*),(*$u$*),(*$q'$*)))
    :eff robot-state = (*$q'$*)
    :cost (*$L_{\rm path}(\tau,u)$*)
:action move-hold
    :param (r - robot, o - object,
        (*$q$*),(*$q'$*) - robot-state, (*$\tau$*) - robot-trajectory, 
        (*$u$*) - control, (*$\phi$*) - object-trajectory)
    :pre (and holding(r,o), 
        collision-free(r,(*$\tau$*)), collision-free(o,(*$\phi$*)),
        valid-motion((*$q$*),(*$\tau$*),(*$u$*),(*$q'$*)), kinematics(r,o,(*$\tau$*),(*$\phi$*)))
    :eff robot-state = (*$q'$*)
    :cost (*$L_{\rm path}(\tau,u)$*)
:action pick
    :param (r - robot, o - object,
        (*$q$*) - robot-state, (*$p$*) - object-state)
    :pre (and hand-empty(r), clear(o), ontable(o), kinematics(r,o,(*$q$*),(*$p$*)))
    :eff (and (not hand-empty(r)), holding(r,o),
        (not ontable(o)))
:action place
    :param (r - robot, o - object, 
        (*$q$*) - robot-state, (*$p$*) - object-state)
    :pre (and holding(r,o), stable(o,(*$p$*)), 
        kinematics(r,o,(*$q$*),(*$p$*)))
    :eff (and hand-empty(r), (not holding(r, o)),
        ontable(o), clear(o))
:action stack
    :param (r - robot, x - object, y - object,
        (*$q$*) - robot-state, (*$p_x$*) - object-state, (*$p_y$*) - object-state)
    :pre (and holding(r,x), kinematics(r,x,(*$q$*),(*$p$*)), 
        stable-stack(x,y,(*$p_x$*),(*$p_y$*)))
    :eff (and hand-empty(r), (not holding(r,o)),
        on(x,y), (not clear(y)), clear(x))
\end{lstlisting}
\vspace{-0.15in}

\subsection{Temporal Logic}\label{sec:domain_representation_temporal_logic}
Temporal logic formalism provides concise expressions for temporal relations between symbolic expressions. 
One of the most popular classes of temporal logic in robotic applications is LTL~\cite{pnueli1977temporal, de2013linear}, which assumes a linear sequence of event, as opposed to the more complex nonlinear temporal logic (e.g. computation tree logic~\cite{emerson1982using}). The syntax of LTL contains a set of propositional variables $AP$, boolean operators $\neg$ (negation), $\land$ (conjunction), $\lor$  (disjunction), and a collection of temporal operators. The most common temporal operators are:
\begin{itemize}
    \item \texttt{eventually} $\Diamond \varphi$: $\varphi$ will hold true at some point in the future;
    \item \texttt{next} $\bigcirc \varphi$: $\varphi$ is true at the next time step;
    \item \texttt{always} $\Box \varphi$: $\varphi$ has to be true for the entire path;
    \item \texttt{until} $\varphi_1 \mathcal{U} \varphi_2$: $\varphi_1$ has to hold true at least until $\varphi_2$ becomes true;
    \item \texttt{release} $\varphi_1 \mathcal{R} \varphi_2$: $\varphi_2$ holds true until $\varphi_1$ becomes true.
\end{itemize}

One limitation of LTL formula is that only boolean variables and discrete time evaluation is allowed. Several extensions of LTL have been proposed to enable real-time and real-valued expressions. MTL~\cite{koymans1990specifying} extends LTL to real-time applications by allowing timing constraints. STL~\cite{maler2004monitoring}
further extends MTL to allow formula evaluation over continuous real-valued signals, which enrich the temporal logic formalism to specify hybrid planning problems in TAMP.

For STL, let \( \mathbf{y}: \mathbb{R}_{\geq 0} \to \mathbb{R}^n \) be a signal and \( t \in \mathbb{R}_{\geq 0} \) be a time. Let $(\mathbf{y}, t):=(\mathbf{y}, [t, \infty))$ denote the suffix of the signal. Let $\pi$ represent an atomic predicate of the form $\mu^\pi(\mathbf{y}) \geq 0$. The satisfaction of an STL formula \( \varphi \) at time \( t \) for signal \( \mathbf{y} \) is defined as:

\begin{itemize}
    \item \( (\mathbf{y}, t) \models \pi \) $\iff$ \( \mu^\pi(\mathbf{y}(t)) \geq 0 \);
    
    \item \( (\mathbf{y}, t) \models \neg \varphi \) $\iff$ \( (\mathbf{y}, t) \not\models \varphi \);
    
    \item \( (\mathbf{y}, t) \models \varphi_1 \land \varphi_2 \) $\iff$ \( (\mathbf{y}, t) \models \varphi_1 \) and \( (\mathbf{y}, t) \models \varphi_2 \);
    
    \item \( (\mathbf{y}, t) \models \varphi_1 \lor \varphi_2 \) $\iff$ \( (\mathbf{y}, t) \models \varphi_1 \) or \( (\mathbf{y}, t) \models \varphi_2 \);
    
    \item \( (\mathbf{y}, t) \models \Box_{[t_1,t_2]} \varphi \) $\iff$ \( \forall t' \in [t_1, t_2], (\mathbf{y}, t') \models \varphi \);
    
    \item \( (\mathbf{y}, t) \models \Diamond_{[t_1,t_2]} \varphi \) $\iff$ \( \exists t' \in [t_1, t_2], (\mathbf{y}, t') \models \varphi \);

    \item \( (\mathbf{y}, t) \models \varphi_1 \mathcal{U}_{[t_1,t_2]} \varphi_2 \) $\iff$ \( \exists t' \in [t_1, t_2], (\mathbf{y}, t') \models \varphi_2 \land \forall t'' \in [t_1, t'], (\mathbf{y}, t'') \models \varphi_1  \).
\end{itemize}

The robustness degree \( \rho(\mathbf{y}, \varphi, t) \) of STL is often used to quantify how well a given signal satisfies or violates an STL specification. The mathematical definition of robustness degree can be found in~\cite{donze2010robust}.

\emph{Encoding STL formula as mixed-integer constraints: }
The STL specification can be encoded into mixed-integer constraints using the big-M method~\cite{raman2014model}. The overall idea is that for each predicate $\pi$, a binary variable $z_t^\pi$ is created at time $t$, where 1 corresponds to true and 0 corresponds to false. Using the big-M method, the robustness degree $\rho$ can be represented with the inequality:
\begin{equation}
    \mu^\pi(\mathbf{y}(t)) - M_t(1-z_t) \geq \epsilon_t, \mu^\pi(\mathbf{y}(t)) - M_t z_t \leq \epsilon_t, 
\end{equation}
where $M_t$ is a sufficiently large constant for all predicates at time $t$, $M_t \geq \max_\pi \mu^\pi(\mathbf{y}(t))$, and $\epsilon_t$ is a sufficiently small positive constant that bounds $\mu^\pi(\mathbf{y}(t))$ away from 0. Using the big-M method, the boolean operations such as disjunction and conjunction are represented by the following:
\begin{align}
    &z = \bigwedge^{n_z}_{i=1}z_i \implies z \leq z_i, i = 1, \dots ,n_z, \\
    &z = \bigvee^{n_z}_{i=1}z_i \implies z \geq \sum_{i=1}^{n_z}z_i.
\end{align}
\cite{kurtz2022mixed} proposes a tree structure for STL formulas, resulting in a more efficient encoding that uses fewer binary variables. 
In comparison to the big-M method, the smoothed approximation approaches, introduced in Sec.~\ref{sec:tamp_to_guided}, represent the STL specifications as continuous constraints via the robustness degrees.

\emph{Example: }
To express the block stacking problem in STL, we first define the signals and predicates and then express the planning domain using STL formulas.
The continuous states and controls in the planning domain are represented as signals in STL. We define signals $\mathbf{p}_{A}(t), \mathbf{p}_{B}(t), \mathbf{p}_{C}(t)$ to be the 3D positions of objects $A, B, C$, and signal $\mathbf{y}(t) = [\mathbf{q}(t);\mathbf{u}(t)]$ to be the joint angles and torques of the robot manipulator at time t. Additionally, the gripper state is represented by signal $g(t)\in\{0, 1\}$, where $0$ means the gripper is closed and $1$ means it is opened. Note that, STL has to instantiate each object individually, which is different from the templated representation in PDDL. The following example in Fig.~\ref{fig:example_stl} provides one instantiation for each type of predicate.

\begin{figure}[H]
\begin{tcolorbox}[
    standard jigsaw,
    opacityback=0,
    colbacktitle=lightblue,
    coltitle=black,
]
The following predicates can be generated from the signals to express the state of the objects. 
\begin{itemize}
    \item $H_A, H_B, H_C$: whether the manipulator is holding an object A, B, or C, $\|\text{FK}(\mathbf{q}(t))-\mathbf{p}_A\|=0 \land g(t) \leq 0.5 \models H_A$, similarly for $H_B$ and $H_C$;
    \item $S_{AB}, S_{AC}, S_{BC}, S_{BA}, S_{CA}, S_{CB}$: whether an object is stacked on top of another. For example, $S_{AB}$ signifies A is stacked directly on top of B, i.e., $\big((\mathbf{p}_{A, x}(t)=\mathbf{p}_{B, x}(t)) \land (\mathbf{p}_{A, y}(t) = \mathbf{p}_{B, y}(t)) \land (\mathbf{p}_{A, z}(t) = \mathbf{p}_{B, z}(t) + h_A)\big) \models S_{AB}$, where $h_A$ is the the height of the object.
\end{itemize}

The STL formulation for the blocking domain is then represented in the following formulas:
\begin{itemize}
    \item Robot can hold only one object at a time: $\Box_{[0, T]}(H_A\Rightarrow\neg H_B \land \neg H_C)\land(H_B\Rightarrow\neg H_A \land \neg H_C)\land(H_C\Rightarrow\neg H_A \land \neg H_B)$;
    \item Objects cannot be free-floating: $\Box_{[0, T]}(\mathbf{p}_{A, z}(t)>0\Rightarrow H_A \lor S_{AB} \lor S_{AC})\land(\mathbf{p}_{B, z}(t)>0\Rightarrow H_B \lor S_{BA} \lor S_{BC})\land(\mathbf{p}_{C, z}(t)>0\Rightarrow H_C \lor S_{CA} \lor S_{CB})$
\end{itemize} 
\end{tcolorbox}
\vspace{-0.1in}
\captionof{figure}{Tabletop manipulation example expressed in STL.}
\label{fig:example_stl}
\vspace{-0.1in}
\end{figure}

\subsection{Learning Operators and State Abstractions}\label{sec:domain_representation_learned_operators}
To facilitate the search for task plan in solving TAMP problems, researchers propose learning symbolic operators, where probabilistic transition models are evaluated. Additionally, learning state abstractions studies the intrinsic structure of the task such as hierarchical structure and object importance in order to help decompose the large search space into two or more levels of abstractions.

\emph{Learning Operators:} Silver et al.~\cite{silver2021learning} propose to learn symbolic operators for TAMP using a relational learning method, where the demonstration data is first converted to symbolic transitions with defined predicates, and then the effects and preconditions are discovered by grouping transitions with similar effects. To alleviate the burden of hand-engineered symbolic predicates, Silver et al.~\cite{silver2023inventing} further propose to learn the symbolic predicates and the operators jointly from the demonstration data by optimizing a surrogate objective that relates to planning efficiency. To improve the generalization over novel objects, Chitnis et al.~\cite{chitnis2022learning} introduce Neuro-Symbolic Relational Transition Models, where high-level planning is achieved through symbolic search, and the learned action sampler and transition models are used to generate continuous motion. 

\emph{Learning State Abstractions:} State abstractions have also been studied to further improve the efficiency and generalization of TAMP systems. 
Chitnis et al.~\cite{chitnis2021camps} introduce a method for acquiring context-specific state abstractions. This approach focuses on considering only task-relevant objects, streamlining the planning process and improving adaptability across different scenarios, Silver et al.~\cite{silver2021planning} develop a GNN-based framework to predict object importance, thus allowing the planner to efficiently search for a solution while only considering the objects that are relevant to the task goal. Zhu et al.~\cite{zhu2021hierarchical} propose a hierarchical framework that constructs the symbolic scene graph and geometric scene graph from visual observations for representing the states, which are used for generating task plans and motion plans. Wang et al.~\cite{wang2022generalizable} suggest utilizing extensive datasets to enhance generalization. They adopt a two-step approach, commencing with the pre-training of visual features through symbolic prediction tasks and semantic reconstruction tasks. Subsequently, they employ the latent feature derived from this pre-training to learn abstract transition models, which in turn aid in guiding the task plan search process.

\subsection{Generating Domain Knowledge by LLMs}\label{sec:domain_representation_llm}
Generating domain knowledge for planning methods, including \emph{action descriptions} and \emph{goal specifications}, typically requires manual input from human experts using specific declarative languages like PDDL.
Manually encoding action description knowledge for task planners can be a tedious process. It requires extensive domain knowledge from human experts and must be regularly maintained to adapt to domain changes. 
It is a long-standing challenge of generating domain knowledge for autonomous agents (including robots) with minimum human involvement. 
Recent advances in LLMs have demonstrated the great potential of automating this process across various planning scenarios.

\emph{Generating Action Description by LLMs:} 
The strategy for generating action descriptions can be divided into two categories.
The first involves LLMs revising existing action descriptions to adapt them to different domains and situations. 
For instance, Ding et al. dynamically enrich original domain knowledge with task-oriented commonsense knowledge extracted from LLMs~\cite{ding2023integrating}.
The second category involves LLMs directly creating new action descriptions for planning. 
Here, researchers may employ various prompting methods to enhance generation performance.
Examples of such methods include specifying detailed prompts that guide the generative model towards producing outputs that are more aligned with the desired outcome~\cite{liu2023llm,silver2024generalized}, and integrating structured data through programming languages to provide a clear context or framework for the generation~\cite{singh2023progprompt}.
A major challenge in this area is ensuring the practicality of these generated descriptions in real planning systems, given the variability of LLM outputs. 
To address this, researchers deploy various evaluation methods, including simulations~\cite{zhao2024large}, comparison against predefined actions~\cite{huang2022language}, or human assessments, to filter the most viable outcomes~\cite{ren2023robots}.

\emph{Generating Goal Description with LLMs:}
Existing studies aim to translate objectives stated in natural language into specific formats such as PDDL~\cite{xie2023translating,liu2023llm} or LTL~\cite{pan2023data}.
This process is challenging, requiring understanding of context, adherence to syntax to avoid errors that may lead to failure, and alignment with the particular domains and problems.
Like action description generation, this requires various prompting methods, such as including examples in the prompts~\cite{liu2023llm,xie2023translating}.
Learning methods are also used to translate natural language directives into LTL specifications, with minimal need for human-annotated training data~\cite{pan2023data}.
Unlike others, AutoTAMP~\cite{chen2023autotamp} employs LLMs to translate task requirements into goals, which can be applicable at both task and motion levels. Fig.~\ref{fig:domainknowledge_llm} illustrates an example of translating natural language into goal specification.
\begin{figure}[H]
\begin{tcolorbox}[
    standard jigsaw,
    opacityback=0,
    colbacktitle=lightblue,
    coltitle=black,
]
\textbf{Problem Description:} \\
The problem description is: 
You have 3 blocks. 
B is on top of C. 
C is on top of A. 
A is on the table. 
B is clear. 
Your arm is empty. 
Your goal is to move the blocks, so that the final height of block A is as high as possible.
Provide me with the problem PDDL file that describes the planning problem directly without further explanations.\\
\vspace{-0.5em}
\\
\textbf{Problem File:}\\
\PlanCode{
(define (problem blocks-problem) \\
(:domain blocks)\\
  (:objects A B C - block)\\
  (:init
    (clear B)
    (on B C)
    (on C A)
    (on-table A)
    (arm-empty)
    (= (height A) 0)
    (= (height B) 0)
    (= (height C) 0)
  )\\
  (:metric (height A))
)
}
\end{tcolorbox}
\vspace{-0.1in}
\captionof{figure}{Tabletop manipulation example expressed in LLM prompt.}\label{fig:domainknowledge_llm}
\end{figure}

\section{Task Planning}\label{sec:task_planning}
Task planning focuses on determining sequences of actions to achieve specific goals using symbolic method. Traditionally, classical AI planning methods address this using graph search algorithms with specialized heuristics. Alternatively, temporal-logic-based techniques, especially those using LTL, employ automata theory and reactive synthesis to generate discrete decision sequences.
However, the aforementioned classical AI planning and temporal-logic-based approaches are not without limitations. One major challenge is the combinatorial complexity that arises when dealing with large-scale planning problems. This complexity can severely hamper the scalability and efficiency of planning algorithms.

To address these challenges, recent advancements in the field aim to bypass the combinatorial bottleneck by leveraging learned models to guide the task sequence search. These approaches utilize insights from learned models, incorporating task decompositions, action affordances, and the effects of skills. Notably, the advent of LLMs has introduced new methodologies. LLM-native planning derives strategies directly from data, while LLM-aided techniques synergize these models with established planning systems. The fusion of classic algorithms with state-of-the-art machine learning encourages a promising evolution in task planning algorithms.

\subsection{Classical Task Planning} \label{sec:task_planning_state}
Classical task planning, as described by Ghallab et al.~\cite{ghallab2004automated}, refers to the problem of planning for a deterministic, static, finite, and fully observable state-transition system with restricted goals and implicit time. 
The most straightforward task planning algorithms are state-space search methods. In this paradigm, the search space is a subset of the state-space itself, where each node in the search represents a state, and each edge symbolizes a transition. The state-space search typically results in a sequential path traversing the state space, effectively detailing the progression from an initial state to a goal state. State-space search is particularly relevant to the field of TAMP, as the underlying motion planning algorithm inherently operates on state space. 
The key considerations for algorithm design include the identification of appropriate search space, the selection of efficient algorithms, and the determination of suitable heuristics to guide the search process. 

The search heuristics in classical AI planning can be seen as the relaxation of the exact search problem. In practice, the heuristics design often involves a trade-off between computational cost and informativeness of the heuristics.
The works in~\cite{hoffmann2001ff, baier2009heuristic} employ heuristics based on the idea of state reachability relaxation, where the heuristics are computed by constructing a relaxed planning graph starting at state $s$, and all negative effects of operators are ignored when growing the graph. Therefore, the resulting planning graph has the properties of monotonic increase in the number of propositions with respect to the depth of the graph. A simple, computationally-cheap heuristics based on the relaxed planning graph is the goal distance function~\cite{zhu2005simultaneous}. Let the distance to goal $h^*(s)$ be defined as the minimum number of operators needed to reach the goal. The lower bound estimation of $h^*(s)$ can be easily calculated by the minimum depth of the node containing all the goal propositions within the relaxed planning graph. As an alternative approach, the Fast Downward-based~\cite{helmert2006fast, richter2011lama} planning systems uses hierarchical decomposition of planning tasks to compute a causal graph heuristic, which uses the causal dependencies in a relaxed causal graph to guide the forward state-space search.

In comparison to state-space search, other AI planning methods, such as hierarchical task network~\cite{georgievski2014overview}, attempt to conduct search on plan-space. However, these methods are not often used in TAMP scenarios due to the difficulty in interfacing plan-space search with motion planners.

For temporal logic based formulations such as LTL, automata based approaches~\cite{hopcroft2001introduction} such as reactive synthesis~\cite{pnueli1989synthesis, maoz2015gr, ehlers2016slugs} are often used to generate a reactive system that ensures the system meets a desired specification irrespective of external inputs.

Note that this survey assumes the readers have basic backgrounds of classical task planning and intentionally keeps this section brief. For more information, readers are referred to the references~\cite{ghallab2004automated, belta2019formal, meli2023logic}.

\subsection{Learning Models for Task Planning}\label{sec:learning_symbolic}
A key challenge of improving the scalability of TAMP is the combinatorial complexity of the discrete planning problem and the large number of motion planning problems to be solved. A promising approach to circumvent this challenge is to use learning methods to guide the high-level task plan search. Pasula et al.~\cite{pasula2007learning} propose to learn probabilistic, relational planning rule representations to model the action effects, which can be used to generate the task plan through search. Similarly, Amir et al.~\cite{amir2008learning} develop a method that learns the deterministic action models in partially observable domains. To allow dealing with uncertain representation and probabilistic plans, Konidaris et al.~\cite{konidaris2015symbol} propose to replace the sets and logical operations by probability distributions and probabilistic operations, and develop a framework that enables autonomous learning of the probabilistic symbols from continuous environments. To address the challenge of goal-directed planning involving a set of predefined motor skills, Konidaris et al.~\cite{konidaris2018skills} present a framework that directly acquires symbolic representations, abstracting the low-level transitions for effective utilization in planning tasks.

More recently, deep learning techniques have been explored to learn the models from large-scale datasets. Ames et al.~\cite{ames2018learning} propose to learn preconditions, action parameters, and effects from execution results of parameterized motor skills, which are then used to construct symbolic models for efficient planning. Neural task programming~\cite{xu2018neural} proposes to learn neural models that recursively decompose a task demonstration video into robot executable action primitives. To further improve the generalization on long-horizon tasks, Neural task graphs~\cite{huang2019neural} learns neural networks for generating conjugate task graphs, where the actions are represented as nodes and the dependencies between actions are modeled by edges, better exploring the compositionality. Regression planning networks~\cite{xu2019regression} learns a neural model to iteratively predict the intermediate subgoals in a reverse order based on the current image observation and the final symbolic goal. Ceola et al.~\cite{ceola2019robot} propose to utilize deep reinforcement learning to train neural models for generating discrete actions. Deep affordance foresight~\cite{xu2021deep} learns the long-term affordance of actions and the latent transition models to guide the search, and thereby informs the robot of the best actions to achieve the final task goal. Similarly, Liang et al.~\cite{liang2022search} propose to learn skill effect models that generate future terminal states of each parameterized skill, and then leverage these models to aid search-based task planning.

\subsection{LLMs for Task Planning}
\begin{figure}
\includegraphics[width=0.5\textwidth]{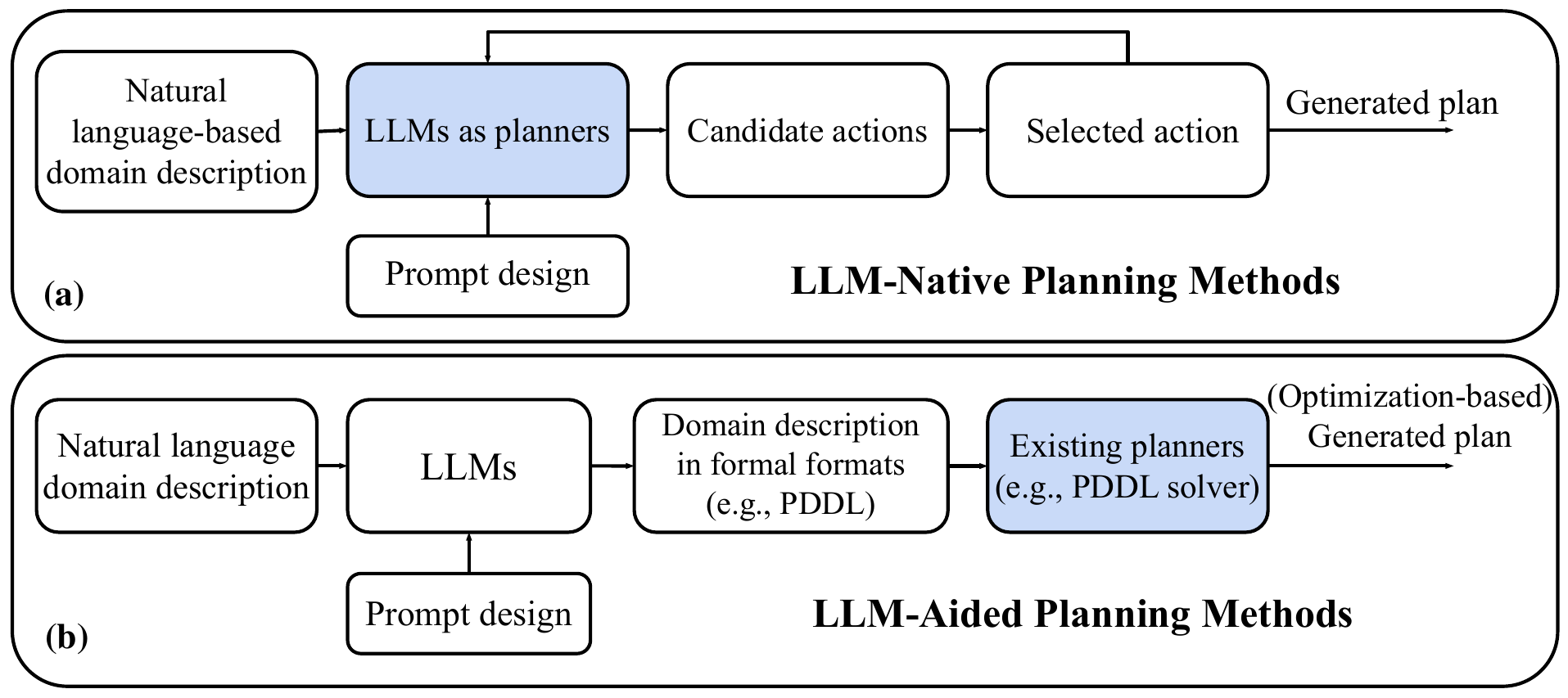}
\caption{Two methodologies for task planning using LLMs, with the key difference lies in the role of LLMs: (a) LLM-Native planning methods use LLMs for planning, while (b) LLM-Aided planning methods LLMs to generate domain descriptions for existing task planning methods such as PDDL and temporal logic.}~\label{fig:llm_for_taskplanning}
\vspace{-0.15in}
\end{figure}

Traditionally, optimizing task plans for robots involves minimizing either the number of actions or the total plan cost, depending on whether action costs are considered. 
The emergence of LLMs, such as Google's Bard, OpenAI's ChatGPT~\cite{openai}, and Meta's LLaMA~\cite{touvron2023llama}, have reshaped the landscape of AI, including task planning for robots~\cite{liu2023pre}. 
We categorize the LLM-based planning methods into the two groups: \emph{LLM-Native Planning Methods} and \emph{LLM-Aided Planning Methods}, where the former does not rely on external knowledge and the latter does, as shown in Fig.~\ref{fig:llm_for_taskplanning}. 
{Comparing to regular learning-based methods, LLMs are typically trained on a large amount of out-of-domain data that contains a great deal of commonsense knowledge. While LLMs are not strong in numerical reasoning (and hence optimization)~\cite{imani2023mathprompter,gaur2023reasoning}, the incorporation of LLMs improves the capabilities of natural language understanding, the acquisition of world knowledge, and commonsense reasoning. Such capabilities enable LLM-based planners to reason about symbolic information such as spatial relationships between objects~\cite{ding2023task} and symbolic correctness of a task sequence~\cite{lin2023text2motion}, without prior interaction with the robot environment. Therefore, LLMs as a task and domain agnostic reasoning module has the potential to enhance the scalability and generalizability of robot planning.}

LLM-Native planning methods often incorporate additional components like reinforcement learning to enhance planning by choosing better actions. 
Conversely, LLM-Aided planning methods can be integrated with classical optimization strategies, ensuring satisfactory planning efficiency and practicality. 
These two approaches are compatible with  optimization methods, while integrating LLMs enhances the overall planning capabilities.

\emph{LLM-Native Planning Methods:}
One method to leverage LLMs for task planning involves directly generating plans from LLMs by providing a domain description (Fig.~\ref{fig:llm_for_taskplanning}(a)). 
This can be done either in a one-shot way or iteratively. 
These methods primarily focus on prompt design for effective communication with LLMs, and the grounding to specific domains and robot skills. 
Multiple systems have made efforts in this field.
Huang et al.~\cite{huang2022language} propose to generate candidate actions and design tools to improve their executability, such as enumerating all permissible actions and mapping the model's output to the most semantically similar action.
Building upon this, SayCan~\cite{brohan2023can} enables robotic planning using affordance functions that determine action feasibility and respond to natural language requests such as ``deliver a Coke''. 
An advanced approach, named Inner Monologue, developed by Huang et al.~\cite{huang2023inner}, integrates environmental feedback for task planning and situation handling. 
Previously, methods typically generate task plans in text forms.
Singh et al.~\cite{singh2023progprompt} develop a system, called ProgPrompt, which employs programmatic LLM prompts to generate task plans and manage situations, by verifying the preconditions of the plan and reacting to failed assertions with suitable recovery actions.
Employing code as the framework for high-level planning provides significant benefits. 
It allows for the expression of complex functions and feedback loops. 
These loops effectively process sensory outputs and enable the parameterization of control primitives within APIs~\cite{hu2023toward}.
Apart from planning for robots, there is also research on whether LLMs can act as universal planners.
They could potentially create programs that efficiently generate plans for various tasks within the same domain~\cite{silver2024generalized}

\emph{LLM-Aided Planning Methods:} 
Prior to the development of LLMs, various tools existed for robot task planning, but they had scalability limitations. 
For example, defining domain knowledge in PDDL demands significant time from human experts (Fig.~\ref{fig:llm_for_taskplanning}(b)). 
The advent of LLMs offers a way to augment these traditional planners by supplementing knowledge, thereby improving their performance and enabling more natural language interaction.
There are a few ways of integrating LLMs and classical task planners. 
First, a series of studies are conducted to explore the conversion of natural language descriptions of planning tasks into standardized languages like PDDL or temporal logic.
LLMs complete these transformations, playing a crucial role in the process.
These translated specifications are then used in existing planning systems.
For example, Xie et al.~\cite{xie2023translating} create optimality-based task-level plans with the PDDL planner, translating natural language inputs into PDDL problems. 
Second, one can dynamically extract commonsense knowledge from LLMs, enhancing PDDL's action knowledge for planning and situational handling~\cite{ding2023task}. 
Third, Zhao et al.~\cite{zhao2024large} utilize LLMs to build world models, and perform heuristic policy in search algorithms such as Monte Carlo Tree Search, which uses the common-sense knowledge provided by LLMs to generate possible world states, thus facilitating efficient decision-making as well as underlying motion planning.

The optimization of those LLM-based planning methods occurs in the interaction with the LLMs, in the plan generation of classical task planners, or both. 
The prompting strategy of LLM-Native planning methods encourage behaviors towards maximizing the overall task completion rate, where the optimization usually occurs in an implicit way (i.e., there is no objective function explicitly specified). 
By comparison, the LLM-Aided planning methods compute optimal plans with or without plan cost in consideration, where the optimality is conditioned on the external knowledge provided by LLMs, and the optimization process is embedded within the deployed task planning system. 

\section{Optimization-based Motion Planning}\label{sec:motion_planning}
Optimization-based motion planning is an important component in robot planning. It aims to generate a continuous robot motion path and a control sequence that optimizes an objective function subject to a set of kinematics and/or dynamics constraints. Numerous methods have been proposed \cite{betts1998survey} to TO\footnote{In this survey, we interchangeably use the terms of ``trajectory optimization" and ``optimization-based motion planning".}. Notable TO techniques include direct methods that transcribe TO into nonlinear programs (NLPs), and indirect methods that leverage the optimality conditions.

In the meantime, with the increasing complexity and diversity of environments that robots operate in and tasks that the robot are required to accomplish, there is an imperative need to enhance the scalability of these TO strategies, especially in handling robot dynamics, complex constraints in physical contact problems, and higher dimensional state spaces in multi-robot scenarios. To this end, distributed optimization techniques have been introduced, with consensus Alternating Direction Method of Multipliers (ADMM) being a notable methodology~\cite{boyd2011distributed}.

In conjunction with model-based TO approaches, recent advancement in combining data-driven approaches and TO has shown capabilities in predictively generating trajectories by imitating offline-generated optimized paths solved by model-based TO techniques~\cite{mordatch2014combining, janner2021offline}. These learned methods hold significant promise in enhancing the efficiency and adaptability of motion planning processes, especially in environments with dynamic and unforeseen challenges.

\subsection{Trajectory Optimization}
A motion planning problem is specified by a motion planning domain $\mathcal{D}^{m}$, an initial state $\mathbf{x}^{\rm init}\in\mathbb{R}^{2n}$, and a goal state $\mathbf{x}^{\rm G}\in\mathbb{R}^{2n}$. A motion plan consists of a state-control trajectory with $T$ knot points: $\langle {\mathbf{X}},{\mathbf{U}} \rangle = \langle \mathbf{x}_0, \mathbf{u}_0, \mathbf{x}_1, \mathbf{u}_1, \cdots, \mathbf{u}_{T-1}, \mathbf{x}_T \rangle$, where $\mathbf{x}_0 = \mathbf{x}^{\rm init}$ and $\mathbf{x}_T = \mathbf{x}^{\rm G}$.

The motion planning problem can be formulated as a constrained NLP:
\begin{subequations}
\begin{align} \label{eq:opt_motion_planning}
    \min_{{\mathbf{X}},{\mathbf{U}}}\ &\sum_{t=0}^{T-1} L_{\rm path}(\mathbf{x}_t, \mathbf{u}_t) + L_{\rm goal}(\mathbf{x}_T)\\
    \text{s.t.} \quad &\mathbf{x}_{t+1}=f(\mathbf{x}_t, \mathbf{u}_t),\label{eq:trajopt_dyn}\\
    &\mathbf{x}_0 = \mathbf{x}^{\rm init}, \;\mathbf{x}_T = \mathbf{x}^{\rm G}, \label{eq:trajopt_boundary_cond}\\
    &g(\mathbf{x}_t, \mathbf{u}_t) \leq \mathbf{0}.\label{eq:trajopt_ineq}
\end{align}
\end{subequations}
where the dynamics equation in Eq.~\eqref{eq:trajopt_dyn} and inequality constraint in Eq.~\eqref{eq:trajopt_ineq} are defined in Sec.~\ref{sec:preliminaries_tamp}.

Direct collocation \cite{kelly2017introduction, pardo2016evaluating} offers a straightforward transcription where both controls and states are treated as decision variables, and complex state constraints can be easily expressed. General-purpose NLP solvers such as IPOPT \cite{wachter2006implementation} and SNOPT \cite{gill2005snopt} can be adopted to solve for optimal solutions. Alternatively, motivated by the real-time computation requirement for many robotics applications, researchers start to devise problem-specific solvers for reliably solving the above NLP. Notably, Differential Dynamic Programming (DDP) \cite{mayne1966second}
is a shooting method that efficiently explores the problem structure through Riccati recursion and handles nonlinear dynamics, but limited to unconstrained TO.
More recently, variants of DDP algorithms have been proposed to handle diverse state and control constraints \cite{tassa2014control, xie2017differential, plancher2017constrained, howell2019altro, sleiman2021constraint, 
jallet2022constrained, wang2023fast}. Readers are referred to  \cite{betts1998survey, nocedal1999numerical} for a comprehensive overview on the numerical TO methods.The recent survey paper~\cite{wensing2023optimization} offers insights into contemporary applications of TO in legged locomotion with an emphasis on handling complex dynamic and contact constraints.

\emph{Example: }
For the tabletop manipulation task, we consider the motion planning problem for a single task of a manipulator moving from a free position to pick up an object $A$. Let forward kinematics function $FK(\cdot)$ denote the end-effector position of the manipulator and $\mathbf{p}_A$ denote the position of object $A$. The running cost consists of a position tracking term and regularization terms for state and control:
\begin{align}
    L_{\text{path}}&(\mathbf{x}_{t}, \mathbf{u}_{t}) = w\|FK(\mathbf{x}_{t})-\mathbf{p}_A\|^2 + \\ \nonumber
    &\mathbf{x}_{t}^\top Q\mathbf{x}_{t} + \mathbf{u}_{t}^\top R\mathbf{u}_{t}.
\end{align}

The goal cost only concerns whether the final configurations of the robot achieves the desired final end-effector position:
\begin{align}
    L_{\text{goal}}(\mathbf{x}_{T}) = w\|FK(\mathbf{x}_{T})-\mathbf{p}_A\|^2.
\end{align}

The dynamics in Eq.~\eqref{eq:trajopt_dyn} is represented by the numerical integration of the rigid-body dynamics equation~\cite{featherstone2014rigid}:
\begin{align}
    \mathbf{M}(\mathbf{q})\ddot{\mathbf{q}} + \mathbf{C}(\mathbf{q}, \dot{\mathbf{q}}) = \mathbf{u} + \mathbf{J}(\mathbf{q})\boldsymbol{\lambda},
\end{align}
where $\mathbf{M}\in\mathbb{R}^{n\cross n}$ represents is inertia matrix; $\mathbf{C}\in\mathbb{R}^n$ is the gravitational, centrifugal, and Coriolis forces; $\mathbf{J}$ represents the Jacobian matrix; and $\boldsymbol{\lambda}$ is the contact forces at the end-effector. 

The following inequality constraints are involved:
\begin{align}
    \textbf{state limit: }&\underline{\mathbf{x}} \leq \mathbf{x}_t \leq \overline{\mathbf{x}},\\
    \textbf{control limit: }&\underline{\mathbf{u}} \leq \mathbf{u}_t \leq \overline{\mathbf{u}},\\
    \textbf{collision avoidance: }&d_i(\mathbf{x}) \geq d_{\min},
\end{align}
where $d_i(\mathbf{x})$ represents the distance between the $i^{\rm th}$ collision pair at state $\mathbf{x}$, and $d_{\min}$ denotes the minimum allowable distance to avoid collisions.

\begin{figure*}
\centering
\includegraphics[width=0.9\textwidth]{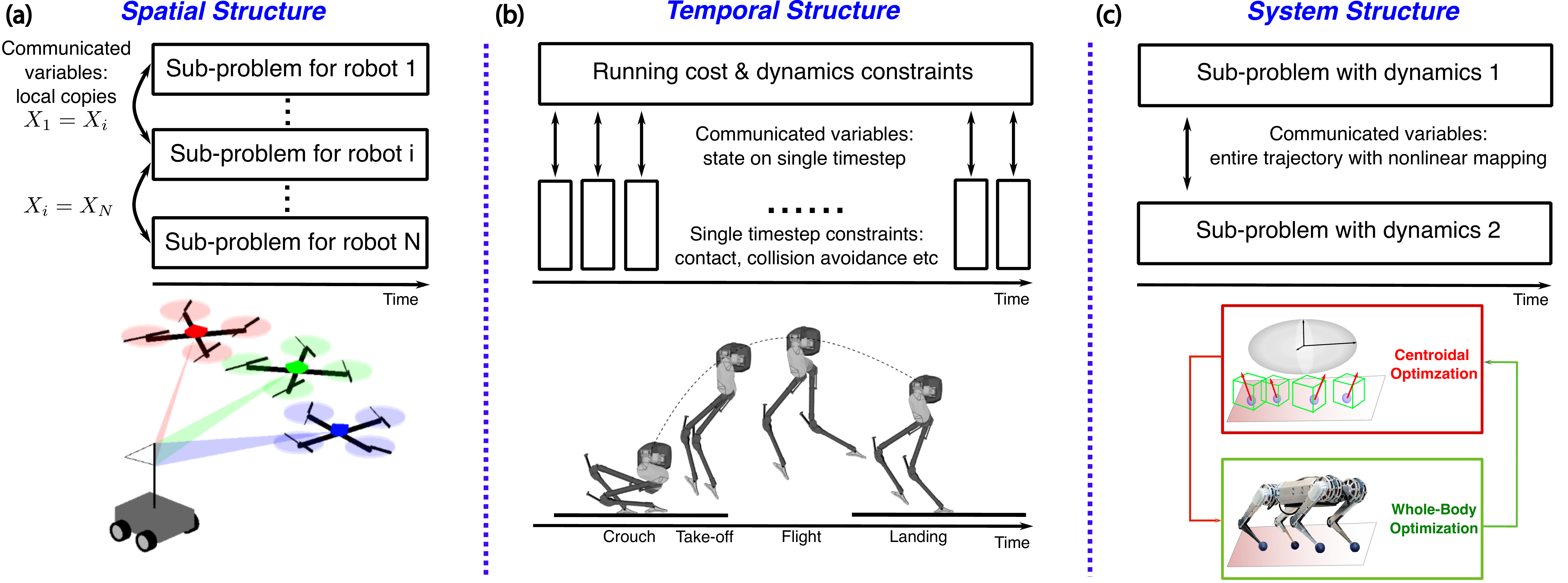}
\caption{Three common structures in distributed TO: (a) spatial structure, illustrated by a multi-agent drone system; (b) temporal structure, illustrated by a Cassie robot jumping over a gap; (c) system structure, illustrated by an alternating centroidal and whole-body optimization for a quadruped.}
\label{fig:admm_structure}
\vspace{-0.15in}
\end{figure*}
\subsection{Distributed Optimization}
Many TO problems have intrinsically distributed structures. Such distributed structures are often formulated and solved via alternating optimization approaches such as ADMM~\cite{boyd2011distributed} in order to improve the efficiency of TO. 
However, the distributed structure might not be immediately apparent and the optimization problems often need to be reformulated into an explicitly distributed format. We focus here on the consensus ADMM as a representative distributed formulation, where copies of decision variables and additional consensus constraints are often introduced to reveal the distributed structure of the TO problems. Consider a optimization problem where the objective is the sum of $N$ functions:
\begin{align}
    \min_{\mathbf{X}}\quad  \sum_{i}^{N}J_{i}(\mathbf{X}),
\end{align}
The optimization can be reformulated in the consensus ADMM format:
\begin{align}
    \min_{\overline{\mathbf{X}}, \mathbf{X}_1,...,\mathbf{X}_N}\quad& \sum_{i}^{N}J_{i}(\mathbf{X})\\
    \text{s.t.}\quad & \mathbf{X}_i = \overline{\mathbf{X}}\quad \forall i \in \{1, ..., N\}\nonumber,
\end{align}
where $\overline{\mathbf{X}}$ is a set of global decision variables.
Practically, one critical aspect to achieve a satisfactory consensus performance is built upon appropriate selection of ADMM parameters through principled mechanisms, such as over-relaxation~\cite{eckstein1992douglas}, varying-penalty parameters~\cite{wohlberg2017admm}, and Nestorov
acceleration method \cite{goldstein2014fast}. 
In the following we focus on the discussion of three structures that can be effeciently solved using consensus ADMM commonly seen in TO problems.

\emph{Spatial Structure: }
The spatial structure of the system can be exploited when the subsystems and their dynamics are separable. This property often exists in multi-robot systems, where the planning of each robot can be treated as a sub-problem. The decision variables are possibily coupled through the objective functions or the constraints (e.g. collision avoidance between robots). Local copies of the full state variables can be created for each robot to decouple the problem~\cite{ferranti2022distributed}, as seen in Fig.~\ref{fig:admm_structure}(a).
Readers are referred to \cite{shorinwa2023distributed, shorinwa2023distributed2} for a detailed review of the multi-robot ADMM. Robustness is further studied in \cite{ni2022robust} given a multi-robot motion planning problem via ADMM.~\cite{amatucci2024accelerating} accelerates TO for loco-manipulation tasks by modeling a quadruped robot with an articulated arm as three sub-robots.

\emph{Temporal Structure: }
TO problems are often formulated in their discretized form. In most cases, the discrete formulation involves a set of decoupled objective and constraint terms that are functions of robot state and control \textit{at a single timestep} (e.g., Eqs.~\eqref{eq:trajopt_boundary_cond} and~\eqref{eq:trajopt_ineq}), and dynamics constraints that couples the states and control trajectory \textit{across consecutive timesteps} (e.g., Eq.~\eqref{eq:trajopt_dyn}). 

For single-timestep objectives or constraints that are computationally expensive (e.g. complementarity constraints for contact \cite{posa2014direct}), it is beneficial to accelerate the optimization process by leveraging the temporal structure of the problem and parallelizing single-timestep objectives and constraints of interest in a distributed fashion(see Fig.~\ref{fig:admm_structure}(b)).

A constraint can be decoupled in a similar fashion by moving the constraint into objective using indicator functions or projection operators. Examples include~\cite{aydinoglu2022real}, where the linear complementarity constraints are independent temporally, and~\cite{le2019fast}, where a $L1$ objective on the control are decoupled. Similarly, in \cite{zhao2021sydebo, wijayarathne2022real}, box constraints are handled separately through a projection operator.

\emph{System Structure: }
The system structure of TO can be exploited when the system dynamics can be characterized by two or more interacting sub-systems, i.e., dynamic models with different complexities. ADMM is used to separate the full optimization problem into sub-problems, each of which corresponds to a sub-system (Fig.~\ref{fig:admm_structure}(c)). This separation often applies to systems with complex robot dynamics with high degrees of freedom. Different from the spatial structure, the system structure often involves non-linear mapping from one sub-system to another one, e.g., a mapping from centroidal dynamics to whole-body dynamics for locomotion problems, as introduced in the next paragraph.

In legged locomotion, there is often a hierarchy of model abstractions, where a whole-body TO and a reduced-order TO are both solved over the planning horizon~\cite{li2021model, khazoom2023optimal}. This hierarchy of model abstractions can be effectively handled via the dynamic splitting strategy of ADMM.
The original rigid body dynamics can be split into centroidal dynamics and whole-body kinematics \cite{herzog2016structured} or dynamics \cite{zhou2022momentum, budhiraja2019dynamics}. Although \cite{herzog2016structured,zhou2022momentum} do not explicitly use ADMM, they iteratively feed optimized trajectory from one sub-system to the other one as the reference trajectory. Empirically, decent results have been reported for converging to local minima \cite{meduri2023biconmp}. A potentially accelerated ADMM updating scheme is also proposed in \cite{zhou2020accelerated}.

\subsection{Learning Methods for Motion Planning}\label{sec:motion_planning_learned}
Despite the improvements in the efficiency of classical TO methods, it remains challenging to achieve real-time TO in many use cases. Moreover, problem-specific objectives and constraints within TO often need to be manually designed, limiting the generalizability of TO approaches. Consequently, learning methods have been extensively explored to facilitate motion generation by: (i) learning objectives and constraints to guide the TO, (ii) learning physical models for integration into TO, and (iii) learning end-to-end policies that imitates the trajectories generated by TO.

\emph{Learned Objectives and Constraints for TO:} 
Objective functions and constraints can be learned from trajectory demonstrations and other task specification inputs such as natural language.
Guided cost learning~\cite{finn2016guided} recovers cost function by adaptively sampling trajectories generated by TO using policy optimization.
For constrained TO scenarios, inverse KKT~\cite{englert2017inverse} learns the cost function and KKT conditions of the underlying constrained optimization problem.
Janner et al.~\cite{janner2021offline} propose to view RL as a generic sequence modeling problem, and then develop a transformer-based architecture to model the distribution of the trajectories, and utilize beam search to solve the planning problem. To allow more flexible task specifications, Sharma et al.~\cite{sharma2022correcting} propose to learn neural networks for mapping natural language sentences to transformations of cost functions, which are then used for optimizing the motion trajectories. Along another line of research, LLM has shown promises as a interface to motion planning by describing robot motions and translating desired robot motions into reward functions~\cite{yu2023language} to guide the optimization of control policy. VoxPoser~\cite{huang2023voxposer} leverages LLM to generate cost maps based on task specifications, and then utilizes search algorithms to derive the robot motion trajectory.

\emph{Learned Physical Models for TO}:
Complexities in physical models, especially the discontinuities in contact models can cause significant numerical challenges for TO. These challenges have spurred the development of learned differentiable contact models, despite the noted difficulty in accurately capturing the behavior of stiff contacts~\cite{parmar2021fundamental}. ContactNets~\cite{pfrommer2021contactnets} proposes to learn inter-body distances and contact Jacobians using a smooth, implicit parameterization, which can potentially be integrated with TO. The work in~\cite{bianchini2023simultaneous} extends upon~\cite{pfrommer2021contactnets} to simultaneously learn continuous and contact dynamics using residual networks. For object manipulation problems, ~\cite{le2023differentiable} builds a dynamic augmented neural object model that simulates the geometry and dynamics of an object as well as a differentiable contact model.~\cite{driess2022learning} proposes to learn object representations as signed distance fields, which are particularly suitable for optimization-based planning approaches.

\emph{End-to-end Policy Learning Guided by TO:} 
To address the inefficiencies encountered in TO and the obstacles associated with executing TO in real-time, research efforts have been made to learn neural policies that imitates the trajectory examples generated by offline TO.
Guided policy search~\cite{levine2013guided, levine2014learning} iteratively trains policy on distributions over guiding samples generated by DDP.
In comparison, the works in~\cite{mordatch2014combining, duburcq2020online, zhao2022adversarially} propose to use ADMM to achieve consensus between neural network policy and trajectory examples provided by TO. OracleNet~\cite{bency2019neural} recovers the motion plans sequentially with learned Recurrent Neural Networks. To address the motion planning problems with task constraints, CoMPNet~\cite{qureshi2020neural} first encodes the task descriptions and environment into latent space, with a Recurrent Neural Network and CNNs, and then sequentially generates the intermediate robot configurations based on the feature embedding, initial configuration, and goal configuration. Similarly, Radosavovic et al.~\cite{radosavovic2023learning} develop a transformer-based framework for tacking the humanoid locomotion task, where the model is first trained in simulation for generating actions in an autoregressive way, and directly deployed in the real world. To handle the multimodal action distribution of low-level skills, diffusion policy~\cite{chi2023diffusionpolicy} iteratively refine the noise into action sequence through a learned gradient field that is conditioned on the observations, which provides stable training and accommodates high-dimensional action sequences. For legged locomotion, ~\cite{viereck2021learning} proposes to learn a neural network that generates the desired centroidal motion real-time, which is subsequently integrated with a whole-body controller.

\section{Integrated Task and Motion Planning}\label{sec:tamp}
\begin{figure*}\centering
\includegraphics[width=0.75\textwidth]{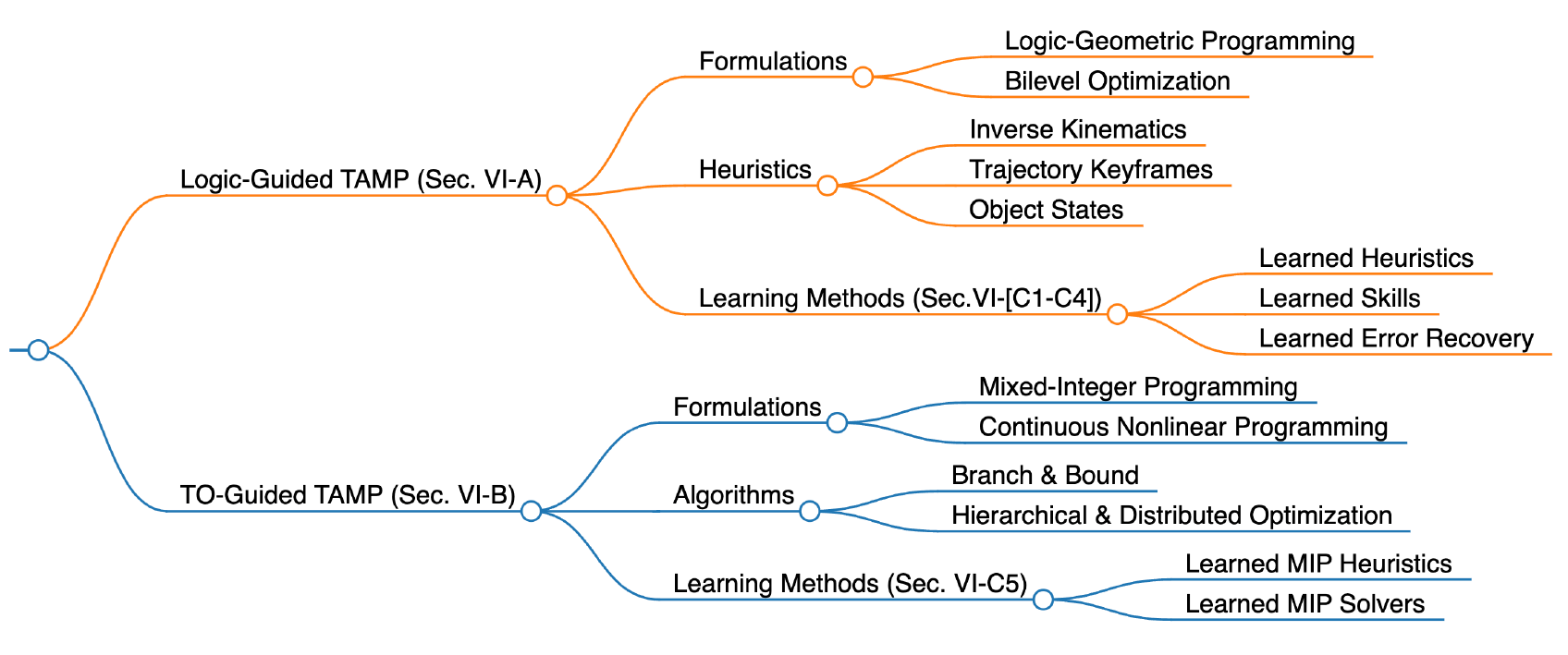}
\caption{Organization and main topics of Integrated TAMP in Sec.~\ref{sec:tamp}. Two main types of optimization-based TAMP are introduced: \textit{logic-guided TAMP} and \textit{TO-guided TAMP}. For each type, the formulations, considerations in classical methods, and relevant learning methods are discussed.}
\label{fig:tamp_overview}
\vspace{-0.15in}
\end{figure*}

Integrated TAMP presents a holistic approach that contrasts with others separately handling task planning (Sec.~\ref{sec:task_planning}) and motion planning (Sec.~\ref{sec:motion_planning}). In optimization-based TAMP, the plan is not merely required to be feasible but also expected to approximate the global optimality. The crucial consideration of integrated TAMP lies in the interdependence between task planning and motion planning. This interplay forms the cornerstone of the design of efficient TAMP algorithms and represents an area of active research. 

The optimization-based formulations for TAMP often involve a hybrid optimization of discrete symbolic-level decisions and continuous motion-level trajectories, as shown in Sec.~\ref{sec:preliminaries_tamp}. 
To this end, we identify two general approaches to formulate and solve the hybrid optimization: \emph{logic-guided TAMP} and \emph{TO-guided TAMP}. {While both approaches inherently solve the hybrid optimization problems, they fundamentally differ in formulations and algorithms, particularly in the definition of discrete variables and the selection of search spaces.}
Fig.~\ref{fig:tamp_overview} shows the overall organization of this section and main topics discussed.
Table~\ref{tab:tamp_summary} presents a representative set of classical approaches for optimization-based TAMP, highlighting their formulations, algorithms, and whether dynamics or kinematics are considered in the application.

\emph{Logic-Guided TAMP} (Sec.~\ref{sec:tamp_logic_guided}) is formulated based on symbolic languages such as PDDL, with the continuous variables and constraints for motion planning embedded as 
continuous-level realization of symbolic planning (referred to as \textit{refinement} hereafter). A notable formulation in logic-guided TAMP is logic-geometric programming (LGP)~\cite{toussaint2015logic}, where logic at the symbolic level governs the constraints imposed on TO, i.e. the motion planner. The algorithm structures for logic-guided TAMP typically involves a state-space search based task planner, with hand-designed heuristics specific to the planning problem (Sec.~\ref{sec:tamp_logic_search}). The motion planner is typically interleaved with the task planner to refine the plan skeleton generated by the task planner (Sec.~\ref{sec:tamp_logic_mmmp}).

\emph{TO-Guided TAMP} (Sec.~\ref{sec:tamp_to_guided}) is formulated as a single TO problem with binary variables that represent discrete decisions. This formulation often views the hybrid optimization problem of TAMP as mixed-integer programming (MIP). Frequently, TO-guided TAMP is derived from temporal logic representations as introduced in Sec.~\ref{sec:domain_representation_temporal_logic}. The methods to solve TO-guided TAMP typically employs general-purpose numerical solvers such as B\&B, without problem-specific heuristics. Unlike logic-guided TAMP, the algorithm's search space is defined not by the explicit state space of the planning problem, but the solution space of the underlying numerical program (Sec.~\ref{sec:tamp_to_bnb}.
Additionally, efforts have been made to improve the scalability of TO-guided TAMP by splitting the MIP into subproblems~\cite{lin2022multi} (Sec.~\ref{sec:tamp_to_dist}) or formulating the planning problem as a fully continuous optimization~\cite{takano2021continuous, gu2023walking} (Sec.~\ref{sec:tamp_to_smooth}).

Despite the progress made in classical optimization techniques for integrated TAMP, these methods still typically have limited scalability due to the combinatorial nature of task planning and numerical complexity of motion planning. One current trend of research is to explore the use of learning-based techniques to improve the efficiency of TAMP algorithms. For logic-guided TAMP, learning methods have been utilized in the interaction between task planning and motion planning layers, for example, learned action feasibility (Sec.~\ref{sec:tamp_learning_feasibility}) and search guidance (Sec.~\ref{sec:tamp_learning_guidance}). Along a different line of research, reusable motion skill acquisition has been explored, which facilitates the efficiency improvement for motion generation in long-horizon tasks (Sec.~\ref{sec:tamp_learning_skills}). For TO-guided TAMP, integrating learning-based techniques to reduce the computational burden of MIP problem has been an active area of research (Sec.~\ref{sec:tamp_learning_mip}).

\begin{table*}[h!]
    \centering
    \begin{tabularx}{0.85\textwidth}
    {c|c|c|c|c}
        \textbf{Papers} & \textbf{Formulation} & \textbf{Mathematical Program} & \textbf{Dynamics/Kinematics} & \textbf{Applications} \\
        \hline
         \cite{migimatsu2020object} & LGP & STRIPS + NLP & Cartesian dynamics & tabletop manipulation \\
         \cite{toussaint2015logic} & LGP & MCTS + KOMO & joint dynamics & tabletop manipulation \\         
         \cite{toussaint2017multi, toussaint2018differentiable} & LGP & MBTS + KOMO & joint
         dynamics & tabletop manipulation \\       
         \cite{zhao2021sydebo} & BO & A* + DDP & joint dynamics & tabletop manipulation\\
         \cite{lo2020petlon} & PDDL or ASP & Heuristics Search + RRT & kinematics & navigation\\
         \cite{stouraitis2020online} & BO & A* + NLP & joint dynamics & dyadic manipulation\\
         \cite{wolff2014optimization} & LTL + MBO & MILP & aerial and mobile robot dynamics & quadrotor and car navigation\\
         \cite{chen2021optimal} & Hybrid Automata & MILP & kinematics & truck and drone delivery\\
         \cite{kogo2021fast} & Hybrid Automata & MILP & Cartesian dynamics & tabletop manipulation\\
         \cite{katayama2020fast} & LTL + MLD & MILP & Cartesian dynamics & manipulation\\
         \cite{adu2022optimal} & PDDL+ & MICP & kinematics & unified loco-manipulation\\
         \cite{saha2017task} & MTL & MILP + gradient descent & Cartesian dynamics & manipulation\\
         \cite{lin2022multi} & MLD & ADMM & joint dynamics & manipulation and locomotion\\
         \cite{pant2018fly} & STL & NLP & aerial dynamics & quadrotor \\
         \cite{takano2021continuous} & STL & NLP & Cartesian dynamics & manipulation\\
         \cite{gu2023walking} & STL & NLP & joint dynamics & locomotion\\
         \hline
         \multicolumn{5}{l}{\rule{0pt}{\normalbaselineskip}\textit{Abbreviations:} MCTS=Monte Carlo Tree Search, MBTS=Multi-bound Tree Search, ASP=Answer Set Programming, KOMO=K-order}\\
        \multicolumn{5}{l}{ Markov Path Optimization, BO=Bilevel Optimization, MBO=Multi-body Optimization, MLD=Mixed Logical-Dynamical System}\\
    \end{tabularx}
    \caption{Overview of classical approaches for optimization-based TAMP}
    \label{tab:tamp_summary}
    \vspace{-0.2in}
\end{table*}

\subsection{Logic-Guided TAMP} \label{sec:tamp_logic_guided}

In logic-guided TAMP, the approach to solving the hybrid planning problem can be conceptualized as constructing a trajectory tree. In this trajectory tree representation, each node corresponds to a symbolic state and each edge represents a trajectory segment. Given the intertwined nature of TAMP, the determination of a symbolic state's feasibility and its associated cost is influenced by a combination of symbolic and continuous domains. 

A naive approach to solve for logic-guided TAMP is to impose a strict hierarchical structure~\cite{garrett2021integrated}, where task planning precedes, followed by motion planning to refine the proposed plan skeleton in continuous domain. This approach hinges on the downward refinement property~\cite{bacchus1994downward}, which posits that for every plan skeleton generated by the task planner, a corresponding continuous motion plan exists. 
However, the downward refinement property does not hold in most real-world scenarios. This necessitates mechanisms for replanning or backtracking at the task planning level upon realizing that a current plan skeleton becomes infeasible in the motion planning level. 

On the other extreme, a fully intertwined algorithm for TAMP might require a call to the motion planner every time a new node in the search tree is expanded, in order to validate the feasibility of the selected symbolic action sequence and to generate a feasible and low-cost continuous motion plan. This method fully determines each symbolic state's reachability and its associated cost is influenced by a combination of planning at symbolic and continuous domains. However, each motion planner call is often computationally expensive, and a majority of symbolic states expanded and trajectory segments solved are unused in the final solution. This often makes the fully intertwined approach computationally intractable.

\begin{figure*}\centering
\includegraphics[width=0.75\textwidth]{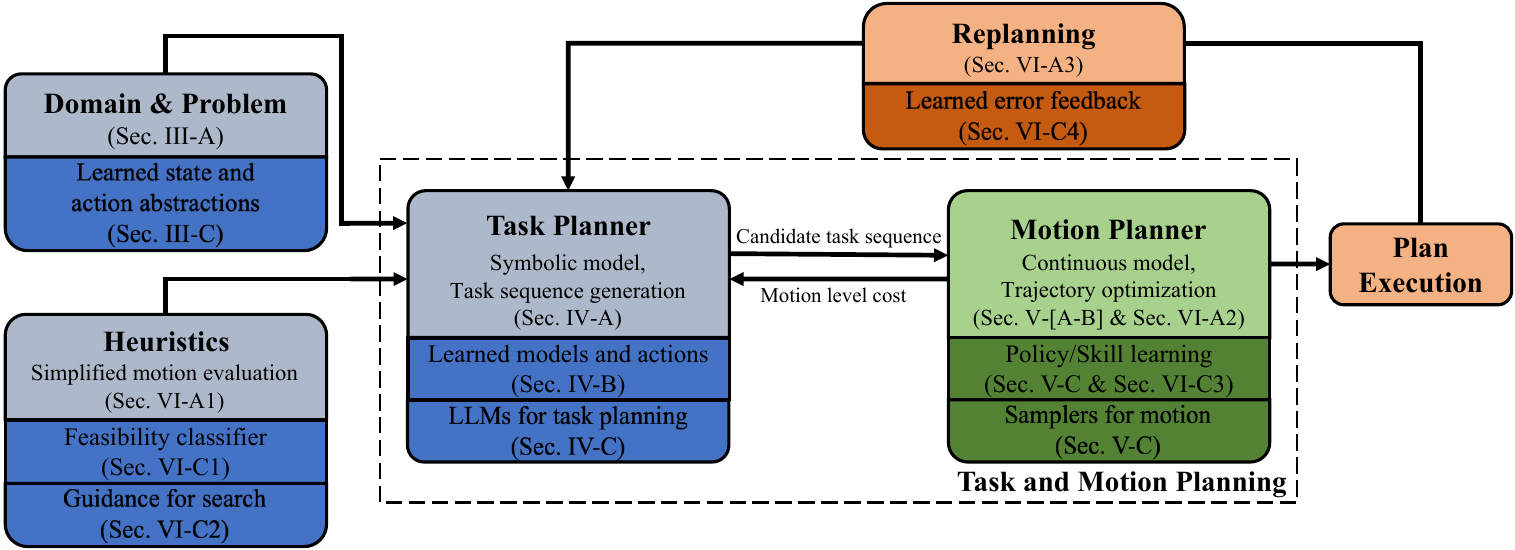}

\caption{Schematic overview of \textit{logic-guided TAMP} and associated learning approaches: blue blocks illustrate domain representation and task planning level; green blocks denote the motion planning level; orange blocks represent plan execution and failure recovery mechanisms.}
\label{fig:logic_guided_tamp}
\vspace{-0.15in}
\end{figure*}

Therefore, the main research question is how to effectively interface between task planning and motion planning layers in order to curtail the size of the search tree and minimize the number of calls to the motion planner, while still effectively solving for feasible and ideally optimal solutions. Fig.~\ref{fig:logic_guided_tamp} illustrates the overall algorithm structure that is commonly seen in logic-guided TAMP.

\subsubsection{Search Heuristics in TAMP}
\label{sec:tamp_logic_search}
Many search heuristics in TAMP attempt to solve a relaxation of the underlying motion planning problem, in order to obtain an estimation of the feasibility and cost of the action. For example,
in TAMP for navigation problems~\cite{lo2020petlon}, Euclidean distance in 2D space serves as an admissible and easily computable heuristic function, which improves planning efficiency while guaranteeing task-level optimality. However, in planning domains with high-dimensional configuration spaces, it is often difficult to generate an analogous distance measure that estimates the action costs. 
One intuitive approach is to evaluate the action feasibility and cost based on the initial and final configurations of the robot or object while ignoring the intermediate trajectories. For example, inverse kinematics (IK) is commonly used to reason about the feasibility of the initial and final robot poses in the action without generating the full trajectory~\cite{akbari2019combined, zhao2021sydebo}. For collaborative robot manipulation, the angular displacement of a manipulated object is often used as heuristics~\cite{stouraitis2020online}.~\cite{agostini2023unified} proposes an object-centric representation of manipulation constraints that unifies TAMP into a single heuristic search that is amenable to existing AI planning heuristics.~\cite{toussaint2015logic} proposes a multi-stage method to solve the TAMP: 1) optimizing over the final configuration given an action sequence, 2) optimizing over all 
kinematics configurations at symbolic state transitions, and 3) optimizing over the entire trajectory. The first two stages effectively act as heuristics during tree search to check the geometric feasibility of a given symbolic action sequence, while the costly TO is only conducted in the final stage.

Heuristics presented so far consider only the initial and final states in a symbolic action. Therefore, no feasibility or cost information about the intermediate states along the trajectory is available, which makes the heuristics easy-to-compute but less informative. This is insufficient to solve more complicated problems, where the path feasibility of the actions plays an important role in the planning process. In comparison,~\cite{toussaint2017multi}, extended from~\cite{toussaint2015logic}, proposes to use a TO with a very coarse time resolution (2 time steps per symbolic action). These heuristics incorporate some path feasibility information while remaining relatively fast to compute, effectively achieving a different level of informativeness-relaxation trade-off.  

Additional work has been done to discover and prune infeasible actions before the corresponding node is reached. If an action is determined to be infeasible by a heuristic function or a motion planner during the search, the same action that exists on other branches of the tree would also be infeasible if no other actions are taken to modify the states relevant to the infeasible action. ~\cite{srivastava2014combined} proposes a planner-independent task-motion interface layer, where additional \texttt{infeasible} predicates are introduced to the task planning domain when an infeasibility is found by the motion planner.~\cite{toussaint2017multi} extends this method to operate in conjunction with Monte-Carlo tree search in an optimization-based TAMP formulation.

\subsubsection{Multi-Modal Motion Planning Solved by TO}
\label{sec:tamp_logic_mmmp}
After a complete or partial plan skeleton is generated by the heuristics search process at the task planning level, the plan skeleton is refined into a continuous trajectory by TO. The problem of TO over a given plan skeleton is akin to the conventional multi-modal motion planning (MMMP) problem proposed in the sampling-based planning community~\cite{hauser2010multi, kingston2022scaling}. TO incorporates the mode transitions and mode constraints derived from the symbolic decisions to form a MMMP problem. Mode constraints are predominantly expressed as manifold constraints in TO. Furthermore, transitions at the symbolic level, often called "symbolic switches", are often represented as continuity constraints between trajectory segments. 

Two main strategies arise to solve the MMMP using TO. The first paradigm emphasizes segment-wise optimization, wherein trajectory segments associated with individual actions in the symbolic sequence are solved independently. For example, LGP-based formulation typically uses k-th order motion optimization~\cite{toussaint2017tutorial} as the underlying motion planner.~\cite{toussaint2018differentiable, toussaint2020describing} further extends LGP to incorporate dynamics constraints and predicates.~\cite{migimatsu2020object} proposes an object-centric TO formulation based on LGP. Similarly,~\cite{zhao2021sydebo} solves the hybrid NLP as trajectory tree, but aims to improve the efficiency of the solver by using ADMM to handle the constraints of the trajectory segments in a distributed mannar. Another line of research attempts to solve for the full motion trajectory as a whole given the symbolic sequence:~\cite{zimmermann2020multi} proposes a multi-level optimization framework that exploits the implicit differentiation method to solve the switch conditions and full trajectories holistically.~\cite{phoon2022constraint} uses a multi-phased TO approach that optimizes the entire motion sequence simultaneously.
{\subsubsection{Receding Horizon TAMP}\label{sec:rh-tamp}
The real-world application of TAMP for long-horizon dynamic tasks is often hindered by failures in plan execution due to changes in the environment, interaction with human, or noisy sensor inputs. Receding horizon TAMP has been explored to mitigate this issue via online replanning. Receding horizon TAMP is analogous to model predictive control (MPC)~\cite{schwenzer2021review}, where the planning problem is solved iteratively over a receding time window. The specific challenge in receding horizon TAMP is to appropriately define the finite time horizon over the hybrid planning domain. The works in~\cite{hartmann2020robust, zhao2021sydebo} rely on task-specific decomposition, where each receding horizon planning iteration achieves the goal for a subtask.~\cite{castaman2021receding, braun2022rhh} propose to plan over a fixed action-horizon, where a full task plan is generated in each iteration, while the motion plan is only computed for a predefined number of actions.~\cite{chen2022interactive} develops branch-MPC, where the objective function is optimized over a scenario tree, which is constructed by enumerating the predicted environmental responses.}

\begin{figure}
\centering
\includegraphics[width=0.45\textwidth]{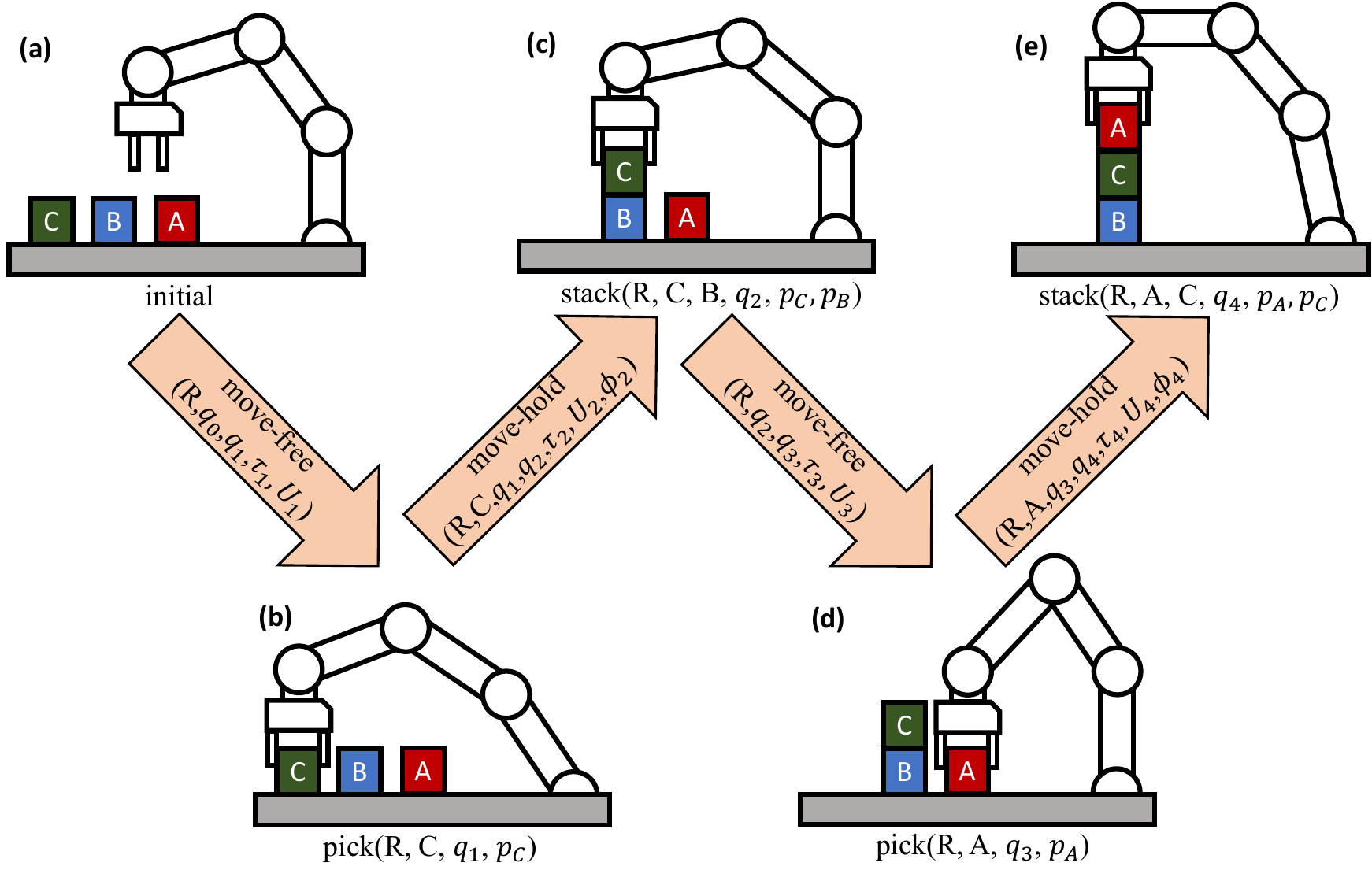}
\caption{
Illustration of a plan skeleton for the tabletop manipulation example. (a)-(e) represent the plan to stack block A on top of C and B using \texttt{pick} and \texttt{stack} actions; the arrows denote the associated \texttt{move-free} and \texttt{move-hold} actions and their required continuous decision variables.}
\label{fig:search_example}
\vspace{-0.15in}
\end{figure}

\emph{Example: }
To address the tabletop manipulation scenario through TAMP, the initial step involves the task planner generating a plan skeleton using tree search. This process involves constructing a tree where nodes represent potential states of the environment and edges represent actions, such as moving or stacking the objects $A, B, C$. The objective is to find a sequence of actions that leads to the desired configuration with maximal height for object $A$. An example plan skeleton can be seen in Fig.~\ref{fig:search_example}.

During the tree search, IK is employed as a heuristic function to check the feasibility of actions. This involves determining whether the robot can physically reach and manipulate the objects as required by the actions in the plan skeleton. The use of IK as a heuristic aids in efficiently pruning the search tree by quickly eliminating infeasible actions, thereby focusing the search on promising solution paths.
Once a preliminary plan skeleton is generated, it is passed to a multi-modal motion planner, which refines this skeleton into a detailed, executable plan. This process is repeated iteratively until the optimal plan is reached or the allocated planning time elapses.

\subsection{TO-Guided TAMP} \label{sec:tamp_to_guided}
In contrast with the approaches discussed in Sec.~\ref{sec:tamp_logic_guided}, where the interactions between task planning and motion planning are expressed \textit{explicitly}, the common methods to solve TO-guided TAMP typically rely on internal features of numerical algorithms such as branch-and-bound (B\&B) and ADMM to achieve interplay between the discrete and continuous decision variables \textit{implicitly}.

\subsubsection{Branch-and-Bound Methods}
\label{sec:tamp_to_bnb}
One typical approach to solve MIP is B\&B-based algorithms~\cite{lawler1966branch, huang2021branch}. 
This method partitions the solution space into smaller subsets (branching) and uses bounds on the objective function to eliminate regions that do not contain an optimal solution. Initially, integer constraints are relaxed to provide an initial bound. The algorithm then branches based on fractional integer variable values, constructing a search tree. By assessing bounds for each subproblem and pruning branches that cannot improve the current best solution, B\&B converges to the global optimum after multiple iterations. However, MIP is classified as a NP-hard problem~\cite{schrijver2003combinatorial}, therefore several branching heuristics are commonly used~\cite{berthold2006primal} to improve scalability, analogous to the state-space search heuristics discussed in Sec.~\ref{sec:tamp_logic_guided}.
For example, strong branching heuristics~\cite{dey2023theoretical} aims to produce a small B\&B tree by selecting the variable to branch that will result in the best improvement of the objective function. Alternatively, local neighborhood search~\cite{fischetti2003local} attempts to improve upon existing feasible solutions by local search.

B\&B-based algorithms are widely implemented in commercial solvers such as Gurobi~\cite{gurobi}, Mosek~\cite{mosek}, and Matlab~\cite{MATLAB}. However, many off-the-shelf implementations are only able to efficiently solve Mixed Integer Linear Programming (MILP) or Mixed Integer Convex Programming (MICP). Therefore, one commonly adopted strategy is to formulate the TAMP problems as MILP or MICP in order to effectively leverage the commercial MIP solvers.

From the formal control community such as temporal logic,~\cite{wolff2014optimization} avoid reactive synthesis by directly encoding LTL formula as mixed-integer linear constraints on nonlinear systems, and aim to find an optimal control sequence.
~\cite{chen2021optimal} encodes the LTL-based hybrid planning problem as a MILP by fixing the number of automaton runs and reasoning over temporally concurrent goals.
~\cite{kogo2021fast} integrates an existing TAMP model with collision avoidance using an MILP formulation with hard constraints on collision and soft constraints on goal positions.
~\cite{katayama2020fast} proposes an object-oriented MILP formulation for dual-arm manipulation by representing the LTL formulas, robot end effector dynamics, and object dynamics as a mixed logical dynamical (MLD) system. 
~\cite{adu2022optimal} proposes Grounded Task Planning as Mixed Integer Programming (GTPMIP), which builds a Hybrid Funnel Graph (HFG) from the hybrid planning problem description in PDDL+, and encodes the HFG as an MICP.

\subsubsection{Hierarchical and Distributed Optimization Methods}
\label{sec:tamp_to_dist}
\begin{figure}
\centering
\includegraphics[width=0.45\textwidth]{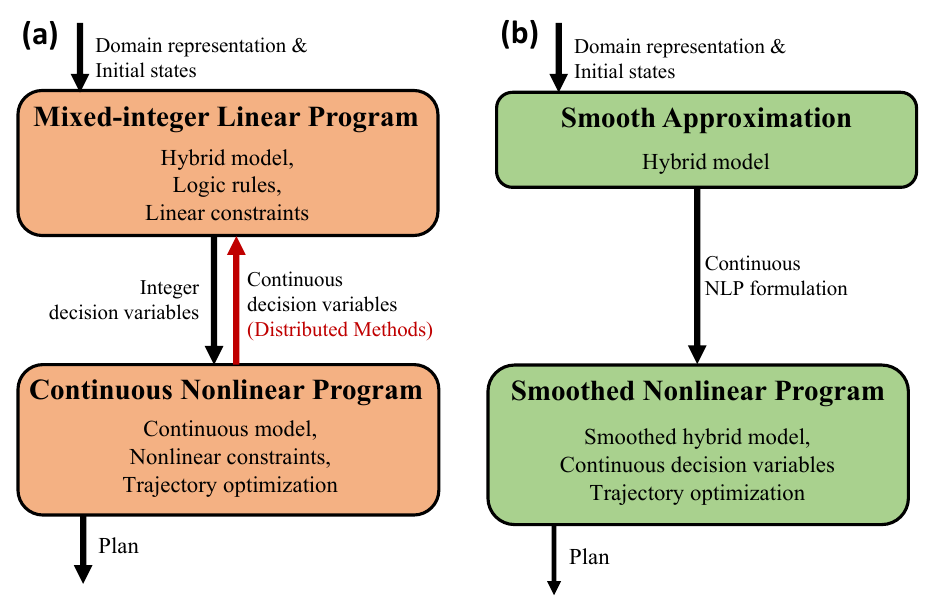}
\caption{Example algorithm structures for \textit{TO-guided TAMP}: (a) hierarchical and distributed methods, (b) smoothed approximation methods.}
\label{fig:to_guided_tamp}
\vspace{-0.15in}
\end{figure}

For systems subject to nonlinear dynamics, the optimization formulation extends to mixed integer nonlinear programming (MINLP). However, the computational burden associated with MINLP is often prohibitive, making them impractical for many real-world applications. In order to manage the computational complexity, the MINLP are often reformulated by decomposing it into solvable sub-problems in a hierarchical or distributed fashion (Fig.~\ref{fig:to_guided_tamp}(a)). 

For the hierarchical methods,~\cite{saha2017task} proposes a hierarchical framework that uses MTL to express specifications for object manipulation tasks and encodes them into a MILP at the high level to solve for task sequence and manipulation poses. Meanwhile, it employs a gradient-descent-based optimization at the low level to compute collision-free robot trajectory.
Similarly, ~\cite{funk2022graph} solves robot assembly discovery problem via a tri-level hierarchical planning structure, where the high level solves a MILP for object arrangement. The work of
~\cite{shamsah2023integrated, warnke2020towards} address bipedal locomotion problem in partially observable environment with a LTL-based task planner and a reduced-order model motion planner.
The main drawback of these hierarchical frameworks is that the low-level TO only attempts to refine the high-level candidate solutions with an detailed continuous-level trajectory, but cannot influence the high-level MILP to achieve a better discrete solution.

{To address the communication issue between the high and low level in hierarchical methods above, the distributed methods}~\cite{lin2022multi, shirai2022simultaneous} use ADMM to convert a MINLP into a consensus problem between a MICP which involves the logical rules and discrete variables, and a continuous NLP that involves the nonlinear kinematics and dynamics. The MICP and NLP shares mutual information through the consensus constraints.
The ADMM-based algorithm for MINLP is demonstrated to be effective in a modular robot climbing \cite{lin2022multi} and manipulation task \cite{shirai2022simultaneous}.
{However, ADMM relies on augmented Lagrangian method, which assumes all decision variables are continuous. Consequently, the presence of integer variables in TAMP can impede ADMM's convergence. To circumvent this limitation, modifications to the ADMM algorithm are required. One potential solution involves the direct copying of integer solutions across subproblems, effectively bypassing the primal-dual update process for integer variables~\cite{lin2022multi}.}

\subsubsection{Smooth Approximation Methods} 
\label{sec:tamp_to_smooth}
Different from the hierarchical and distributed methods above,
another line of research to circumvent the combinatorial complexity of MIP is reformulate the hybrid optimization into a continuous NLP with a specific cost function or constraint representing the smooth approximation of the discrete task planning (Fig.~\ref{fig:to_guided_tamp}(b)). Such a formulation can be solved more efficiently with gradient-based solvers. Recent works in STL utilize smooth approximations of a task specification formula and encode the corresponding robustness degrees into the NLP cost functions~\cite{pant2017smooth, pant2018fly, mehdipour2019arithmetic, gilpin2020smooth}. The work in~\cite{takano2021continuous} builds on the smooth approximation approach but focuses on handling multiple dynamic modes in robot manipulation tasks.~\cite{gu2023walking, gu2024robust} apply the STL specifications and robustness degrees on the push recovery scenarios for bipedal locomotion.
~\cite{envall2023differentiable} proposes a differentiable scheme for multi-arm manipulation problems by treating robot task assignment implicitly as continuous constraints that associate the states of robots and objects.
{Analogous to receding horizon TAMP discussed in Sec.~\ref{sec:rh-tamp}, MPC-based methods have been developed based on STL to enable formal guarantees or reasoning about robustness of the task satisfaction in an online fashion~\cite{gu2024robust, sun2022multi, farahani2015robust, sadraddini2015robust, sadigh2016safe}.}

\subsection{Learning for combined TAMP}\label{sec:tamp_learned}
{Classical TAMP frameworks~\cite{garrett2021integrated, kaelbling2011hierarchical} require accurate, special-purpose perception systems and hand engineered manipulation skills, rendering these approaches less effective while handling novel problems. To overcome this issue, in recent years, there has been extensive exploration of learning techniques within TAMP community. Data-driven approaches allow robots to make informed decisions based on prior examples and experiences, which enhance flexibility and generalizability. Furthermore, the scalability of classical TAMP methods is often limited by the problem size of the tree search for complex problems and the computational cost to evaluate heuristics and optimal trajectories. Learning-based methods show potential to accelerate or replace some of the computationally expensive components of classical methods, such as feasibility checking, search guidance, and skill learning.} Categorizing the works based on the roles and functionalities of the learned components, we primarily classify them into the following five categories. Note that, although some of the low-level motion planners used by works cited in this section might not be optimization-based, the methodologies discussed here are fundamental and highly relevant to optimization-based TAMP. A list of representative learning approaches is presented in table~\ref{tab:learning}.

\begin{table*}[]
    \centering
    \begin{tabular}{M{0.06\linewidth}|M{0.2\linewidth}|M{0.25\linewidth}|M{0.1\linewidth}|M{0.2\linewidth}}
\textbf{Papers} & \textbf{Learned Components} & \textbf{Data} & \textbf{Collection Strategy} & \textbf{Application} \\
\hline
\cite{wells2019learning} & Feasibility Classifier & object states; motion feasibility & offline & tabletop manipulation \\
\cite{noseworthy2021active} & Feasibility Classifier & abstract action sequence; plan feasibility & online & tabletop manipulation \\
\cite{yang2023sequence} & Feasibility Classifier &  abstract action sequence, goal descriptions, images, and object states; plan feasibility & offline & mobile manipulation \\
\cite{silver2021learning} & State and Action Abstractions & low-level transitions with object states and actions; symbolic predicates & offline & mobile manipulation \\
\cite{wang2022generalizable} & State and Action Abstractions & low-level transitions with images, segmentation masks, and actions; symbolic predicates &  offline & mobile manipulation \\
\cite{xu2021deep} & State and Action Abstractions & low-level transitions with images and actions; skill feasibility & offline + online & tabletop manipulation \\
\cite{liang2022search} & State and Action Abstractions & low-level transitions with object states and actions; skill execution cost & offline + online & tabletop manipulation \\
\cite{10227514} & Control Policy & low-level transitions with object states and actions; rewards & online & tabletop manipulation \\
\cite{mandlekar2023hitltamp} & Control Policy & low-level transitions with object states and actions & offline & tabletop manipulation \\
\cite{silver2022learning} & Control Policy; State and Action Abstractions & low-level transitions with object states and actions; symbolic predicates & offline & tabletop manipulation \\
\cite{cauligi2020learning} & Combinatorial Decisions & MICP solutions & offline & free-flyer \& dexterous grasping \\
\hline  
\end{tabular}
\vspace{0.05in}
\caption{Overview of learning techniques for planning}
\label{tab:learning}
\vspace{-0.25in}
\end{table*}

\subsubsection{Learning Feasibility Classifier}\label{sec:tamp_learning_feasibility}
Traditionally, TAMP methods leverage the geometric and dynamic information in the continuous domain to determine the feasibility of tasks during discrete search. However, it can be challenging to incorporate this feasibility check mechanism into a discrete planner. A single feasibility check might involve computationally expensive operations such as collision checking, IK, or even TO. Additionally, the selection of an action or associated geometric parameters can have long-horizon implications on the feasibility of the plan.

To address these issues, Wells et al.~\cite{wells2019learning} propose to train a classifier for evaluating feasible motions and use the classifier as a heuristic for discrete task plan search. Driess et al.~\cite{driess2020deep} propose to learn a neural model for evaluating the hypothesized discrete actions based on visual images. Noseworthy et al.~\cite{noseworthy2021active} leverage active learning to efficiently collect the data for training the plan feasibility classifier, and then utilize the learned classifier to guide the planning and execution. Xu et al.~\cite{xu2022accelerating} similarly train a feasibility classifier with a neural network, which estimates the feasibility of proposed TAMP actions from images of the robot's workspace. To avoid exhaustively reevaluating infeasible motion-level actions, Sung et al.~\cite{sung2023learning} propose learning backjumping heuristics to identify infeasible actions for efficient backtracking during the discrete search. Alternatively, Yang et al.~\cite{yang2023sequence} develop a transformer-based framework for directly predicting the feasibility of finding motion trajectories for the given task plan conditioned on the environment state. Curtis et al.~\cite{curtis2022long} learn to predict the affordances of actions from color and depth images, which helps the TAMP solver generalize to environments with unknown object models.

\subsubsection{Learning Search Guidance}\label{sec:tamp_learning_guidance}
When dealing with planning challenges with extended continuous state-action spaces, relying on random uniform sampling of action parameters without guidance until a path to a goal is discovered proves to be extremely inefficient. Additionally, gradient-based methods frequently struggle when the optimization manifold of a specific problem lacks smoothness. To tackle these challenges, some researchers propose to learn samplers to speed up the TAMP solver for sequential manipulation. Wang et al.~\cite{wang2018active, wang2021learning} propose to jointly learn the action samplers and the conditions of the models. Kim et al.~\cite{kim2018guiding} propose to learn an action sampling distribution with adversarial training to guide the search towards the task goal, then they develop a score space representation and leverage it for transferring constraints to novel situations~\cite{kim2019learning}, which facilitates the speed-up of the search in TAMP. Chitnis et al~\cite{chitnis2016guided} propose to formulate the task plan refinement as a Markov decision process and leverage RL to learn a policy to guide the task plan search. Similarly, Kim et al.~\cite{kim2020learning} develop an abstract representation of states and goals, and learn a value function using graph neural networks to guide TAMP. Ortiz et al.~\cite{ortiz2021learning} propose to represent the problem as constraint graphs and break the overall problem into smaller sequential sampling problems, which are solved by learning assignment orders with Monte-Carlo Tree Search, then they propose to utilize generative models for learning to sample solutions on constraint manifolds~\cite{ortiz2022structured}.

{For complex scenarios involving heterogeneous multi-agent systems, the planning framework must effectively handle task allocation and scheduling. Traditionally, these problems are addressed using heuristics-based approaches, but the combinatorial growth of the search space makes them computationally demanding. To tackle this challenge, reinforcement learning (RL) methods have been widely explored for both task allocation~\cite{noureddine2017multi, ding2021graph} and task scheduling~\cite{shyalika2020reinforcement}. These studies demonstrate RL's capability to significantly improve the time and memory requirements for multi-agent planning.}

\subsubsection{Learning Skill Policies}\label{sec:tamp_learning_skills}
In recent advancements, some researchers have proposed the concept of acquiring foundational skills within the operational framework of TAMP systems. Notably these methods differ from conventional motion planning by learning skill policies, which are capable of generating trajectories through rollout and can be viewed as implicit motion plans. Learning skill policies at motion planning level has shown significant potential to improve the overall efficiency of TAMP due to the reusability and adaptability of the skills.
McDonald et al.~\cite{mcdonald2022guided} propose to distill the knowledge of a TAMP system into a hierarchical RL policy, where they first leverage the TAMP solver to generate supervision data for imitation learning, and then the learned control policies are utilized to further speed up the TAMP solver. LEAGUE~\cite{10227514} proposes to learn RL policies with the guidance of a task planner, where the acquired skills are reused in the TAMP system to accelerate the learning of new skills, which progressively grows its capability for solving long-horizon manipulation tasks in a more efficient manner. The work in~\cite{li2024league++} extends~\cite{10227514} to leverage LLMs to guide skill learning. HITL-TAMP~\cite{mandlekar2023hitltamp} develops an efficient teleoperation system that leverages TAMP to reach the beginning state of the demonstration phase, where the low-level control policies are learned with the collected data and integrated into the system in the testing stage. Meng et al.~\cite{meng2023signal} propose to learn control policies for satisfying the long-term tasks specified with STL, where the learned models are used to generate trajectories via MPC during execution. Silver et al.~\cite{silver2022learning} propose to jointly learn the symbolic operators and low-level skill policies with demonstration data, the extracted operators are first used to generate abstract task plans, and the learned policies are then invoked for generating motions to achieve subgoals.

\subsubsection{Learning for Error Recovery}\label{sec:tamp_learning_recovery}
Robots tend to make mistakes during  the execution phase of TAMP in unstructured environments. 
Therefore, robust policies are critical to both prevent and recover from such errors.
Pan et al.~\cite{pan2022failure} propose a TAMP framework that accounts for potential failures during execution, enabling the robot to calculate and perform necessary actions to achieve the goal despite potential failures. 
The strength of this framework lies in its continuous reassessment and adjustment of the basic beliefs associated with actions, minimizing the likelihood of execution failures.
Wang et al.~\cite{wang2019learning} use multimodal information to create a system to improve the robustness of robot manipulation tasks in unstructured environments.
This procedure involves developing a multimodal state transition model, grounded on task contact dynamics and observed transitions.
Similar to~\cite{wang2019learning}, Luo et al.~\cite{luo2021endowing} use learning-from-demonstration techniques to enable error recovery in robots.

Recently, vision language models (VLMs) and LLMs have been utilized in error recovery. 
For instance, Zhang et al.~\cite{zhang2023grounding} develop a system, called TPVQA. 
This system leverages VLMs to identify action failures and validate action possibilities, thereby increasing the likelihood of successful plan execution.
REFLECT~\cite{liu2023reflect} uses LLMs to autonomously identify and explain robot failures. 
It creates a hierarchical summary of past robot experiences, utilizing multi-sensory data, systematically formulates plans to correct failures, and effectively managing complex tasks.
SayPlan~\cite{rana2023sayplan} offers an error recovery strategy via an iterative replanning process. 
It produces an initial plan tested in a simulator, and any unexecutable actions detected are reported back to the LLM for plan refinement. 
This iterative feedback system, a method of error recovery learning, enables the LLM to steadily boost its planning efficiency, adapting to expansive, complex environments.
Huang et al.~\cite{huang2023inner} leverages three types of environmental feedback in LLMs to generate task plans and handle failures. 

\subsubsection{Learning for Mixed-Integer Programming}\label{sec:tamp_learning_mip}
One main limitation of MIP-based methods for TAMP is that MIP is expensive to solve if a large number of integer variables are involved. 
Recent works in the optimization literature have shown promises to learn the branching heuristics and initializations for MIP.~\cite{zhang2023survey, bengio2021machine} provide detailed surveys for learning-based MIP solvers.
~\cite{marcos2014supervised} first proposes to imitate the strong branching heuristics via supervised learning. The work in~\cite{khalil2016learning} similarly imitates strong branching but develops a framework to solve MIP in an instance-specific manner.~\cite{gasse2019exact} further extends the imitation learning method to incorporate GNN-based models and represent the MIP problem as a bipartite graph. Along another line of research, ~\cite{song2020general, sonnerat2021learning, liu2022learning} combine RL with local neighborhood search for MIP and learn to determine the neighborhood selection, initialization and search neighborhood, and neighborhood size, respectively. 

Aside from the advancements in  learning-based general MIP solvers, recent works also specifically explore learning for MIP in TAMP tasks.~\cite{srinivasan2021fast} trains a neural network offline that imitate the solution of MILP solvers as a warm-start to online multi-robot planning tasks. Cauligi et al.~\cite{cauligi2020learning, cauligi2021coco} speeds up MICP for robot planning and control by solving a dataset of MICP offline and learning the combinatorial decisions via a strategy classifier. The classifier is used to predict the discrete variables and the remaining convex program is solved online during execution.~\cite{deits2019lvis} learns the optimal value function from mixed-integer optimizations to guide policy search.

\section{Future Challenges and Opportunities}
\label{sec:challenges}

\emph{Foundation Models for TAMP:}
The integration of LLMs and VLMs in robot planning is emerging but faces a few challenges.
In robot planning, it is crucial to break down complex task specifications into actionable steps suited for particular environments~\cite{ajay2024compositional,qiu2023large,li2022systematic,zhao2024large,chen2023say}.
However, LLMs and VLMs often struggle with this. 
Their plans may be too abstract, failing to consider the practical constraints of the physical world. 
This presents a significant issue for robots that need concrete and feasible instructions for physical operations.
Furthermore, the current capabilities of LLMs and VLMs are limited, which in turn restricts their effectiveness in robot planning.
For instance, spatial reasoning is essential in robot planning, yet LLMs and VLMs may not accurately understand physical spaces and dynamic environments~\cite{chen2024spatialvlm,rocamonde2023vision,chen2023large}. 
Robot planning also requires considering historical data and long-term goals. 
The limited short-term memory of LLMs could result in information loss during sequential or multi-stage tasks, impacting the coherence and efficiency of planning~\cite{liu2023think,wang2024augmenting,kannan2023smart}. 
Although recent developments aim to enhance the capabilities of LLMs and VLMs, they seem not to fundamentally solve these limitations~\cite{chen2024spatialvlm,yang2023set}.
Additionally, in open-world settings such as homes, malls, and hospitals,
robots also need the ability to adapt to new, unforeseen tasks. Despite the complexities, progress in developing LLMs and VLMs for these purposes is emerging~\cite{chen2023open, wang2023gensim}.

\emph{Diffusion Models for TAMP: }
The use of diffusion models in motion planning, such as diffuser~\cite{janner2022planning}, decision diffuser~\cite{ajay2022conditional} and diffusion policies~\cite{chi2023diffusionpolicy, ze20243d}, has been explored for their flexibility and composability. \cite{pan2024modelbased} proposes a model-based diffusion planner that solves trajectory optimization using the diffusion process without external data. In the context of TAMP, diffusion models can serve as trajectory samplers for individual skills, offering a robust method for generating diverse and feasible motion plans. \cite{mishra2023generative} presents generative skill chaining, where short-horizon skill-centric diffusion models are learned and a compositional framework is established to directly generate long-horizon plans given a plan skeleton. \cite{fang2023dimsam} integrates diffusion skill samplers into a classical TAMP method, which adapts the framework to partially observable planning domains. Promising future research can explore the synergy between LLMs and diffusion models to develop generative multi-task models and end-to-end TAMP frameworks.

\emph{Multi-Modal Sensing for TAMP: }
Currently, most TAMP frameworks rely predominantly on visual sensing. However, integrating multi-modal sensing, such as visual, tactile, and acoustic modules, can significantly improve the robots' capabilities for contact-rich tasks in imitation learning~\cite{lee2020making, yu2023mimictouch, li2023see}, since each sensing modality provides unique and useful contact information related to the manipulation task that covers a wide range of geometric scales and frequency bandwidths. Future research opportunities include further integrating multi-modal sensing with TAMP by advanced sensor fusion as well as contact information representation, extraction, and utilization. Given these potentials, we can potentially enhance the performance of TAMP with better heuristics for task planning and more accurate contact models for trajectory optimization.

\emph{Policy Learning in and for TAMP}:
In the realm of RL-based policy learning, key obstacles revolve around sample inefficiency ({the cost of trial and error}) and the reliance on meticulously crafted dense reward functions. {These challenges are exacerbated in the realm of complex, long-term tasks and often mandate the initial training of RL algorithms within simulators}, thereby adding complexity to the transfer of acquired behaviors from simulation to real world~\cite{10227514}. While tapping into robot teleoperation data directly from real-world environments could alleviate generalization issues, this strategy demands careful system design to ensure seamless robot-human handover and maintain efficient data collection~\cite{mandlekar2023hitltamp}. {Moreover, constructing diverse task configurations that encompass an adequate range of scenarios in terms of object geometries, spatial arrangements, and lighting conditions is nontrivial, presenting additional challenges for acquiring generalizable models in long-horizon tasks. Additionally, the design of observation and action representations is pivotal in enhancing the generalization and reusability of acquired skills.} 

\emph{TAMP for Locomotion and Manipulation: }
Although the duality between locomotion and manipulation~\cite{mason2018toward} means that they can be viewed as equivalent problems, the distinct nature of locomotion and manipulation presents challenges in applying TAMP more complex robotic systems, e.g., humanoid robots with loco-manipulation capabilities. 
From the TAMP perspective, both involve hybrid planning for dynamic contact interactions between the robot and the environment but differ in the representation and frequency of the hybrid events. Manipulation tasks are usually object-centric~\cite{garrett2021integrated} and occur at lower frequencies for contact switching, focusing on precise interactions with objects. In contrast, locomotion involves robot-centric motions with higher frequency contact switching, typically handled through a hierarchical approach of contact planning followed by trajectory generation~\cite{wensing2023optimization}, often facilitated by a centroidal trajectory planning approach~\cite{zhang2021efficient, mcgreavy2022reachability}. 
This difference in representation and operational frequency raises important research questions in developing a unified TAMP framework for loco-manipulation, requiring a balanced integration of these aspects. 
\cite{sleiman2023versatile} shows promises in integrating graph search and TO for long-horizon loco-manipulation tasks by handling the object-centric and robot-centric tasks separately and transferring offline generated plans to online execution. \cite{sferrazza2024humanoidbench} illustrates the recent trend to use hierarchical RL to solve whole-body loco-manipulation problems for humanoid robots with dexterous hands. 
Still, TAMP for unified locomotion and manipulation remains a challenge, due to complicating factors such as the complexity of dexterous grasping and locomotion planning in uneven terrains. 

\emph{TAMP for Human-Robot Collaboration: }
TAMP for human-robot collaboration (HRC) faces challenges due to the uncertainties of human intentions and behaviors. For effective collaboration, it is essential for robots to predict human's symbolic intentions and continuous motions and integrate this understanding into the planning process~\cite{liu2023task}. Recent development in this area include human-aware task planning~\cite{cheng2021human}, hierarchical planning approaches~\cite{darvish2020hierarchical, faroni2023optimal}, and the incorporation of human motion prediction into LGP~\cite{le2021hierarchical}. 
An emerging area of research in HRC is the exploration of novel communication modalities between humans and robots, for example, the integration of conversation interactions~\cite{zhang2023large} in the robot planning framework. Moreover, human intent can be expressed via physical interactions in physical HRC scenarios, such as hand-over~\cite{kshirsagar2019specifying} and collaborative transport of objects where the object and human dynamics are considered~\cite{mortl2012role}. How physical HRC can be achieved efficiently via TAMP remains an active area of research. These recent trends signals a move towards more intuitive and natural human-robot interactions through both physical and language interfaces.

\emph{TAMP in Real-World Applications: }
Optimization-based TAMP has diverse real-world applications across various industries. In industrial settings, TAMP is used for construction planning~\cite{hartmann2020robust}, and rebar grid traversal~\cite{asselmeier2024hierarchical}. Additionally, TAMP enables unmanned aerial vehicles (UAVs) to navigate complex environments for delivery services and environmental monitoring~\cite{otto2018optimization}, as well as agricultural tasks~\cite{conesa2016route}. In domestic applications, TAMP facilitates household tasks such as cooking~\cite{siburianintegrated}, and manipulation of doors and dishwashers~\cite{sleiman2023versatile}. In lab environment, TAMP is deployed for medical test tube rearrangements~\cite{wan2022arranging}. 
Expanding TAMP to broader real-world applications requires overcoming current challenges such as developing more robust methods to accommodate varying environmental factors, problem settings, and human interactions. Additionally, proper scene understanding and representation that captures spatial and semantic relationships is essential and remains an open research area for deploying TAMP in open-world environments. Furthermore, implementing low-level control for robots in real-world scenarios involves handling environmental variability, sensor noise, and calibration issues, which affect feedback reliability. Addressing these challenges is crucial for effective and safe robot deployment.

\emph{Ethical and Societal Implications: }The deployment of TAMP and robotics raise several ethical and societal considerations~\cite{lin2014robot, wu2022sustainable}, including (i) ensuring the safety and reliability of robot planning, particularly in human-populated environments; (ii) developing strategies to mitigate socio-economic impact of job displacement caused by increased automation; (iii) responsible handling of data collected during training of machine learning models to protect privacy and intellectual property; (iv) minimizing environmental impact via more energy efficient training process for machine learning models and sustainable practices in the the production, operation, and disposal of robots.

\vspace{-0.1in}
{
\section*{Glossaries of Terms}}
\noindent{\footnotesize\textbf{Alternating Direction Method of Multipliers (ADMM): }An algorithm that solves convex optimization problems by breaking it into smaller pieces, each of which is easier to handle~\cite{boyd2011distributed}. 
}

\noindent{\footnotesize\textbf{AI Planning: }An area of artificial intelligence that studies the process of automated generation of sequence of actions to achieve specific goals~\cite{ghallab2004automated}.} 

\noindent{\footnotesize\textbf{Bilevel Optimization: } A mathematical program, where an optimization problem contains another optimization problem as a constraint~\cite{sinha2017review}.}

\noindent{\footnotesize\textbf{Foundation Model: }A general purpose model, typically trained on a large amount of data, that can be adapted for various downstream applications~\cite{bommasani2021opportunities}.}

\noindent{\footnotesize\textbf{Heuristics: }Methods to organize the search space and guide the search algorithm~\cite{ghallab2004automated}.}

\noindent{\footnotesize\textbf{Large Language Model: }A large pretrained language model designed to understand and generate human-like text~\cite{liu2023pre}.}

\noindent{\footnotesize\textbf{Mixed Integer Programming: }An optimization problem with a combination of continuous and discrete decision variables~\cite{schrijver2003combinatorial}.}

\noindent{\footnotesize\textbf{Mixed Logical Dynamical System: }A formulation of system dynamics with mixed integer constraints~\cite{belta2019formal}.}

\noindent{\footnotesize\textbf{Model Predictive Control: }A class of control methods where a model is used to predict the future of the controlled system over a receding planning horizon~\cite{schwenzer2021review}.}

\noindent{\footnotesize\textbf{Motion Planning: }The problem of moving a mechanical system from a start state to a goal region~\cite{orthey2023sampling}.}

\noindent{\footnotesize\textbf{Nonlinear Programming: }An optimization problem where the objective function or constraints are nonlinear~\cite{betts1998survey}.}

\noindent{\footnotesize\textbf{Reinforcement Learning: }A problem where an agent learns to make decisions through a goal-directed interaction with the uncertain environment~\cite{sutton2018reinforcement}.}

\noindent{\footnotesize\textbf{Task and Motion Planning: }A hybrid planning problem that integrates high-level task planning and low-level motion planning, which enables reasoning over long-horizon, dynamic tasks~\cite{garrett2021integrated}.}

\noindent{\footnotesize\textbf{Task Planning: }The problem of generating an action sequence that accomplishes the goal of the task and satisfies the task specifications~\cite{meli2023logic}.}

\noindent{\footnotesize\textbf{Temporal Logic: }The formal methods for describing time dependent rules and symbols, often used to specify the correctness of a finite-state transition system~\cite{belta2019formal}.}

\noindent{\footnotesize\textbf{Trajectory Optimization: }The problem of generating a continuous robot motion path and a control sequence that optimizes an objective function subject to a set of kinematics and/or dynamics constraints~\cite{betts1998survey}.}

{\footnotesize  
\bibliographystyle{IEEEtran}
\bibliography{references.bib}

\begin{thebibliography}{100}
\providecommand{\url}[1]{#1}
\csname url@samestyle\endcsname
\providecommand{\newblock}{\relax}
\providecommand{\bibinfo}[2]{#2}
\providecommand{\BIBentrySTDinterwordspacing}{\spaceskip=0pt\relax}
\providecommand{\BIBentryALTinterwordstretchfactor}{4}
\providecommand{\BIBentryALTinterwordspacing}{\spaceskip=\fontdimen2\font plus
\BIBentryALTinterwordstretchfactor\fontdimen3\font minus
  \fontdimen4\font\relax}
\providecommand{\BIBforeignlanguage}[2]{{%
\expandafter\ifx\csname l@#1\endcsname\relax
\typeout{** WARNING: IEEEtran.bst: No hyphenation pattern has been}%
\typeout{** loaded for the language `#1'. Using the pattern for}%
\typeout{** the default language instead.}%
\else
\language=\csname l@#1\endcsname
\fi
#2}}
\providecommand{\BIBdecl}{\relax}
\BIBdecl

\bibitem{adu2022optimal}
A.~Adu-Bredu, N.~Devraj, and O.~C. Jenkins, ``Optimal constrained task planning
  as mixed integer programming,'' in \emph{Proc. IEEE/RSJ Int. Conf. Intell.
  Robots Syst.}, 2022, pp. 12\,029--12\,036.

\bibitem{garrett2021integrated}
C.~R. Garrett, R.~Chitnis, R.~Holladay, B.~Kim, T.~Silver, L.~P. Kaelbling, and
  T.~Lozano-P{\'e}rez, ``Integrated task and motion planning,'' \emph{Annu.
  Rev. Control Robot. Auton. Syst.}, vol.~4, pp. 265--293, 2021.

\bibitem{envall2023differentiable}
J.~Envall, R.~Poranne, and S.~Coros, ``Differentiable task assignment and
  motion planning,'' in \emph{Proc. IEEE/RSJ Int. Conf. Intell. Robots Syst.},
  2023, pp. 2049--2056.

\bibitem{dantam2018incremental}
N.~T. Dantam, Z.~K. Kingston, S.~Chaudhuri, and L.~E. Kavraki, ``An incremental
  constraint-based framework for task and motion planning,'' \emph{Int. J.
  Robot. Res.}, vol.~37, no.~10, pp. 1134--1151, 2018.

\bibitem{lozano2014constraint}
T.~Lozano-P{\'e}rez and L.~P. Kaelbling, ``A constraint-based method for
  solving sequential manipulation planning problems,'' in \emph{Proc. IEEE/RSJ
  Int. Conf. Intell. Robots Syst.}, 2014, pp. 3684--3691.

\bibitem{garrett2020pddlstream}
C.~R. Garrett, T.~Lozano-P{\'e}rez, and L.~P. Kaelbling, ``Pddlstream:
  Integrating symbolic planners and blackbox samplers via optimistic adaptive
  planning,'' in \emph{Proc. Int. Conf. Autom. Planning Scheduling}, vol.~30,
  2020, pp. 440--448.

\bibitem{krontiris2016efficiently}
A.~Krontiris and K.~E. Bekris, ``Efficiently solving general rearrangement
  tasks: A fast extension primitive for an incremental sampling-based
  planner,'' in \emph{Proc. IEEE Int. Conf. Robot. Autom.}, 2016, pp.
  3924--3931.

\bibitem{toussaint2015logic}
M.~Toussaint, ``Logic-geometric programming: An optimization-based approach to
  combined task and motion planning.'' in \emph{Proc. Int. Joint Conf. Artif.
  Intell.}, 2015, pp. 1930--1936.

\bibitem{takano2021continuous}
R.~Takano, H.~Oyama, and M.~Yamakita, ``Continuous optimization-based task and
  motion planning with signal temporal logic specifications for sequential
  manipulation,'' in \emph{Proc. IEEE Int. Conf. Robot. Autom.}, 2021, pp.
  8409--8415.

\bibitem{guo2023recent}
H.~Guo, F.~Wu, Y.~Qin, R.~Li, K.~Li, and K.~Li, ``Recent trends in task and
  motion planning for robotics: A survey,'' \emph{ACM Comput. Surv.}, 2023.

\bibitem{orthey2023sampling}
A.~Orthey, C.~Chamzas, and L.~E. Kavraki, ``Sampling-based motion planning: A
  comparative review,'' \emph{Annu. Rev. Control Robot. Auton. Syst.}, vol.~7,
  pp. 285--310, 2023.

\bibitem{karaman2011sampling}
S.~Karaman and E.~Frazzoli, ``Sampling-based algorithms for optimal motion
  planning,'' \emph{Int. J. Robot. Res.}, vol.~30, no.~7, pp. 846--894, 2011.

\bibitem{schmitt2017optimal}
P.~S. Schmitt, W.~Neubauer, W.~Feiten, K.~M. Wurm, G.~V. Wichert, and
  W.~Burgard, ``Optimal, sampling-based manipulation planning,'' in \emph{Proc.
  IEEE Int. Conf. Robot. Autom.}, 2017, pp. 3426--3432.

\bibitem{gammell2021asymptotically}
J.~D. Gammell and M.~P. Strub, ``Asymptotically optimal sampling-based motion
  planning methods,'' \emph{Annu. Rev. Control Robot. Auton. Syst.}, vol.~4,
  pp. 295--318, 2021.

\bibitem{toussaint2018differentiable}
M.~A. Toussaint, K.~R. Allen, K.~A. Smith, and J.~B. Tenenbaum,
  ``Differentiable physics and stable modes for tool-use and manipulation
  planning,'' in \emph{Proc. Robot. Sci. Syst.}, 2018.

\bibitem{migimatsu2020object}
T.~Migimatsu and J.~Bohg, ``Object-centric task and motion planning in dynamic
  environments,'' \emph{IEEE Robot. Autom. Lett.}, vol.~5, no.~2, pp. 844--851,
  2020.

\bibitem{stouraitis2020online}
T.~Stouraitis, I.~Chatzinikolaidis, M.~Gienger, and S.~Vijayakumar, ``Online
  hybrid motion planning for dyadic collaborative manipulation via bilevel
  optimization,'' \emph{IEEE Trans. Robot.}, vol.~36, no.~5, pp. 1452--1471,
  2020.

\bibitem{aceituno2017simultaneous}
B.~Aceituno-Cabezas, C.~Mastalli, H.~Dai, M.~Focchi, A.~Radulescu, D.~G.
  Caldwell, J.~Cappelletto, J.~C. Grieco, G.~Fern{\'a}ndez-L{\'o}pez, and
  C.~Semini, ``Simultaneous contact, gait, and motion planning for robust
  multilegged locomotion via mixed-integer convex optimization,'' \emph{IEEE
  Robot. Autom. Lett.}, vol.~3, no.~3, pp. 2531--2538, 2017.

\bibitem{sleiman2023versatile}
J.-P. Sleiman, F.~Farshidian, and M.~Hutter, ``Versatile multicontact planning
  and control for legged loco-manipulation,'' \emph{Sci. Robot.}, vol.~8,
  no.~81, p. eadg5014, 2023.

\bibitem{zhao2022reactive}
Y.~Zhao, Y.~Li, L.~Sentis, U.~Topcu, and J.~Liu, ``Reactive task and motion
  planning for robust whole-body dynamic locomotion in constrained
  environments,'' \emph{Int. J. Robot. Res.}, vol.~41, no.~8, pp. 812--847,
  2022.

\bibitem{asselmeier2024hierarchical}
M.~Asselmeier, J.~Ivanova, Z.~Zhou, P.~A. Vela, and Y.~Zhao, ``Hierarchical
  experience-informed navigation for multi-modal quadrupedal rebar grid
  traversal,'' in \emph{Proc. IEEE Int. Conf. Robot. Autom.}, 2024, pp.
  8065--8072.

\bibitem{ghallab2004automated}
M.~Ghallab, D.~Nau, and P.~Traverso, \emph{Automated Planning: theory and
  practice}.\hskip 1em plus 0.5em minus 0.4em\relax Elsevier, 2004.

\bibitem{meli2023logic}
D.~Meli, H.~Nakawala, and P.~Fiorini, ``Logic programming for deliberative
  robotic task planning,'' \emph{Artif. Intell. Rev.}, pp. 1--39, 2023.

\bibitem{wensing2023optimization}
P.~M. Wensing, M.~Posa, Y.~Hu, A.~Escande, N.~Mansard, and A.~Del~Prete,
  ``Optimization-based control for dynamic legged robots,'' \emph{IEEE Trans.
  Robot.}, 2023.

\bibitem{posa2014direct}
M.~Posa, C.~Cantu, and R.~Tedrake, ``A direct method for trajectory
  optimization of rigid bodies through contact,'' \emph{Int. J. Robot. Res.},
  vol.~33, no.~1, pp. 69--81, 2014.

\bibitem{tassa2014control}
Y.~Tassa, N.~Mansard, and E.~Todorov, ``Control-limited differential dynamic
  programming,'' in \emph{Proc. IEEE Int. Conf. Robot. Autom.}, 2014, pp.
  1168--1175.

\bibitem{xu2022accelerating}
L.~Xu, T.~Ren, G.~Chalvatzaki, and J.~Peters, ``Accelerating integrated task
  and motion planning with neural feasibility checking,''
  \emph{arXiv:2203.10568}, 2022.

\bibitem{yang2023sequence}
Z.~Yang, C.~Garrett, T.~Lozano-Perez, L.~Kaelbling, and D.~Fox,
  ``Sequence-based plan feasibility prediction for efficient task and motion
  planning,'' in \emph{Proc. Robot. Sci. Syst.}, 2023.

\bibitem{driess2020deep}
D.~Driess, J.-S. Ha, and M.~Toussaint, ``Deep visual reasoning: Learning to
  predict action sequences for task and motion planning from an initial scene
  image,'' in \emph{Proc. Robot. Sci. Syst.}, 2020.

\bibitem{lin2023text2motion}
K.~Lin, C.~Agia, T.~Migimatsu, M.~Pavone, and J.~Bohg, ``Text2motion: From
  natural language instructions to feasible plans,'' \emph{Auton. Robots},
  vol.~47, no.~8, pp. 1345--1365, 2023.

\bibitem{silver2024generalized}
T.~Silver, S.~Dan, K.~Srinivas, J.~B. Tenenbaum, L.~Kaelbling, and M.~Katz,
  ``Generalized planning in pddl domains with pretrained large language
  models,'' in \emph{Proc. AAAI Conf. Artif. Intell.}, vol.~38, no.~18, 2024,
  pp. 20\,256--20\,264.

\bibitem{huang2022language}
W.~Huang, P.~Abbeel, D.~Pathak, and I.~Mordatch, ``Language models as zero-shot
  planners: Extracting actionable knowledge for embodied agents,'' in
  \emph{Int. Conf. Mach. Learn.}\hskip 1em plus 0.5em minus 0.4em\relax PMLR,
  2022, pp. 9118--9147.

\bibitem{10227514}
S.~Cheng and D.~Xu, ``League: Guided skill learning and abstraction for
  long-horizon manipulation,'' \emph{IEEE Robot. Autom. Lett.}, vol.~8, no.~10,
  pp. 6451--6458, 2023.

\bibitem{mcdonald2022guided}
M.~J. McDonald and D.~Hadfield-Menell, ``Guided imitation of task and motion
  planning,'' in \emph{Proc. Conf. Robot. Learn.}\hskip 1em plus 0.5em minus
  0.4em\relax PMLR, 2022, pp. 630--640.

\bibitem{mansouri2021combining}
M.~Mansouri, F.~Pecora, and P.~Sch{\"u}ller, ``Combining task and motion
  planning: Challenges and guidelines,'' \emph{Front. Robot. AI}, vol.~8, p.
  637888, 2021.

\bibitem{antonyshyn2023multiple}
L.~Antonyshyn, J.~Silveira, S.~Givigi, and J.~Marshall, ``Multiple mobile robot
  task and motion planning: A survey,'' \emph{ACM Comput. Surv.}, vol.~55,
  no.~10, pp. 1--35, 2023.

\bibitem{belta2019formal}
C.~Belta and S.~Sadraddini, ``Formal methods for control synthesis: An
  optimization perspective,'' \emph{Annu. Rev. Control Robot. Auton. Syst.},
  vol.~2, pp. 115--140, 2019.

\bibitem{shorinwa2023distributed}
O.~Shorinwa, T.~Halsted, J.~Yu, and M.~Schwager, ``Distributed optimization
  methods for multi-robot systems: Part 1—a tutorial,'' \emph{IEEE Robot.
  Autom. Mag.}, vol.~31, pp. 121--138, 2024.

\bibitem{silver2021learning}
T.~Silver, R.~Chitnis, J.~Tenenbaum, L.~P. Kaelbling, and T.~Lozano-P{\'e}rez,
  ``Learning symbolic operators for task and motion planning,'' in \emph{Proc.
  IEEE/RSJ Int. Conf. Intell. Robots Syst.}, 2021, pp. 3182--3189.

\bibitem{pnueli1977temporal}
A.~Pnueli, ``The temporal logic of programs,'' in \emph{Annu. Symp. Found.
  Comput. Sci.}\hskip 1em plus 0.5em minus 0.4em\relax IEEE, 1977, pp. 46--57.

\bibitem{de2013linear}
G.~De~Giacomo and M.~Y. Vardi, ``Linear temporal logic and linear dynamic logic
  on finite traces,'' in \emph{Proc. Int. Joint Conf. Artif. Intell.}, 2013,
  pp. 854--860.

\bibitem{maler2004monitoring}
O.~Maler and D.~Nickovic, ``Monitoring temporal properties of continuous
  signals,'' in \emph{Int. Symp. Formal Tech. Real-Time Fault-Tolerant
  Syst.}\hskip 1em plus 0.5em minus 0.4em\relax Springer, 2004, pp. 152--166.

\bibitem{koymans1990specifying}
R.~Koymans, ``Specifying real-time properties with metric temporal logic,''
  \emph{Real-time systems}, vol.~2, no.~4, pp. 255--299, 1990.

\bibitem{aeronautiques1998pddl}
C.~Aeronautiques, A.~Howe, C.~Knoblock, I.~D. McDermott, A.~Ram, M.~Veloso,
  D.~Weld, D.~W. SRI, A.~Barrett, D.~Christianson \emph{et~al.}, ``Pddl|the
  planning domain definition language,'' \emph{Tech. Rep.}, 1998.

\bibitem{haslum2019introduction}
P.~Haslum, N.~Lipovetzky, D.~Magazzeni, and C.~Muise, ``An introduction to the
  planning domain definition language,'' \emph{Synth. Lect. Artif. Intell.
  Mach. Learn.}, vol.~13, no.~2, pp. 1--187, 2019.

\bibitem{fox2003pddl2}
M.~Fox and D.~Long, ``Pddl2. 1: An extension to pddl for expressing temporal
  planning domains,'' \emph{J. Artif. Intell. Res.}, vol.~20, pp. 61--124,
  2003.

\bibitem{kovacs2011bnf}
D.~L. Kovacs, ``Bnf definition of pddl 3.1,'' \emph{Unpublished manuscript from
  the IPC-2011 website}, vol.~15, 2011.

\bibitem{fox2006modelling}
M.~Fox and D.~Long, ``Modelling mixed discrete-continuous domains for
  planning,'' \emph{J. Artif. Intell. Res.}, vol.~27, pp. 235--297, 2006.

\bibitem{younes2004ppddl1}
H.~L. Younes and M.~L. Littman, ``Ppddl1. 0: An extension to pddl for
  expressing planning domains with probabilistic effects,'' \emph{Techn. Rep.
  CMU-CS-04-162}, vol.~2, p.~99, 2004.

\bibitem{kovacs2012multi}
D.~L. Kov{\'a}cs, ``A multi-agent extension of pddl3.1,'' in \emph{Proc. 3rd
  Workshop Int. Planning Compet.}, 2012, pp. 19--37.

\bibitem{emerson1982using}
E.~A. Emerson and E.~M. Clarke, ``Using branching time temporal logic to
  synthesize synchronization skeletons,'' \emph{Sci. Comput. Program.}, vol.~2,
  no.~3, pp. 241--266, 1982.

\bibitem{donze2010robust}
A.~Donz{\'e} and O.~Maler, ``Robust satisfaction of temporal logic over
  real-valued signals,'' in \emph{Int. Conf. Formal Model. Anal. Timed
  Syst.}\hskip 1em plus 0.5em minus 0.4em\relax Springer, 2010, pp. 92--106.

\bibitem{raman2014model}
V.~Raman, A.~Donz{\'e}, M.~Maasoumy, R.~M. Murray, A.~Sangiovanni-Vincentelli,
  and S.~A. Seshia, ``Model predictive control with signal temporal logic
  specifications,'' in \emph{Proc. IEEE Conf. Decis. Control}, 2014, pp.
  81--87.

\bibitem{kurtz2022mixed}
V.~Kurtz and H.~Lin, ``Mixed-integer programming for signal temporal logic with
  fewer binary variables,'' \emph{IEEE Control Syst. Lett.}, vol.~6, pp.
  2635--2640, 2022.

\bibitem{silver2023inventing}
T.~Silver, R.~Chitnis, N.~Kumar, W.~McClinton, T.~Lozano-Perez, L.~P.
  Kaelbling, and J.~Tenenbaum, ``Predicate invention for bilevel planning,'' in
  \emph{Proc. AAAI Conf. Artif. Intell.}, 2023, pp. 12\,120--12\,129.

\bibitem{chitnis2022learning}
R.~Chitnis, T.~Silver, J.~B. Tenenbaum, T.~Lozano-Perez, and L.~P. Kaelbling,
  ``Learning neuro-symbolic relational transition models for bilevel
  planning,'' in \emph{Proc. IEEE/RSJ Int. Conf. Intell. Robots Syst.}, 2022,
  pp. 4166--4173.

\bibitem{chitnis2021camps}
R.~Chitnis, T.~Silver, B.~Kim, L.~Kaelbling, and T.~Lozano-Perez, ``Camps:
  Learning context-specific abstractions for efficient planning in factored
  mdps,'' in \emph{Proc. Conf. Robot. Learn.}\hskip 1em plus 0.5em minus
  0.4em\relax PMLR, 2021, pp. 64--79.

\bibitem{silver2021planning}
T.~Silver, R.~Chitnis, A.~Curtis, J.~B. Tenenbaum, T.~Lozano-P{\'e}rez, and
  L.~P. Kaelbling, ``Planning with learned object importance in large problem
  instances using graph neural networks,'' in \emph{Proc. AAAI Conf. Artif.
  Intell.}, vol.~35, no.~13, 2021, pp. 11\,962--11\,971.

\bibitem{zhu2021hierarchical}
Y.~Zhu, J.~Tremblay, S.~Birchfield, and Y.~Zhu, ``Hierarchical planning for
  long-horizon manipulation with geometric and symbolic scene graphs,'' in
  \emph{Proc. IEEE Int. Conf. Robot. Autom.}, 2021, pp. 6541--6548.

\bibitem{wang2022generalizable}
C.~Wang, D.~Xu, and L.~Fei-Fei, ``Generalizable task planning through
  representation pretraining,'' \emph{IEEE Robot. Autom. Lett.}, vol.~7, no.~3,
  pp. 8299--8306, 2022.

\bibitem{ding2023integrating}
Y.~Ding, X.~Zhang, S.~Amiri, N.~Cao, H.~Yang, A.~Kaminski, C.~Esselink, and
  S.~Zhang, ``Integrating action knowledge and llms for task planning and
  situation handling in open worlds,'' \emph{Auton. Robots}, vol.~47, no.~8,
  pp. 981--997, 2023.

\bibitem{liu2023llm}
B.~Liu, Y.~Jiang, X.~Zhang, Q.~Liu, S.~Zhang, J.~Biswas, and P.~Stone, ``Llm+p:
  Empowering large language models with optimal planning proficiency,''
  \emph{arXiv:2304.11477}, 2023.

\bibitem{singh2023progprompt}
I.~Singh, V.~Blukis, A.~Mousavian, A.~Goyal, D.~Xu, J.~Tremblay, D.~Fox,
  J.~Thomason, and A.~Garg, ``Progprompt: Generating situated robot task plans
  using large language models,'' in \emph{Proc. IEEE Int. Conf. Robot. Autom.},
  2023, pp. 11\,523--11\,530.

\bibitem{zhao2024large}
Z.~Zhao, W.~S. Lee, and D.~Hsu, ``Large language models as commonsense
  knowledge for large-scale task planning,'' \emph{Adv. Neural Inf. Process.
  Syst.}, vol.~36, pp. 31\,967--31\,987, 2024.

\bibitem{ren2023robots}
A.~Z. Ren, A.~Dixit, A.~Bodrova, S.~Singh, S.~Tu, N.~Brown, P.~Xu, L.~Takayama,
  F.~Xia, J.~Varley \emph{et~al.}, ``Robots that ask for help: Uncertainty
  alignment for large language model planners,'' in \emph{Proc. Conf. Robot.
  Learn.}\hskip 1em plus 0.5em minus 0.4em\relax PMLR, 2023, pp. 661--682.

\bibitem{xie2023translating}
Y.~Xie, C.~Yu, T.~Zhu, J.~Bai, Z.~Gong, and H.~Soh, ``Translating natural
  language to planning goals with large-language models,''
  \emph{arXiv:2302.05128}, 2023.

\bibitem{pan2023data}
J.~Pan, G.~Chou, and D.~Berenson, ``Data-efficient learning of natural language
  to linear temporal logic translators for robot task specification,'' in
  \emph{Proc. IEEE Int. Conf. Robot. Autom.}, 2023, pp. 11\,554--11\,561.

\bibitem{chen2023autotamp}
Y.~Chen, J.~Arkin, C.~Dawson, Y.~Zhang, N.~Roy, and C.~Fan, ``Autotamp:
  Autoregressive task and motion planning with llms as translators and
  checkers,'' in \emph{IEEE Int. Conf. Robot. Autom.}\hskip 1em plus 0.5em
  minus 0.4em\relax IEEE, 2024, pp. 6695--6702.

\bibitem{hoffmann2001ff}
J.~Hoffmann, ``Ff: The fast-forward planning system,'' \emph{AI Mag.}, vol.~22,
  no.~3, pp. 57--57, 2001.

\bibitem{baier2009heuristic}
J.~A. Baier, F.~Bacchus, and S.~A. McIlraith, ``A heuristic search approach to
  planning with temporally extended preferences,'' \emph{Artif. Intell.}, vol.
  173, no. 5-6, pp. 593--618, 2009.

\bibitem{zhu2005simultaneous}
L.~Zhu and R.~Givan, ``Simultaneous heuristic search for conjunctive
  subgoals,'' in \emph{Proc. Nat. Conf. Artif. Intell.}, vol.~3, 2005, pp.
  1235--1240.

\bibitem{helmert2006fast}
M.~Helmert, ``The fast downward planning system,'' \emph{J. Artif. Intell.
  Res.}, vol.~26, pp. 191--246, 2006.

\bibitem{richter2011lama}
S.~Richter, M.~Westphal, and M.~Helmert, ``Lama 2008 and 2011,'' in \emph{Int.
  Planning Compet.}, 2011, pp. 117--124.

\bibitem{georgievski2014overview}
I.~Georgievski and M.~Aiello, ``An overview of hierarchical task network
  planning,'' \emph{arXiv:1403.7426}, 2014.

\bibitem{hopcroft2001introduction}
J.~E. Hopcroft, R.~Motwani, and J.~D. Ullman, ``Introduction to automata
  theory, languages, and computation,'' \emph{ACM SIGACT News}, vol.~32, no.~1,
  pp. 60--65, 2001.

\bibitem{pnueli1989synthesis}
A.~Pnueli and R.~Rosner, ``On the synthesis of a reactive module,'' in
  \emph{Proc. ACM SIGPLAN-SIGACT Symp. Principles Program. Lang.}, 1989, pp.
  179--190.

\bibitem{maoz2015gr}
S.~Maoz and J.~O. Ringert, ``Gr (1) synthesis for ltl specification patterns,''
  in \emph{Proc. Joint Meet. Found. Softw. Eng.}, 2015, pp. 96--106.

\bibitem{ehlers2016slugs}
R.~Ehlers and V.~Raman, ``Slugs: Extensible gr (1) synthesis,'' in
  \emph{Computer Aided Verif.: 28th Int. Conf.}\hskip 1em plus 0.5em minus
  0.4em\relax Springer, 2016, pp. 333--339.

\bibitem{pasula2007learning}
H.~M. Pasula, L.~S. Zettlemoyer, and L.~P. Kaelbling, ``Learning symbolic
  models of stochastic domains,'' \emph{J. Artif. Intell. Res.}, vol.~29, pp.
  309--352, 2007.

\bibitem{amir2008learning}
E.~Amir and A.~Chang, ``Learning partially observable deterministic action
  models,'' \emph{J. Artif. Intell. Res.}, vol.~33, pp. 349--402, 2008.

\bibitem{konidaris2015symbol}
G.~Konidaris, L.~P. Kaelbling, and T.~Lozano-Perez, ``Symbol acquisition for
  probabilistic high-level planning,'' in \emph{Int. Joint Conf. Artif.
  Intell.}, 2015, pp. 3619--3627.

\bibitem{konidaris2018skills}
G.~Konidaris, L.~P. Kaelbling, and T.~Lozano-Perez, ``From skills to symbols:
  Learning symbolic representations for abstract high-level planning,''
  \emph{J. Artif. Intell. Res.}, vol.~61, pp. 215--289, 2018.

\bibitem{ames2018learning}
B.~Ames, A.~Thackston, and G.~Konidaris, ``Learning symbolic representations
  for planning with parameterized skills,'' in \emph{Proc. IEEE/RSJ Int. Conf.
  Intell. Robots Syst.}, 2018, pp. 526--533.

\bibitem{xu2018neural}
D.~Xu, S.~Nair, Y.~Zhu, J.~Gao, A.~Garg, L.~Fei-Fei, and S.~Savarese, ``Neural
  task programming: Learning to generalize across hierarchical tasks,'' in
  \emph{Proc. IEEE Int. Conf. Robot. Autom.}, 2018, pp. 3795--3802.

\bibitem{huang2019neural}
D.-A. Huang, S.~Nair, D.~Xu, Y.~Zhu, A.~Garg, L.~Fei-Fei, S.~Savarese, and
  J.~C. Niebles, ``Neural task graphs: Generalizing to unseen tasks from a
  single video demonstration,'' in \emph{Proc. IEEE/CVF Conf. Comput. Vision
  Pattern Recognit.}, 2019, pp. 8565--8574.

\bibitem{xu2019regression}
D.~Xu, R.~Mart{\'\i}n-Mart{\'\i}n, D.-A. Huang, Y.~Zhu, S.~Savarese, and L.~F.
  Fei-Fei, ``Regression planning networks,'' \emph{Adv. Neural Inf. Process.
  Syst.}, vol.~32, pp. 1319--1929, 2019.

\bibitem{ceola2019robot}
F.~Ceola, E.~Tosello, L.~Tagliapietra, G.~Nicola, and S.~Ghidoni, ``Robot task
  planning via deep reinforcement learning: a tabletop object sorting
  application,'' in \emph{Proc. IEEE Int. Conf. Syst., Man, Cybern.}, 2019, pp.
  486--492.

\bibitem{xu2021deep}
D.~Xu, A.~Mandlekar, R.~Mart{\'\i}n-Mart{\'\i}n, Y.~Zhu, S.~Savarese, and
  L.~Fei-Fei, ``Deep affordance foresight: Planning through what can be done in
  the future,'' in \emph{Proc. IEEE Int. Conf. Robot. Autom.}, 2021, pp.
  6206--6213.

\bibitem{liang2022search}
J.~Liang, M.~Sharma, A.~LaGrassa, S.~Vats, S.~Saxena, and O.~Kroemer,
  ``Search-based task planning with learned skill effect models for lifelong
  robotic manipulation,'' in \emph{Proc. IEEE Int. Conf. Robot. Autom.}, 2022,
  pp. 6351--6357.

\bibitem{openai}
\BIBentryALTinterwordspacing
OpenAI, ``Chatgpt,'' Accessed: 2023-02-08, 2023, cit. on pp. 1, 16. [Online].
  Available: \url{https://openai.com/blog/chatgpt/}
\BIBentrySTDinterwordspacing

\bibitem{touvron2023llama}
H.~Touvron, T.~Lavril, G.~Izacard, X.~Martinet, M.-A. Lachaux, T.~Lacroix,
  B.~Rozi{\`e}re, N.~Goyal, E.~Hambro, F.~Azhar \emph{et~al.}, ``Llama: Open
  and efficient foundation language models,'' \emph{arXiv:2302.13971}, 2023.

\bibitem{liu2023pre}
P.~Liu, W.~Yuan, J.~Fu, Z.~Jiang, H.~Hayashi, and G.~Neubig, ``Pre-train,
  prompt, and predict: A systematic survey of prompting methods in natural
  language processing,'' \emph{ACM Comput. Surv.}, vol.~55, no.~9, pp. 1--35,
  2023.

\bibitem{imani2023mathprompter}
S.~Imani, L.~Du, and H.~Shrivastava, ``Mathprompter: Mathematical reasoning
  using large language models,'' in \emph{Proc. Annu. Meet. Assoc. Comput.
  Linguist.}, vol.~5, 2023, pp. 37--42.

\bibitem{gaur2023reasoning}
V.~Gaur and N.~Saunshi, ``Reasoning in large language models through symbolic
  math word problems,'' in \emph{Findings Assoc. Comput. Linguist.: ACL 2023},
  pp. 5889--5903.

\bibitem{ding2023task}
Y.~Ding, X.~Zhang, C.~Paxton, and S.~Zhang, ``Task and motion planning with
  large language models for object rearrangement,'' in \emph{Proc. IEEE/RSJ
  Int. Conf. Intell. Robots Syst.}, 2023, pp. 2086--2092.

\bibitem{brohan2023can}
A.~Brohan, Y.~Chebotar, C.~Finn, K.~Hausman, A.~Herzog, D.~Ho, J.~Ibarz,
  A.~Irpan, E.~Jang, R.~Julian \emph{et~al.}, ``Do as i can, not as i say:
  Grounding language in robotic affordances,'' in \emph{Proc. Conf. Robot.
  Learn.}\hskip 1em plus 0.5em minus 0.4em\relax PMLR, 2023, pp. 287--318.

\bibitem{huang2023inner}
W.~Huang, F.~Xia, T.~Xiao, H.~Chan, J.~Liang, P.~Florence, A.~Zeng, J.~Tompson,
  I.~Mordatch, Y.~Chebotar \emph{et~al.}, ``Inner monologue: Embodied reasoning
  through planning with language models,'' in \emph{Proc. Conf. Robot.
  Learn.}\hskip 1em plus 0.5em minus 0.4em\relax PMLR, 2023, pp. 1769--1782.

\bibitem{hu2023toward}
Y.~Hu, Q.~Xie, V.~Jain, J.~Francis, J.~Patrikar, N.~Keetha, S.~Kim, Y.~Xie,
  T.~Zhang, Z.~Zhao \emph{et~al.}, ``Toward general-purpose robots via
  foundation models: A survey and meta-analysis,'' \emph{arXiv:2312.08782},
  2023.

\bibitem{betts1998survey}
J.~T. Betts, ``Survey of numerical methods for trajectory optimization,''
  \emph{J. Guid. Control Dyn.}, vol.~21, no.~2, pp. 193--207, 1998.

\bibitem{boyd2011distributed}
S.~Boyd, N.~Parikh, E.~Chu, B.~Peleato, J.~Eckstein \emph{et~al.},
  ``Distributed optimization and statistical learning via the alternating
  direction method of multipliers,'' \emph{Found. Trends Mach. Learn.}, vol.~3,
  no.~1, pp. 1--122, 2011.

\bibitem{mordatch2014combining}
I.~Mordatch and E.~Todorov, ``Combining the benefits of function approximation
  and trajectory optimization.'' in \emph{Proc. Robot. Sci. Syst.}, vol.~4,
  2014, p.~23.

\bibitem{janner2021offline}
M.~Janner, Q.~Li, and S.~Levine, ``Offline reinforcement learning as one big
  sequence modeling problem,'' \emph{Adv. Neural Inf. Process. Syst.}, vol.~34,
  pp. 1273--1286, 2021.

\bibitem{kelly2017introduction}
M.~Kelly, ``An introduction to trajectory optimization: How to do your own
  direct collocation,'' \emph{SIAM Review}, vol.~59, no.~4, pp. 849--904, 2017.

\bibitem{pardo2016evaluating}
D.~Pardo, L.~M{\"o}ller, M.~Neunert, A.~W. Winkler, and J.~Buchli, ``Evaluating
  direct transcription and nonlinear optimization methods for robot motion
  planning,'' \emph{IEEE Robot. Autom. Lett.}, vol.~1, no.~2, pp. 946--953,
  2016.

\bibitem{wachter2006implementation}
A.~W{\"a}chter and L.~T. Biegler, ``On the implementation of an interior-point
  filter line-search algorithm for large-scale nonlinear programming,''
  \emph{Math. Program.}, vol. 106, pp. 25--57, 2006.

\bibitem{gill2005snopt}
P.~E. Gill, W.~Murray, and M.~A. Saunders, ``Snopt: An sqp algorithm for
  large-scale constrained optimization,'' \emph{SIAM review}, vol.~47, no.~1,
  pp. 99--131, 2005.

\bibitem{mayne1966second}
D.~Mayne, ``A second-order gradient method for determining optimal trajectories
  of non-linear discrete-time systems,'' \emph{Int. J. Control}, vol.~3, no.~1,
  pp. 85--95, 1966.

\bibitem{xie2017differential}
Z.~Xie, C.~K. Liu, and K.~Hauser, ``Differential dynamic programming with
  nonlinear constraints,'' in \emph{Proc. IEEE Int. Conf. Robot. Autom.}, 2017,
  pp. 695--702.

\bibitem{plancher2017constrained}
B.~Plancher, Z.~Manchester, and S.~Kuindersma, ``Constrained unscented dynamic
  programming,'' in \emph{Proc. IEEE/RSJ Int. Conf. Intell. Robots Syst.},
  2017, pp. 5674--5680.

\bibitem{howell2019altro}
T.~A. Howell, B.~E. Jackson, and Z.~Manchester, ``Altro: A fast solver for
  constrained trajectory optimization,'' in \emph{Proc. IEEE/RSJ Int. Conf.
  Intell. Robots Syst.}, 2019, pp. 7674--7679.

\bibitem{sleiman2021constraint}
J.-P. Sleiman, F.~Farshidian, and M.~Hutter, ``Constraint handling in
  continuous-time ddp-based model predictive control,'' in \emph{Proc. IEEE
  Int. Conf. Robot. Autom.}, 2021, pp. 8209--8215.

\bibitem{jallet2022constrained}
W.~Jallet, A.~Bambade, N.~Mansard, and J.~Carpentier, ``Constrained
  differential dynamic programming: A primal-dual augmented lagrangian
  approach,'' in \emph{Proc. IEEE/RSJ Int. Conf. Intell. Robots Syst.}, 2022,
  pp. 13\,371--13\,378.

\bibitem{wang2023fast}
Y.~Wang, H.~Li, Y.~Zhao, X.~Chen, X.~Huang, and Z.~Jiang, ``A fast coordinated
  motion planning method for dual-arm robot based on parallel constrained
  ddp,'' \emph{IEEE/ASME Trans. Mechatronics}, 2023.

\bibitem{nocedal1999numerical}
J.~Nocedal and S.~J. Wright, \emph{Numerical optimization}.\hskip 1em plus
  0.5em minus 0.4em\relax Springer, 1999.

\bibitem{featherstone2014rigid}
R.~Featherstone, \emph{Rigid body dynamics algorithms}.\hskip 1em plus 0.5em
  minus 0.4em\relax Springer, 2014.

\bibitem{eckstein1992douglas}
J.~Eckstein and D.~P. Bertsekas, ``On the douglas—rachford splitting method
  and the proximal point algorithm for maximal monotone operators,''
  \emph{Math. Program.}, vol.~55, pp. 293--318, 1992.

\bibitem{wohlberg2017admm}
B.~Wohlberg, ``Admm penalty parameter selection by residual balancing,''
  \emph{arXiv:1704.06209}, 2017.

\bibitem{goldstein2014fast}
T.~Goldstein, B.~O'Donoghue, S.~Setzer, and R.~Baraniuk, ``Fast alternating
  direction optimization methods,'' \emph{SIAM J. Imaging Sci.}, vol.~7, no.~3,
  pp. 1588--1623, 2014.

\bibitem{ferranti2022distributed}
L.~Ferranti, L.~Lyons, R.~R. Negenborn, T.~Keviczky, and J.~Alonso-Mora,
  ``Distributed nonlinear trajectory optimization for multi-robot motion
  planning,'' \emph{IEEE Trans. Control Syst. Technol.}, 2022.

\bibitem{shorinwa2023distributed2}
O.~Shorinwa, T.~Halsted, J.~Yu, and M.~Schwager, ``Distributed optimization
  methods for multi-robot systems: Part 2—a survey,'' \emph{IEEE Robot.
  Autom. Mag.}, vol.~31, pp. 154--169, 2024.

\bibitem{ni2022robust}
R.~Ni, Z.~Pan, and X.~Gao, ``Robust multi-robot trajectory optimization using
  alternating direction method of multiplier,'' \emph{IEEE Robot. Autom.
  Lett.}, vol.~7, no.~3, pp. 5950--5957, 2022.

\bibitem{amatucci2024accelerating}
L.~Amatucci, G.~Turrisi, A.~Bratta, V.~Barasuol, and C.~Semini, ``Accelerating
  model predictive control for legged robots through distributed
  optimization,'' \emph{arXiv:2403.11742}, 2024.

\bibitem{aydinoglu2022real}
A.~Aydinoglu and M.~Posa, ``Real-time multi-contact model predictive control
  via admm,'' in \emph{Proc. IEEE Int. Conf. Robot. Autom.}, 2022, pp.
  3414--3421.

\bibitem{le2019fast}
S.~Le~Cleac’h and Z.~Manchester, ``Fast solution of optimal control problems
  with l1 cost,'' in \emph{AAS/AIAA Astrodyn. Spec. Conf.}, vol. 904, 2019, pp.
  1--11.

\bibitem{zhao2021sydebo}
Z.~Zhao, Z.~Zhou, M.~Park, and Y.~Zhao, ``Sydebo: Symbolic-decision-embedded
  bilevel optimization for long-horizon manipulation in dynamic environments,''
  \emph{IEEE Access}, vol.~9, pp. 128\,817--128\,826, 2021.

\bibitem{wijayarathne2022real}
L.~Wijayarathne, Z.~Zhou, Y.~Zhao, and F.~L. Hammond, ``Real-time
  deformable-contact-aware model predictive control for force-modulated
  manipulation,'' \emph{IEEE Trans. Robot.}, pp. 1--18, 2023.

\bibitem{li2021model}
H.~Li, R.~J. Frei, and P.~M. Wensing, ``Model hierarchy predictive control of
  robotic systems,'' \emph{IEEE Robot. Autom. Lett.}, vol.~6, no.~2, pp.
  3373--3380, 2021.

\bibitem{khazoom2023optimal}
C.~Khazoom, S.~Heim, D.~Gonzalez-Diaz, and S.~Kim, ``Optimal scheduling of
  models and horizons for model hierarchy predictive control,'' in \emph{Proc.
  IEEE Int. Conf. Robot. Autom.}, 2023, pp. 9952--9958.

\bibitem{herzog2016structured}
A.~Herzog, S.~Schaal, and L.~Righetti, ``Structured contact force optimization
  for kino-dynamic motion generation,'' in \emph{Proc. IEEE/RSJ Int. Conf.
  Intell. Robots Syst.}, 2016, pp. 2703--2710.

\bibitem{zhou2022momentum}
Z.~Zhou, B.~Wingo, N.~Boyd, S.~Hutchinson, and Y.~Zhao, ``Momentum-aware
  trajectory optimization and control for agile quadrupedal locomotion,''
  \emph{IEEE Robot. Autom. Lett.}, vol.~7, no.~3, pp. 7755--7762, 2022.

\bibitem{budhiraja2019dynamics}
R.~Budhiraja, J.~Carpentier, and N.~Mansard, ``Dynamics consensus between
  centroidal and whole-body models for locomotion of legged robots,'' in
  \emph{Proc. IEEE Int. Conf. Robot. Autom.}, 2019, pp. 6727--6733.

\bibitem{meduri2023biconmp}
A.~Meduri, P.~Shah, J.~Viereck, M.~Khadiv, I.~Havoutis, and L.~Righetti,
  ``Biconmp: A nonlinear model predictive control framework for whole body
  motion planning,'' \emph{IEEE Trans. Robot.}, 2023.

\bibitem{zhou2020accelerated}
Z.~Zhou and Y.~Zhao, ``Accelerated admm based trajectory optimization for
  legged locomotion with coupled rigid body dynamics,'' in \emph{Amer. Control
  Conf.}\hskip 1em plus 0.5em minus 0.4em\relax IEEE, 2020, pp. 5082--5089.

\bibitem{finn2016guided}
C.~Finn, S.~Levine, and P.~Abbeel, ``Guided cost learning: Deep inverse optimal
  control via policy optimization,'' in \emph{Int. Conf. Mach. Learn.}\hskip
  1em plus 0.5em minus 0.4em\relax PMLR, 2016, pp. 49--58.

\bibitem{englert2017inverse}
P.~Englert, N.~A. Vien, and M.~Toussaint, ``Inverse kkt: Learning cost
  functions of manipulation tasks from demonstrations,'' \emph{Int. J. Robot.
  Res.}, vol.~36, no. 13-14, pp. 1474--1488, 2017.

\bibitem{sharma2022correcting}
P.~Sharma, B.~Sundaralingam, V.~Blukis, C.~Paxton, T.~Hermans, A.~Torralba,
  J.~Andreas, and D.~Fox, ``{Correcting Robot Plans with Natural Language
  Feedback},'' in \emph{Proc. Robot. Sci. Syst.}, 2022.

\bibitem{yu2023language}
W.~Yu, N.~Gileadi, C.~Fu, S.~Kirmani, K.-H. Lee, M.~G. Arenas, H.-T.~L. Chiang,
  T.~Erez, L.~Hasenclever, J.~Humplik \emph{et~al.}, ``Language to rewards for
  robotic skill synthesis,'' in \emph{Proc. Conf. Robot. Learn.}\hskip 1em plus
  0.5em minus 0.4em\relax PMLR, 2023, pp. 374--404.

\bibitem{huang2023voxposer}
W.~Huang, C.~Wang, R.~Zhang, Y.~Li, J.~Wu, and L.~Fei-Fei, ``Voxposer:
  Composable 3d value maps for robotic manipulation with language models,'' in
  \emph{Proc. Conf. Robot. Learn.}\hskip 1em plus 0.5em minus 0.4em\relax PMLR,
  2023, pp. 540--562.

\bibitem{parmar2021fundamental}
M.~Parmar, M.~Halm, and M.~Posa, ``Fundamental challenges in deep learning for
  stiff contact dynamics,'' in \emph{Proc. IEEE/RSJ Int. Conf. Intell. Robots
  Syst.}, 2021, pp. 5181--5188.

\bibitem{pfrommer2021contactnets}
S.~Pfrommer, M.~Halm, and M.~Posa, ``Contactnets: Learning discontinuous
  contact dynamics with smooth, implicit representations,'' in \emph{Proc.
  Conf. Robot. Learn.}\hskip 1em plus 0.5em minus 0.4em\relax PMLR, 2021, pp.
  2279--2291.

\bibitem{bianchini2023simultaneous}
B.~Bianchini, M.~Halm, and M.~Posa, ``Simultaneous learning of contact and
  continuous dynamics,'' in \emph{Proc. Conf. Robot. Learn.}\hskip 1em plus
  0.5em minus 0.4em\relax PMLR, 2023, pp. 3966--3978.

\bibitem{le2023differentiable}
S.~Le~Cleac'h, H.-X. Yu, M.~Guo, T.~Howell, R.~Gao, J.~Wu, Z.~Manchester, and
  M.~Schwager, ``Differentiable physics simulation of dynamics-augmented neural
  objects,'' \emph{IEEE Robot. Autom. Lett.}, vol.~8, no.~5, pp. 2780--2787,
  2023.

\bibitem{driess2022learning}
D.~Driess, J.-S. Ha, M.~Toussaint, and R.~Tedrake, ``Learning models as
  functionals of signed-distance fields for manipulation planning,'' in
  \emph{Proc. Conf. Robot. Learn.}\hskip 1em plus 0.5em minus 0.4em\relax PMLR,
  2022, pp. 245--255.

\bibitem{levine2013guided}
S.~Levine and V.~Koltun, ``Guided policy search,'' in \emph{Int. Conf. Mach.
  Learn.}\hskip 1em plus 0.5em minus 0.4em\relax PMLR, 2013, pp. 1--9.

\bibitem{levine2014learning}
S.~Levine and V.~Koltun, ``Learning complex neural network policies with
  trajectory optimization,'' in \emph{Int. Conf. Mach. Learn.}\hskip 1em plus
  0.5em minus 0.4em\relax PMLR, 2014, pp. 829--837.

\bibitem{duburcq2020online}
A.~Duburcq, Y.~Chevaleyre, N.~Bredeche, and G.~Bo{\'e}ris, ``Online trajectory
  planning through combined trajectory optimization and function approximation:
  Application to the exoskeleton atalante,'' in \emph{Proc. IEEE Int. Conf.
  Robot. Autom.}, 2020, pp. 3756--3762.

\bibitem{zhao2022adversarially}
Z.~Zhao, S.~Zuo, T.~Zhao, and Y.~Zhao, ``Adversarially regularized policy
  learning guided by trajectory optimization,'' in \emph{Learn. Dyn. Control
  Conf.}\hskip 1em plus 0.5em minus 0.4em\relax PMLR, 2022, pp. 844--857.

\bibitem{bency2019neural}
M.~J. Bency, A.~H. Qureshi, and M.~C. Yip, ``Neural path planning: Fixed time,
  near-optimal path generation via oracle imitation,'' in \emph{Proc. IEEE/RSJ
  Int. Conf. Intell. Robots Syst.}, 2019, pp. 3965--3972.

\bibitem{qureshi2020neural}
A.~H. Qureshi, J.~Dong, A.~Choe, and M.~C. Yip, ``Neural manipulation planning
  on constraint manifolds,'' \emph{IEEE Robot. Autom. Lett.}, vol.~5, no.~4,
  pp. 6089--6096, 2020.

\bibitem{radosavovic2023learning}
I.~Radosavovic, T.~Xiao, B.~Zhang, T.~Darrell, J.~Malik, and K.~Sreenath,
  ``Real-world humanoid locomotion with reinforcement learning,'' \emph{Science
  Robotics}, vol.~9, no.~89, p. eadi9579, 2024.

\bibitem{chi2023diffusionpolicy}
C.~Chi, S.~Feng, Y.~Du, Z.~Xu, E.~Cousineau, B.~Burchfiel, and S.~Song,
  ``Diffusion policy: Visuomotor policy learning via action diffusion,'' in
  \emph{Proc. Robot. Sci. Syst.}, 2023.

\bibitem{viereck2021learning}
J.~Viereck and L.~Righetti, ``Learning a centroidal motion planner for legged
  locomotion,'' in \emph{Proc. IEEE Int. Conf. Robot. Autom.}, 2021, pp.
  4905--4911.

\bibitem{lin2022multi}
X.~Lin, G.~I. Fernandez, Y.~Liu, T.~Zhu, Y.~Shirai, and D.~Hong, ``Multi-modal
  multi-agent optimization for limms, a modular robotics approach to delivery
  automation,'' in \emph{Proc. IEEE/RSJ Int. Conf. Intell. Robots Syst.}, 2022,
  pp. 12\,674--12\,681.

\bibitem{gu2023walking}
Z.~Gu, R.~Guo, W.~Yates, Y.~Chen, Y.~Zhao, and Y.~Zhao, ``Walking-by-logic:
  Signal temporal logic-guided model predictive control for bipedal locomotion
  resilient to external perturbations,'' in \emph{Proc. IEEE Int. Conf. Robot.
  Autom.}, 2024, pp. 1121--1127.

\bibitem{toussaint2017multi}
M.~Toussaint and M.~Lopes, ``Multi-bound tree search for logic-geometric
  programming in cooperative manipulation domains,'' in \emph{Proc. IEEE Int.
  Conf. Robot. Autom.}, 2017, pp. 4044--4051.

\bibitem{lo2020petlon}
S.-Y. Lo, S.~Zhang, and P.~Stone, ``The petlon algorithm to plan efficiently
  for task-level-optimal navigation,'' \emph{J. Artif. Intell. Res.}, vol.~69,
  pp. 471--500, 2020.

\bibitem{wolff2014optimization}
E.~M. Wolff, U.~Topcu, and R.~M. Murray, ``Optimization-based trajectory
  generation with linear temporal logic specifications,'' in \emph{Proc. IEEE
  Int. Conf. Robot. Autom.}, 2014, pp. 5319--5325.

\bibitem{chen2021optimal}
J.~Chen, B.~C. Williams, and C.~Fan, ``Optimal mixed discrete-continuous
  planning for linear hybrid systems,'' in \emph{Proc. Int. Conf. Hybrid Syst.:
  Comput. Control}, 2021, pp. 1--12.

\bibitem{kogo2021fast}
T.~Kogo, K.~Takaya, and H.~Oyama, ``Fast milp-based task and motion planning
  for pick-and-place with hard/soft constraints of collision-free route,'' in
  \emph{Proc. IEEE Int. Conf. Syst., Man, Cybern.}\hskip 1em plus 0.5em minus
  0.4em\relax IEEE, 2021, pp. 1020--1027.

\bibitem{katayama2020fast}
M.~Katayama, S.~Tokuda, M.~Yamakita, and H.~Oyama, ``Fast ltl-based flexible
  planning for dual-arm manipulation,'' in \emph{Proc. IEEE/RSJ Int. Conf.
  Intell. Robots Syst.}, 2020, pp. 6605--6612.

\bibitem{saha2017task}
S.~Saha and A.~A. Julius, ``Task and motion planning for manipulator arms with
  metric temporal logic specifications,'' \emph{IEEE Robot. Autom. Lett.},
  vol.~3, no.~1, pp. 379--386, 2017.

\bibitem{pant2018fly}
Y.~V. Pant, H.~Abbas, R.~A. Quaye, and R.~Mangharam, ``Fly-by-logic: Control of
  multi-drone fleets with temporal logic objectives,'' in \emph{Proc. ACM/IEEE
  Int. Conf. Cyber-Phys. Syst.}, 2018, pp. 186--197.

\bibitem{bacchus1994downward}
F.~Bacchus and Q.~Yang, ``Downward refinement and the efficiency of
  hierarchical problem solving,'' \emph{Artif. Intell.}, vol.~71, no.~1, pp.
  43--100, 1994.

\bibitem{akbari2019combined}
A.~Akbari, F.~Lagriffoul, and J.~Rosell, ``Combined heuristic task and motion
  planning for bi-manual robots,'' \emph{Auton. Robots}, vol.~43, no.~6, pp.
  1575--1590, 2019.

\bibitem{agostini2023unified}
A.~Agostini and J.~Piater, ``Unified task and motion planning using
  object-centric abstractions of motion constraints,'' \emph{arXiv:2312.17605},
  2023.

\bibitem{srivastava2014combined}
S.~Srivastava, E.~Fang, L.~Riano, R.~Chitnis, S.~Russell, and P.~Abbeel,
  ``Combined task and motion planning through an extensible planner-independent
  interface layer,'' in \emph{Proc. IEEE Int. Conf. Robot. Autom.}, 2014, pp.
  639--646.

\bibitem{hauser2010multi}
K.~Hauser and J.-C. Latombe, ``Multi-modal motion planning in non-expansive
  spaces,'' \emph{Int. J. Robot. Res.}, vol.~29, no.~7, pp. 897--915, 2010.

\bibitem{kingston2022scaling}
Z.~Kingston and L.~E. Kavraki, ``Scaling multimodal planning: Using experience
  and informing discrete search,'' \emph{IEEE Trans. Robot.}, vol.~39, no.~1,
  pp. 128--146, 2022.

\bibitem{toussaint2017tutorial}
M.~Toussaint, ``A tutorial on newton methods for constrained trajectory
  optimization and relations to slam, gaussian process smoothing, optimal
  control, and probabilistic inference,'' \emph{Geom. Numer. Found. Mov.}, pp.
  361--392, 2017.

\bibitem{toussaint2020describing}
M.~Toussaint, J.-S. Ha, and D.~Driess, ``Describing physics for physical
  reasoning: Force-based sequential manipulation planning,'' \emph{IEEE Robot.
  Autom. Lett.}, vol.~5, no.~4, pp. 6209--6216, 2020.

\bibitem{zimmermann2020multi}
S.~Zimmermann, G.~Hakimifard, M.~Zamora, R.~Poranne, and S.~Coros, ``A
  multi-level optimization framework for simultaneous grasping and motion
  planning,'' \emph{IEEE Robot. Autom. Lett.}, vol.~5, no.~2, pp. 2966--2972,
  2020.

\bibitem{phoon2022constraint}
M.~S. Phoon, P.~S. Schmitt, and G.~v. Wichert, ``Constraint-based task
  specification and trajectory optimization for sequential manipulation,'' in
  \emph{Proc. IEEE/RSJ Int. Conf. Intell. Robots Syst.}, 2022, pp. 197--202.

\bibitem{schwenzer2021review}
M.~Schwenzer, M.~Ay, T.~Bergs, and D.~Abel, ``Review on model predictive
  control: An engineering perspective,'' \emph{Int. J. Adv. Manuf. Technol.},
  vol. 117, no.~5, pp. 1327--1349, 2021.

\bibitem{hartmann2020robust}
V.~N. Hartmann, O.~S. Oguz, D.~Driess, M.~Toussaint, and A.~Menges, ``Robust
  task and motion planning for long-horizon architectural construction
  planning,'' in \emph{Proc. IEEE/RSJ Int. Conf. Intell. Robots Syst.}, 2020,
  pp. 6886--6893.

\bibitem{castaman2021receding}
N.~Castaman, E.~Pagello, E.~Menegatti, and A.~Pretto, ``Receding horizon task
  and motion planning in changing environments,'' \emph{Robot. Auton. Syst.},
  vol. 145, p. 103863, 2021.

\bibitem{braun2022rhh}
C.~V. Braun, J.~Ortiz-Haro, M.~Toussaint, and O.~S. Oguz, ``Rhh-lgp: Receding
  horizon and heuristics-based logic-geometric programming for task and motion
  planning,'' in \emph{Proc. IEEE/RSJ Int. Conf. Intell. Robots Syst.}\hskip
  1em plus 0.5em minus 0.4em\relax IEEE, 2022, pp. 13\,761--13\,768.

\bibitem{chen2022interactive}
Y.~Chen, U.~Rosolia, W.~Ubellacker, N.~Csomay-Shanklin, and A.~D. Ames,
  ``Interactive multi-modal motion planning with branch model predictive
  control,'' \emph{IEEE Robot. Autom. Lett.}, vol.~7, no.~2, pp. 5365--5372,
  2022.

\bibitem{lawler1966branch}
E.~L. Lawler and D.~E. Wood, ``Branch-and-bound methods: A survey,''
  \emph{Oper. Res.}, vol.~14, no.~4, pp. 699--719, 1966.

\bibitem{huang2021branch}
L.~Huang, X.~Chen, W.~Huo, J.~Wang, F.~Zhang, B.~Bai, and L.~Shi, ``Branch and
  bound in mixed integer linear programming problems: A survey of techniques
  and trends,'' \emph{arXiv:2111.06257}, 2021.

\bibitem{schrijver2003combinatorial}
A.~Schrijver \emph{et~al.}, \emph{Combinatorial optimization: polyhedra and
  efficiency}.\hskip 1em plus 0.5em minus 0.4em\relax Springer, 2003, vol.~24,
  no.~2.

\bibitem{berthold2006primal}
T.~Berthold, ``Primal heuristics for mixed integer programs,'' Ph.D.
  dissertation, Zuse Institute Berlin (ZIB), 2006.

\bibitem{dey2023theoretical}
S.~S. Dey, Y.~Dubey, M.~Molinaro, and P.~Shah, ``A theoretical and
  computational analysis of full strong-branching,'' \emph{Math. Program.}, pp.
  1--34, 2023.

\bibitem{fischetti2003local}
M.~Fischetti and A.~Lodi, ``Local branching,'' \emph{Math. Program.}, vol.~98,
  pp. 23--47, 2003.

\bibitem{gurobi}
\BIBentryALTinterwordspacing
{Gurobi Optimization, LLC}, ``{Gurobi Optimizer Reference Manual},'' 2023.
  [Online]. Available: \url{https://www.gurobi.com}
\BIBentrySTDinterwordspacing

\bibitem{mosek}
\BIBentryALTinterwordspacing
M.~ApS, \emph{The MOSEK optimization toolbox for MATLAB manual. Version 9.0.},
  2019. [Online]. Available: \url{http://docs.mosek.com/9.0/toolbox/index.html}
\BIBentrySTDinterwordspacing

\bibitem{MATLAB}
\BIBentryALTinterwordspacing
T.~M. Inc., ``Matlab version: 9.13.0 (r2022b),'' Natick, Massachusetts, United
  States, 2022. [Online]. Available: \url{https://www.mathworks.com}
\BIBentrySTDinterwordspacing

\bibitem{funk2022graph}
N.~Funk, S.~Menzenbach, G.~Chalvatzaki, and J.~Peters, ``Graph-based
  reinforcement learning meets mixed integer programs: An application to 3d
  robot assembly discovery,'' in \emph{Proc. IEEE/RSJ Int. Conf. Intell. Robots
  Syst.}, 2022, pp. 10\,215--10\,222.

\bibitem{shamsah2023integrated}
A.~Shamsah, Z.~Gu, J.~Warnke, S.~Hutchinson, and Y.~Zhao, ``Integrated task and
  motion planning for safe legged navigation in partially observable
  environments,'' \emph{IEEE Trans. Robot.}, 2023.

\bibitem{warnke2020towards}
J.~Warnke, A.~Shamsah, Y.~Li, and Y.~Zhao, ``Towards safe locomotion navigation
  in partially observable environments with uneven terrain,'' in \emph{Proc.
  IEEE Conf. Decis. Control}, 2020, pp. 958--965.

\bibitem{shirai2022simultaneous}
Y.~Shirai, X.~Lin, A.~Schperberg, Y.~Tanaka, H.~Kato, V.~Vichathorn, and
  D.~Hong, ``Simultaneous contact-rich grasping and locomotion via distributed
  optimization enabling free-climbing for multi-limbed robots,'' in \emph{Proc.
  IEEE/RSJ Int. Conf. Intell. Robots Syst.}, 2022, pp. 13\,563--13\,570.

\bibitem{pant2017smooth}
Y.~V. Pant, H.~Abbas, and R.~Mangharam, ``Smooth operator: Control using the
  smooth robustness of temporal logic,'' in \emph{Proc. IEEE Conf. Control
  Technol. Appl.}, 2017, pp. 1235--1240.

\bibitem{mehdipour2019arithmetic}
N.~Mehdipour, C.-I. Vasile, and C.~Belta, ``Arithmetic-geometric mean
  robustness for control from signal temporal logic specifications,'' in
  \emph{Amer. Control Conf.}\hskip 1em plus 0.5em minus 0.4em\relax IEEE, 2019,
  pp. 1690--1695.

\bibitem{gilpin2020smooth}
Y.~Gilpin, V.~Kurtz, and H.~Lin, ``A smooth robustness measure of signal
  temporal logic for symbolic control,'' \emph{IEEE Control Syst. Lett.},
  vol.~5, no.~1, pp. 241--246, 2020.

\bibitem{gu2024robust}
Z.~Gu, Y.~Zhao, Y.~Chen, R.~Guo, J.~K. Leestma, G.~S. Sawicki, and Y.~Zhao,
  ``Robust-locomotion-by-logic: Perturbation-resilient bipedal locomotion via
  signal temporal logic guided model predictive control,''
  \emph{arXiv:2403.15993}, 2024.

\bibitem{sun2022multi}
D.~Sun, J.~Chen, S.~Mitra, and C.~Fan, ``Multi-agent motion planning from
  signal temporal logic specifications,'' \emph{IEEE Robot. Autom. Lett.},
  vol.~7, no.~2, pp. 3451--3458, 2022.

\bibitem{farahani2015robust}
S.~S. Farahani, V.~Raman, and R.~M. Murray, ``Robust model predictive control
  for signal temporal logic synthesis,'' \emph{Int. Fed. Autom. Control},
  vol.~48, no.~27, pp. 323--328, 2015.

\bibitem{sadraddini2015robust}
S.~Sadraddini and C.~Belta, ``Robust temporal logic model predictive control,''
  in \emph{Annu. Allerton Conf. Commun., Control, Comput.}\hskip 1em plus 0.5em
  minus 0.4em\relax IEEE, 2015, pp. 772--779.

\bibitem{sadigh2016safe}
D.~Sadigh and A.~Kapoor, ``Safe control under uncertainty with probabilistic
  signal temporal logic,'' in \emph{Proc. Robot. Sci. Syst.}, 2016.

\bibitem{kaelbling2011hierarchical}
L.~P. Kaelbling and T.~Lozano-P{\'e}rez, ``Hierarchical task and motion
  planning in the now,'' in \emph{Proc. IEEE Int. Conf. Robot. Autom.}, 2011,
  pp. 1470--1477.

\bibitem{wells2019learning}
A.~M. Wells, N.~T. Dantam, A.~Shrivastava, and L.~E. Kavraki, ``Learning
  feasibility for task and motion planning in tabletop environments,''
  \emph{IEEE Robot. Autom. Lett.}, vol.~4, no.~2, pp. 1255--1262, 2019.

\bibitem{noseworthy2021active}
M.~Noseworthy, C.~Moses, I.~Brand, S.~Castro, L.~Kaelbling,
  T.~Lozano-P{\'e}rez, and N.~Roy, ``Active learning of abstract plan
  feasibility,'' in \emph{Proc. Robot. Sci. Syst.}, 2021.

\bibitem{mandlekar2023hitltamp}
A.~Mandlekar, C.~Garrett, D.~Xu, and D.~Fox, ``Human-in-the-loop task and
  motion planning for imitation learning,'' in \emph{Proc. Conf. Robot.
  Learn.}\hskip 1em plus 0.5em minus 0.4em\relax PMLR, 2023, pp. 3030--3060.

\bibitem{silver2022learning}
T.~Silver, A.~Athalye, J.~B. Tenenbaum, T.~Lozano-Perez, and L.~P. Kaelbling,
  ``Learning neuro-symbolic skills for bilevel planning,'' in \emph{Proc. Conf.
  Robot. Learn.}\hskip 1em plus 0.5em minus 0.4em\relax PMLR, 2022, pp.
  701--714.

\bibitem{cauligi2020learning}
A.~Cauligi, P.~Culbertson, B.~Stellato, D.~Bertsimas, M.~Schwager, and
  M.~Pavone, ``Learning mixed-integer convex optimization strategies for robot
  planning and control,'' in \emph{Proc. IEEE Conf. Decis. Control}.\hskip 1em
  plus 0.5em minus 0.4em\relax IEEE, 2020, pp. 1698--1705.

\bibitem{sung2023learning}
Y.~Sung, Z.~Wang, and P.~Stone, ``Learning to correct mistakes: Backjumping in
  long-horizon task and motion planning,'' in \emph{Proc. Conf. Robot.
  Learn.}\hskip 1em plus 0.5em minus 0.4em\relax PMLR, 2023, pp. 2115--2124.

\bibitem{curtis2022long}
A.~Curtis, X.~Fang, L.~P. Kaelbling, T.~Lozano-P{\'e}rez, and C.~R. Garrett,
  ``Long-horizon manipulation of unknown objects via task and motion planning
  with estimated affordances,'' in \emph{Proc. IEEE Int. Conf. Robot. Autom.},
  2022, pp. 1940--1946.

\bibitem{wang2018active}
Z.~Wang, C.~R. Garrett, L.~P. Kaelbling, and T.~Lozano-P{\'e}rez, ``Active
  model learning and diverse action sampling for task and motion planning,'' in
  \emph{Proc. IEEE/RSJ Int. Conf. Intell. Robots Syst.}, 2018, pp. 4107--4114.

\bibitem{wang2021learning}
Z.~Wang, C.~R. Garrett, L.~P. Kaelbling, and T.~Lozano-P{\'e}rez, ``Learning
  compositional models of robot skills for task and motion planning,''
  \emph{Int. J. Robot. Res.}, vol.~40, no. 6-7, pp. 866--894, 2021.

\bibitem{kim2018guiding}
B.~Kim, L.~Kaelbling, and T.~Lozano-P{\'e}rez, ``Guiding search in continuous
  state-action spaces by learning an action sampler from off-target search
  experience,'' in \emph{Proc. AAAI Conf. Artif. Intell.}, vol.~32, no.~1,
  2018, pp. 6509--6516.

\bibitem{kim2019learning}
B.~Kim, Z.~Wang, L.~P. Kaelbling, and T.~Lozano-P{\'e}rez, ``Learning to guide
  task and motion planning using score-space representation,'' \emph{Int. J.
  Robot. Res.}, vol.~38, no.~7, pp. 793--812, 2019.

\bibitem{chitnis2016guided}
R.~Chitnis, D.~Hadfield-Menell, A.~Gupta, S.~Srivastava, E.~Groshev, C.~Lin,
  and P.~Abbeel, ``Guided search for task and motion plans using learned
  heuristics,'' in \emph{Proc. IEEE Int. Conf. Robot. Autom.}, 2016, pp.
  447--454.

\bibitem{kim2020learning}
B.~Kim and L.~Shimanuki, ``Learning value functions with relational state
  representations for guiding task-and-motion planning,'' in \emph{Proc. Conf.
  Robot. Learn.}\hskip 1em plus 0.5em minus 0.4em\relax PMLR, 2020, pp.
  955--968.

\bibitem{ortiz2021learning}
J.~Ortiz-Haro, V.~N. Hartmann, O.~S. Oguz, and M.~Toussaint, ``Learning
  efficient constraint graph sampling for robotic sequential manipulation,'' in
  \emph{Proc. IEEE Int. Conf. Robot. Autom.}, 2021, pp. 4606--4612.

\bibitem{ortiz2022structured}
J.~Ortiz-Haro, J.-S. Ha, D.~Driess, and M.~Toussaint, ``Structured deep
  generative models for sampling on constraint manifolds in sequential
  manipulation,'' in \emph{Proc. Conf. Robot. Learn.}\hskip 1em plus 0.5em
  minus 0.4em\relax PMLR, 2022, pp. 213--223.

\bibitem{noureddine2017multi}
D.~B. Noureddine, A.~Gharbi, and S.~B. Ahmed, ``Multi-agent deep reinforcement
  learning for task allocation in dynamic environment.'' in \emph{Int. Conf.
  Softw. Technol.}, 2017, pp. 17--26.

\bibitem{ding2021graph}
S.~Ding, D.~Lin, and X.~Zhou, ``Graph convolutional reinforcement learning for
  dependent task allocation in edge computing,'' in \emph{Proc. IEEE Int. Conf.
  Agents}.\hskip 1em plus 0.5em minus 0.4em\relax IEEE, 2021, pp. 25--30.

\bibitem{shyalika2020reinforcement}
C.~Shyalika, T.~Silva, and A.~Karunananda, ``Reinforcement learning in dynamic
  task scheduling: A review,'' \emph{SN Computer Science}, vol.~1, no.~6, p.
  306, 2020.

\bibitem{li2024league++}
Z.~Li, K.~Yu, S.~Cheng, and D.~Xu, ``League++: Empowering continual robot
  learning through guided skill acquisition with large language models,'' in
  \emph{Int. Conf. Learn. Represent. 2024 Workshop Large Lang. Model Agents}.

\bibitem{meng2023signal}
Y.~Meng and C.~Fan, ``Signal temporal logic neural predictive control,''
  \emph{IEEE Robot. Autom. Lett.}, 2023.

\bibitem{pan2022failure}
T.~Pan, A.~M. Wells, R.~Shome, and L.~E. Kavraki, ``Failure is an option: Task
  and motion planning with failing executions,'' in \emph{Proc. IEEE Int. Conf.
  Robot. Autom.}, 2022, pp. 1947--1953.

\bibitem{wang2019learning}
A.~S. Wang and O.~Kroemer, ``Learning robust manipulation strategies with
  multimodal state transition models and recovery heuristics,'' in \emph{Proc.
  IEEE Int. Conf. Robot. Autom.}, 2019, pp. 1309--1315.

\bibitem{luo2021endowing}
S.~Luo, H.~Wu, S.~Duan, Y.~Lin, and J.~Rojas, ``Endowing robots with
  longer-term autonomy by recovering from external disturbances in manipulation
  through grounded anomaly classification and recovery policies,'' \emph{J.
  Intell. Robot. Syst.}, vol. 101, pp. 1--40, 2021.

\bibitem{zhang2023grounding}
X.~Zhang, Y.~Ding, S.~Amiri, H.~Yang, A.~Kaminski, C.~Esselink, and S.~Zhang,
  ``Grounding classical task planners via vision-language models,'' in
  \emph{Proc. ICRA Workshop Robot Exec. Failures Fail. Manag. Strateg.}, 2023.

\bibitem{liu2023reflect}
Z.~Liu, A.~Bahety, and S.~Song, ``Reflect: Summarizing robot experiences for
  failure explanation and correction,'' in \emph{Proc. Conf. Robot.
  Learn.}\hskip 1em plus 0.5em minus 0.4em\relax PMLR, 2023, pp. 3468--3484.

\bibitem{rana2023sayplan}
K.~Rana, J.~Haviland, S.~Garg, J.~Abou-Chakra, I.~Reid, and N.~Suenderhauf,
  ``Sayplan: Grounding large language models using 3d scene graphs for scalable
  robot task planning,'' in \emph{Proc. Conf. Robot. Learn.}\hskip 1em plus
  0.5em minus 0.4em\relax PMLR, 2023, pp. 23--72.

\bibitem{zhang2023survey}
J.~Zhang, C.~Liu, X.~Li, H.-L. Zhen, M.~Yuan, Y.~Li, and J.~Yan, ``A survey for
  solving mixed integer programming via machine learning,''
  \emph{Neurocomputing}, vol. 519, pp. 205--217, 2023.

\bibitem{bengio2021machine}
Y.~Bengio, A.~Lodi, and A.~Prouvost, ``Machine learning for combinatorial
  optimization: a methodological tour d’horizon,'' \emph{Eur. J. Oper. Res.},
  vol. 290, no.~2, pp. 405--421, 2021.

\bibitem{marcos2014supervised}
A.~Marcos~Alvarez, Q.~Louveaux, and L.~Wehenkel, ``A supervised machine
  learning approach to variable branching in branch-and-bound,'' \emph{Tech.
  Rep.}, 2014.

\bibitem{khalil2016learning}
E.~Khalil, P.~Le~Bodic, L.~Song, G.~Nemhauser, and B.~Dilkina, ``Learning to
  branch in mixed integer programming,'' in \emph{Proc. AAAI Conf. Artif.
  Intell.}, vol.~30, no.~1, 2016, pp. 724--731.

\bibitem{gasse2019exact}
M.~Gasse, D.~Ch{\'e}telat, N.~Ferroni, L.~Charlin, and A.~Lodi, ``Exact
  combinatorial optimization with graph convolutional neural networks,''
  \emph{Adv. Neural Inf. Process. Syst.}, vol.~32, 2019.

\bibitem{song2020general}
J.~Song, Y.~Yue, B.~Dilkina \emph{et~al.}, ``A general large neighborhood
  search framework for solving integer linear programs,'' \emph{Adv. Neural
  Inf. Process. Syst.}, vol.~33, pp. 20\,012--20\,023, 2020.

\bibitem{sonnerat2021learning}
N.~Sonnerat, P.~Wang, I.~Ktena, S.~Bartunov, and V.~Nair, ``Learning a large
  neighborhood search algorithm for mixed integer programs,''
  \emph{arXiv:2107.10201}, 2021.

\bibitem{liu2022learning}
D.~Liu, M.~Fischetti, and A.~Lodi, ``Learning to search in local branching,''
  in \emph{Proc. AAAI Conf. Artif. Intell.}, vol.~36, no.~4, 2022, pp.
  3796--3803.

\bibitem{srinivasan2021fast}
M.~Srinivasan, A.~Chakrabarty, R.~Quirynen, N.~Yoshikawa, T.~Mariyama, and
  S.~Di~Cairano, ``Fast multi-robot motion planning via imitation learning of
  mixed-integer programs,'' \emph{Int. Fed. Autom. Control}, vol.~54, no.~20,
  pp. 598--604, 2021.

\bibitem{cauligi2021coco}
A.~Cauligi, P.~Culbertson, E.~Schmerling, M.~Schwager, B.~Stellato, and
  M.~Pavone, ``Coco: Online mixed-integer control via supervised learning,''
  \emph{IEEE Robot. Autom. Lett.}, vol.~7, no.~2, pp. 1447--1454, 2021.

\bibitem{deits2019lvis}
R.~Deits, T.~Koolen, and R.~Tedrake, ``Lvis: Learning from value function
  intervals for contact-aware robot controllers,'' in \emph{Proc. IEEE Int.
  Conf. Robot. Autom.}, 2019, pp. 7762--7768.

\bibitem{ajay2024compositional}
A.~Ajay, S.~Han, Y.~Du, S.~Li, A.~Gupta, T.~Jaakkola, J.~Tenenbaum,
  L.~Kaelbling, A.~Srivastava, and P.~Agrawal, ``Compositional foundation
  models for hierarchical planning,'' \emph{Adv. Neural Inf. Process. Syst.},
  vol.~36, 2024.

\bibitem{qiu2023large}
Y.~Qiu, Z.~Zhao, Y.~Ziser, A.~Korhonen, E.~Ponti, and S.~B. Cohen, ``Are large
  language model temporally grounded?'' in \emph{Proc. Conf. North Amer. Chap.
  Assoc. Comput. Linguist.: Hum. Lang. Technol.}, vol.~1, 2024, pp. 7057--7076.

\bibitem{li2022systematic}
X.~L. Li, A.~Kuncoro, J.~Hoffmann, C.~de~Masson~d’Autume, P.~Blunsom, and
  A.~Nematzadeh, ``A systematic investigation of commonsense knowledge in large
  language models,'' in \emph{Proc. Conf. Empir. Methods Nat. Lang. Process.},
  2022, pp. 11\,838--11\,855.

\bibitem{chen2023say}
J.~Chen, W.~Shi, Z.~Fu, S.~Cheng, L.~Li, and Y.~Xiao, ``Say what you mean!
  large language models speak too positively about negative commonsense
  knowledge,'' in \emph{Proc. Annu. Meet. Assoc. Comput. Linguist.}, vol.~1,
  2023, pp. 9890--9908.

\bibitem{chen2024spatialvlm}
B.~Chen, Z.~Xu, S.~Kirmani, B.~Ichter, D.~Sadigh, L.~Guibas, and F.~Xia,
  ``Spatialvlm: Endowing vision-language models with spatial reasoning
  capabilities,'' in \emph{Proc. IEEE/CVF Conf. Comput. Vis. Pattern
  Recognit.}, June 2024, pp. 14\,455--14\,465.

\bibitem{rocamonde2023vision}
J.~Rocamonde, V.~Montesinos, E.~Nava, E.~Perez, and D.~Lindner,
  ``Vision-language models are zero-shot reward models for reinforcement
  learning,'' in \emph{Proc. Int. Conf. Learn. Represent.}, 2024.

\bibitem{chen2023large}
L.~Chen, B.~Li, S.~Shen, J.~Yang, C.~Li, K.~Keutzer, T.~Darrell, and Z.~Liu,
  ``Large language models are visual reasoning coordinators,'' \emph{Adv.
  Neural Inf. Process. Syst.}, vol.~36, 2024.

\bibitem{liu2023think}
L.~Liu, X.~Yang, Y.~Shen, B.~Hu, Z.~Zhang, J.~Gu, and G.~Zhang,
  ``Think-in-memory: Recalling and post-thinking enable llms with long-term
  memory,'' \emph{arXiv:2311.08719}, 2023.

\bibitem{wang2024augmenting}
W.~Wang, L.~Dong, H.~Cheng, X.~Liu, X.~Yan, J.~Gao, and F.~Wei, ``Augmenting
  language models with long-term memory,'' \emph{Adv. Neural Inf. Process.
  Syst.}, vol.~36, 2024.

\bibitem{kannan2023smart}
S.~S. Kannan, V.~L. Venkatesh, and B.-C. Min, ``Smart-llm: Smart multi-agent
  robot task planning using large language models,'' \emph{arXiv:2309.10062},
  2023.

\bibitem{yang2023set}
J.~Yang, H.~Zhang, F.~Li, X.~Zou, C.~Li, and J.~Gao, ``Set-of-mark prompting
  unleashes extraordinary visual grounding in gpt-4v,''
  \emph{arXiv:2310.11441}, 2023.

\bibitem{chen2023open}
B.~Chen, F.~Xia, B.~Ichter, K.~Rao, K.~Gopalakrishnan, M.~S. Ryoo, A.~Stone,
  and D.~Kappler, ``Open-vocabulary queryable scene representations for real
  world planning,'' in \emph{Proc. IEEE Int. Conf. Robot. Autom.}, 2023, pp.
  11\,509--11\,522.

\bibitem{wang2023gensim}
L.~Wang, Y.~Ling, Z.~Yuan, M.~Shridhar, C.~Bao, Y.~Qin, B.~Wang, H.~Xu, and
  X.~Wang, ``Gensim: Generating robotic simulation tasks via large language
  models,'' in \emph{Proc. Int. Conf. Learn. Represent.}, 2023.

\bibitem{janner2022planning}
M.~Janner, Y.~Du, J.~Tenenbaum, and S.~Levine, ``Planning with diffusion for
  flexible behavior synthesis,'' in \emph{Int. Conf. Mach. Learn.}\hskip 1em
  plus 0.5em minus 0.4em\relax PMLR, 2022, pp. 9902--9915.

\bibitem{ajay2022conditional}
A.~Ajay, Y.~Du, A.~Gupta, J.~Tenenbaum, T.~Jaakkola, and P.~Agrawal, ``Is
  conditional generative modeling all you need for decision-making?''
  \emph{arXiv:2211.15657}, 2022.

\bibitem{ze20243d}
Y.~Ze, G.~Zhang, K.~Zhang, C.~Hu, M.~Wang, and H.~Xu, ``{3D Diffusion Policy:
  Generalizable Visuomotor Policy Learning via Simple 3D Representations},'' in
  \emph{Proc. Robot. Sci. Syst.}, 2024.

\bibitem{pan2024modelbased}
\BIBentryALTinterwordspacing
C.~Pan, Z.~Yi, G.~Shi, and G.~Qu, ``Model-based diffusion for trajectory
  optimization,'' 2024, accessed on June 5, 2024. [Online]. Available:
  \url{https://lecar-lab.github.io/mbd/}
\BIBentrySTDinterwordspacing

\bibitem{mishra2023generative}
U.~A. Mishra, S.~Xue, Y.~Chen, and D.~Xu, ``Generative skill chaining:
  Long-horizon skill planning with diffusion models,'' in \emph{Proc. Conf.
  Robot. Learn.}\hskip 1em plus 0.5em minus 0.4em\relax PMLR, 2023, pp.
  2905--2925.

\bibitem{fang2023dimsam}
X.~Fang, C.~Garrett, C.~Eppner, T.~Lozano-P{\'e}rez, L.~Kaelbling, and D.~Fox,
  ``Dimsam: Diffusion models as samplers for task and motion planning under
  partial observability,'' in \emph{Proc. CoRL 2023 Workshop Learn. Effective
  Abstr. Plan.}, 2023.

\bibitem{lee2020making}
M.~A. Lee, Y.~Zhu, P.~Zachares, M.~Tan, K.~Srinivasan, S.~Savarese, L.~Fei-Fei,
  A.~Garg, and J.~Bohg, ``Making sense of vision and touch: Learning multimodal
  representations for contact-rich tasks,'' \emph{IEEE Trans. Robot.}, vol.~36,
  no.~3, pp. 582--596, 2020.

\bibitem{yu2023mimictouch}
K.~Yu, Y.~Han, M.~Zhu, and Y.~Zhao, ``Mimictouch: Learning human's control
  strategy with multi-modal tactile feedback,'' in \emph{NeurIPS Workshop Touch
  Process.: New Sens. Modality AI}, 2023.

\bibitem{li2023see}
H.~Li, Y.~Zhang, J.~Zhu, S.~Wang, M.~A. Lee, H.~Xu, E.~Adelson, L.~Fei-Fei,
  R.~Gao, and J.~Wu, ``See, hear, and feel: Smart sensory fusion for robotic
  manipulation,'' in \emph{Proc. Conf. Robot. Learn.}\hskip 1em plus 0.5em
  minus 0.4em\relax PMLR, 2023, pp. 1368--1378.

\bibitem{mason2018toward}
M.~T. Mason, ``Toward robotic manipulation,'' \emph{Annu. Rev. Control Robot.
  Auton. Syst.}, vol.~1, pp. 1--28, 2018.

\bibitem{zhang2021efficient}
Z.~Zhang, J.~Yan, X.~Kong, G.~Zhai, and Y.~Liu, ``Efficient motion planning
  based on kinodynamic model for quadruped robots following persons in confined
  spaces,'' \emph{IEEE/ASME Trans. Mechatronics}, vol.~26, no.~4, pp.
  1997--2006, 2021.

\bibitem{mcgreavy2022reachability}
C.~McGreavy and Z.~Li, ``Reachability map for diverse and energy efficient
  stepping of humanoids,'' \emph{IEEE/ASME Trans. Mechatronics}, vol.~27,
  no.~6, pp. 5307--5317, 2022.

\bibitem{sferrazza2024humanoidbench}
C.~Sferrazza, D.-M. Huang, X.~Lin, Y.~Lee, and P.~Abbeel, ``Humanoidbench:
  Simulated humanoid benchmark for whole-body locomotion and manipulation,'' in
  \emph{Proc. Robot. Sci. Syst.}, 2024.

\bibitem{liu2023task}
W.~Liu, X.~Liang, and M.~Zheng, ``Task-constrained motion planning considering
  uncertainty-informed human motion prediction for human--robot collaborative
  disassembly,'' \emph{IEEE/ASME Trans. Mechatronics}, vol.~28, no.~4, pp.
  2056--2063, 2023.

\bibitem{cheng2021human}
Y.~Cheng, L.~Sun, and M.~Tomizuka, ``Human-aware robot task planning based on a
  hierarchical task model,'' \emph{IEEE Robot. Autom. Lett.}, vol.~6, no.~2,
  pp. 1136--1143, 2021.

\bibitem{darvish2020hierarchical}
K.~Darvish, E.~Simetti, F.~Mastrogiovanni, and G.~Casalino, ``A hierarchical
  architecture for human--robot cooperation processes,'' \emph{IEEE Trans.
  Robot.}, vol.~37, no.~2, pp. 567--586, 2020.

\bibitem{faroni2023optimal}
M.~Faroni, A.~Umbrico, M.~Beschi, A.~Orlandini, A.~Cesta, and N.~Pedrocchi,
  ``Optimal task and motion planning and execution for multiagent systems in
  dynamic environments,'' \emph{IEEE Trans. Cybern.}, 2023.

\bibitem{le2021hierarchical}
A.~T. Le, P.~Kratzer, S.~Hagenmayer, M.~Toussaint, and J.~Mainprice,
  ``Hierarchical human-motion prediction and logic-geometric programming for
  minimal interference human-robot tasks,'' in \emph{Proc. IEEE Int. Conf.
  Robot. Human Interact. Commun.}, 2021, pp. 7--14.

\bibitem{zhang2023large}
C.~Zhang, J.~Chen, J.~Li, Y.~Peng, and Z.~Mao, ``Large language models for
  human-robot interaction: A review,'' \emph{Biomimetic Intelligence and
  Robotics}, p. 100131, 2023.

\bibitem{kshirsagar2019specifying}
A.~Kshirsagar, H.~Kress-Gazit, and G.~Hoffman, ``Specifying and synthesizing
  human-robot handovers,'' in \emph{Proc. IEEE/RSJ Int. Conf. Intell. Robots
  Syst.}, 2019, pp. 5930--5936.

\bibitem{mortl2012role}
A.~M{\"o}rtl, M.~Lawitzky, A.~Kucukyilmaz, M.~Sezgin, C.~Basdogan, and
  S.~Hirche, ``The role of roles: Physical cooperation between humans and
  robots,'' \emph{Int. J. Robot. Res.}, vol.~31, no.~13, pp. 1656--1674, 2012.

\bibitem{otto2018optimization}
A.~Otto, N.~Agatz, J.~Campbell, B.~Golden, and E.~Pesch, ``Optimization
  approaches for civil applications of unmanned aerial vehicles (uavs) or
  aerial drones: A survey,'' \emph{Networks}, vol.~72, no.~4, pp. 411--458,
  2018.

\bibitem{conesa2016route}
J.~Conesa-Mu{\~n}oz, J.~M. Bengochea-Guevara, D.~Andujar, and A.~Ribeiro,
  ``Route planning for agricultural tasks: A general approach for fleets of
  autonomous vehicles in site-specific herbicide applications,'' \emph{Comput.
  Electron. Agric.}, vol. 127, pp. 204--220, 2016.

\bibitem{siburianintegrated}
J.~Siburian, C.~C. Beltran-Hernandez, and M.~Hamaya, ``Integrated task and
  motion planning for real-world cooking tasks,'' in \emph{IEEE Int. Conf.
  Robot. Autom. 2024 Workshop Cook. Robot.: Percept. Motion Plann.}

\bibitem{wan2022arranging}
W.~Wan, T.~Kotaka, and K.~Harada, ``Arranging test tubes in racks using
  combined task and motion planning,'' \emph{Robot. Auton. Syst.}, vol. 147, p.
  103918, 2022.

\bibitem{lin2014robot}
P.~Lin, K.~Abney, and G.~A. Bekey, \emph{Robot ethics: the ethical and social
  implications of robotics}.\hskip 1em plus 0.5em minus 0.4em\relax MIT press,
  2014.

\bibitem{wu2022sustainable}
C.-J. Wu, R.~Raghavendra, U.~Gupta, B.~Acun, N.~Ardalani, K.~Maeng, G.~Chang,
  F.~Aga, J.~Huang, C.~Bai \emph{et~al.}, ``Sustainable ai: Environmental
  implications, challenges and opportunities,'' \emph{Proc. Mach. Learn.
  Syst.}, vol.~4, pp. 795--813, 2022.

\bibitem{sinha2017review}
A.~Sinha, P.~Malo, and K.~Deb, ``A review on bilevel optimization: From
  classical to evolutionary approaches and applications,'' \emph{IEEE Trans.
  Evol. Comput.}, vol.~22, no.~2, pp. 276--295, 2017.

\bibitem{bommasani2021opportunities}
R.~Bommasani, D.~A. Hudson, E.~Adeli, R.~Altman, S.~Arora, S.~von Arx, M.~S.
  Bernstein, J.~Bohg, A.~Bosselut, E.~Brunskill \emph{et~al.}, ``On the
  opportunities and risks of foundation models,'' \emph{arXiv:2108.07258},
  2021.

\bibitem{sutton2018reinforcement}
R.~S. Sutton and A.~G. Barto, \emph{Reinforcement learning: An
  introduction}.\hskip 1em plus 0.5em minus 0.4em\relax MIT press, 2018.

\end{thebibliography}
}

\vspace{-0.25in}
\begin{IEEEbiography}[{\includegraphics[width=1in,height=1.25in,clip,keepaspectratio]{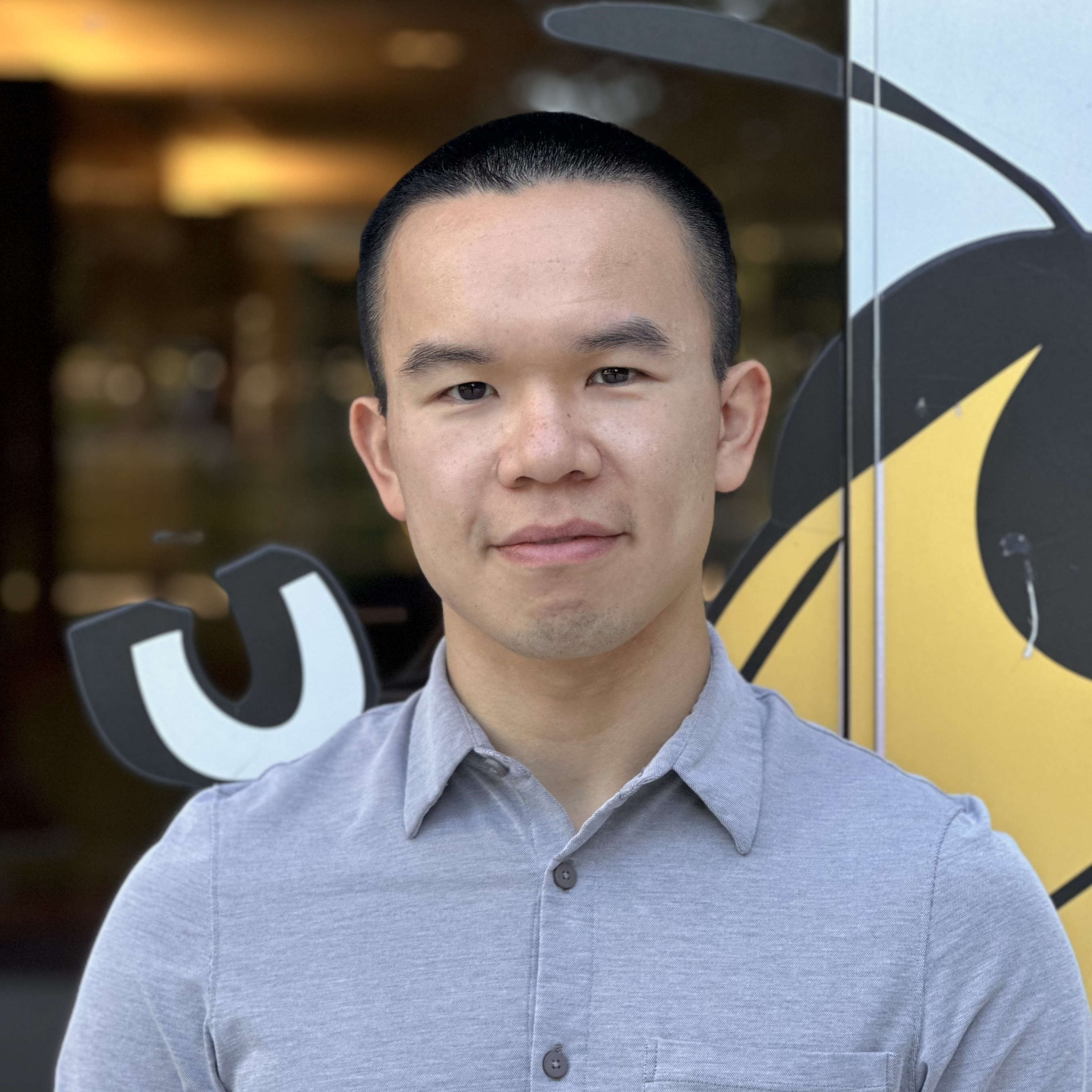}}]{Zhigen Zhao}
received the B.S. degree in Mechanical Engineering from Georgia Institute of Technology, Atlanta, GA, in 2018. He is currently pursuing a Ph.D. degree in Robotics at Georgia Institute of Technology. His research interests lie in task and motion planning, trajectory optimization, and combining model-based and learning approaches for robot planning. His primary focus is on applications involving robot manipulation and loco-manipulation tasks, where he aims to advance the capabilities of robotic systems in complex, dynamic environments.
\end{IEEEbiography}

\vspace{-0.25in}

\begin{IEEEbiography}[{\includegraphics[width=1in,height=1.25in,clip,keepaspectratio]{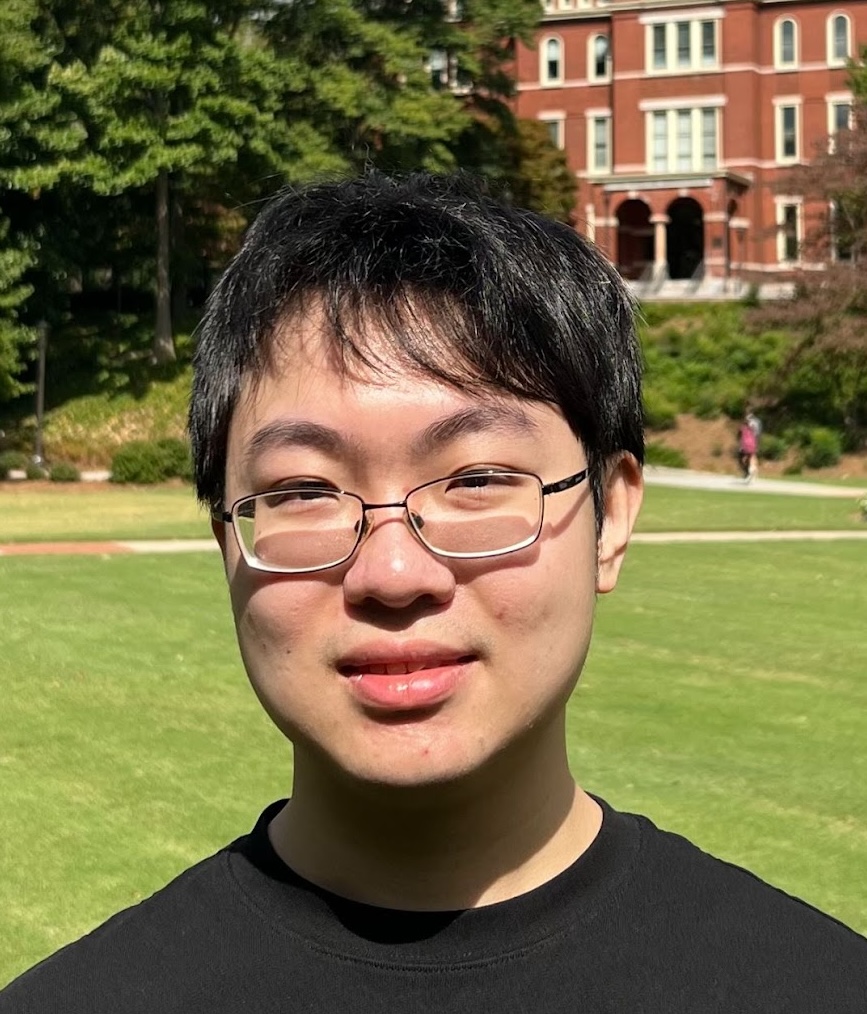}}]{Shuo Cheng}
 is currently working toward a Ph.D. degree in Computer Science at the Georgia Institute of Technology, GA, US. His research focuses on enabling robots with the ability to reason and perform in complex and highly variable environments for achieving long-horizon tasks. To approach this objective, he studies novel representations for efficient, generalizable, and scalable robot skill acquisition and develop systems that compose and reuse the skills robustly, by exploring techniques such as task and motion planning, imitation learning, reinforcement learning, and visual perception learning. His work has been nominated for Best Paper at IEEE RA-L.
\end{IEEEbiography}

\vspace{-0.25in}

\begin{IEEEbiography}[{\includegraphics[width=1in,height=1.25in,clip,keepaspectratio]{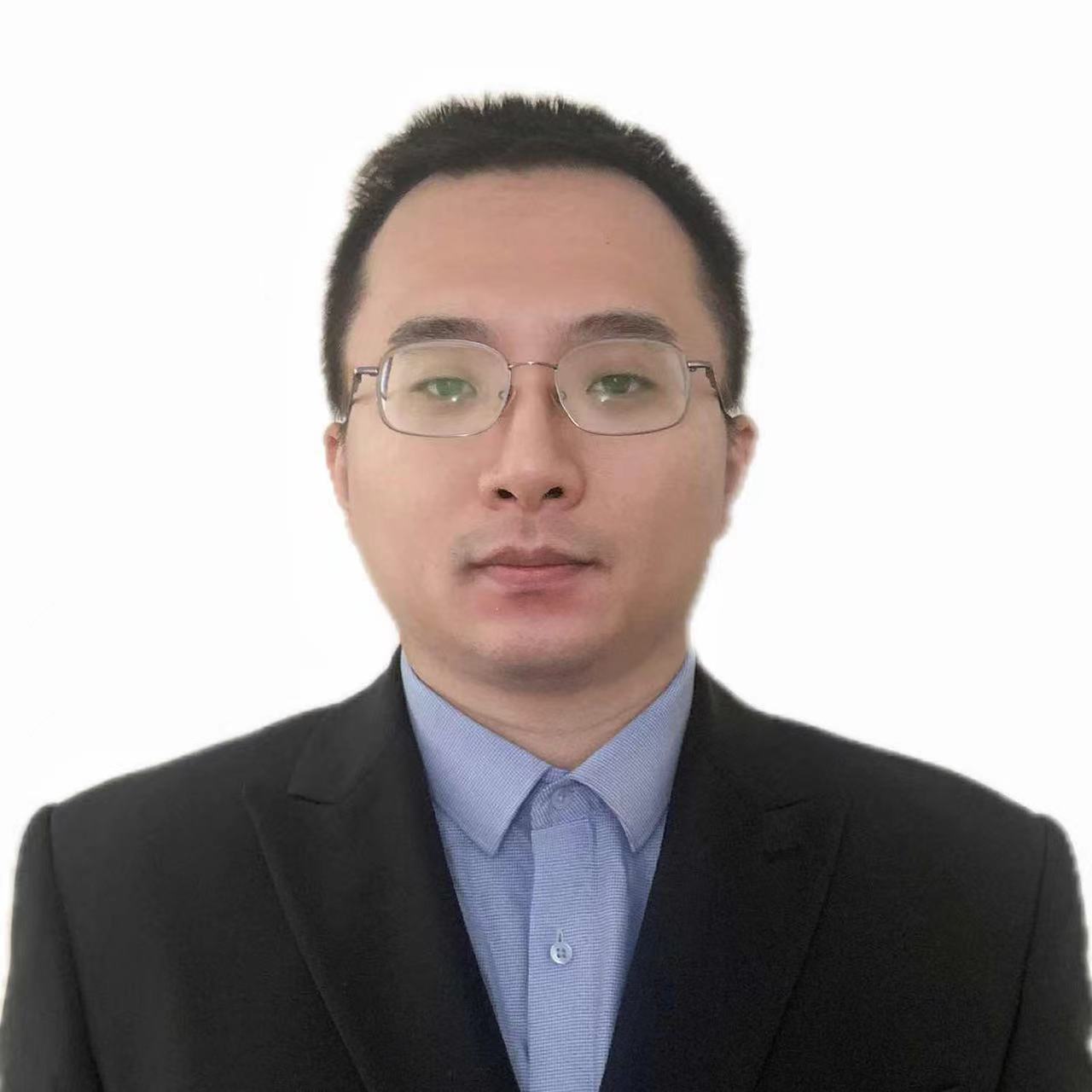}}]{Yan Ding}
is a researcher at the Shanghai AI Laboratory. He earned his Ph.D. in Computer Science from the State University of New York at Binghamton in 2024, following his Bachelor's and Master's degrees from Chongqing University, China, in 2016 and 2019, respectively. His research interests lie in spatial intelligence for robotics, where he aims to empower robots with the capability to understand and interact with the real world, and skill learning for robotics, focusing on enabling robots to effectively transform the real world.
\end{IEEEbiography}

\vspace{-0.25in}

\begin{IEEEbiography}[{\includegraphics[width=1in,height=1.25in,clip,keepaspectratio]{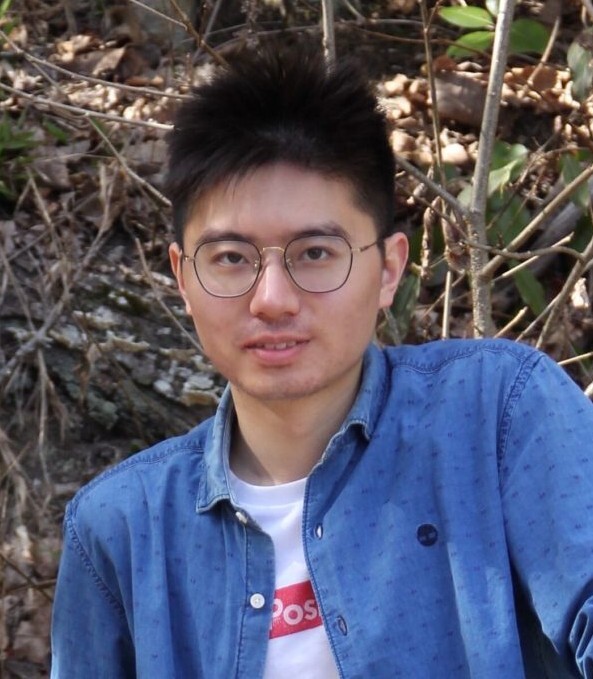}}]{Ziyi Zhou}
received the B.S. degree in Automation from Northeastern University, Shenyang, China in 2018 and the M.S. degree in Electrical and Computer Engineering from Georgia Institute of Technology in 2021. He is currently pursuing a Ph.D. degree in electrical and computer engineering at Georgia Institute of Technology with a focus on robotics. His research interests center around contact-rich trajectory optimization and task planning applied to legged robots.
\end{IEEEbiography}

\vspace{-0.25in}

\begin{IEEEbiography}[{\includegraphics[width=1in,height=1.25in,clip,keepaspectratio]{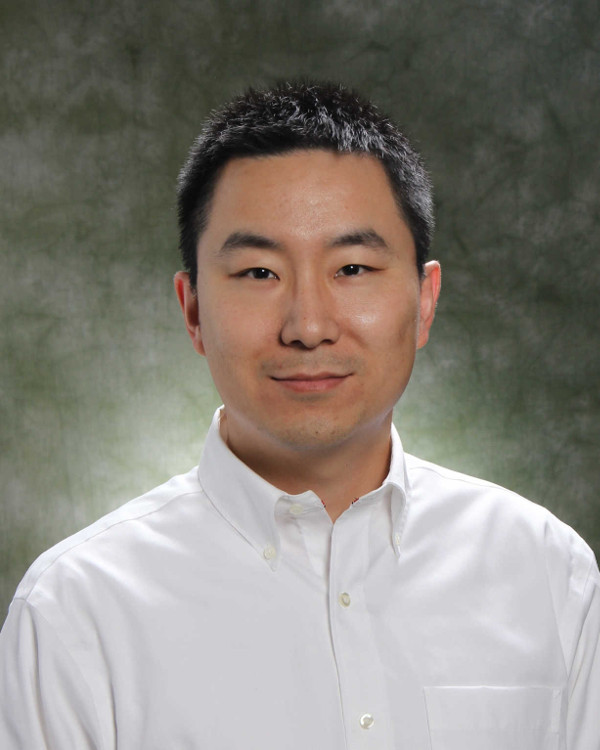}}]{Shiqi Zhang}
is an Associate Professor with the School of Computing, the State University of New York at Binghamton (SUNY Binghamton). He was an Assistant Professor at Cleveland State University and before that was a Postdoctoral Fellow at the University of Texas at Austin. He received his Ph.D. in Computer Science (2013) from Texas Tech University, and received his M.S. and B.S. degrees from Harbin Institute of Technology. He received the AAMAS-2018 Best Robotics Paper Award, Ford URP Award in 2019, OPPO Faculty Research Award in 2020, Top Cited Article recognition from AI Magazine in 2023 and Outstanding Associate Editor recognition from IEEE Robotics and Automation Letters in 2024. He served on the organizing committees of the AAMAS-2022 and KR-2023 conferences.
\end{IEEEbiography}

\vspace{-0.25in}

\begin{IEEEbiography}[{\includegraphics[width=1in,height=1.25in,clip,keepaspectratio]{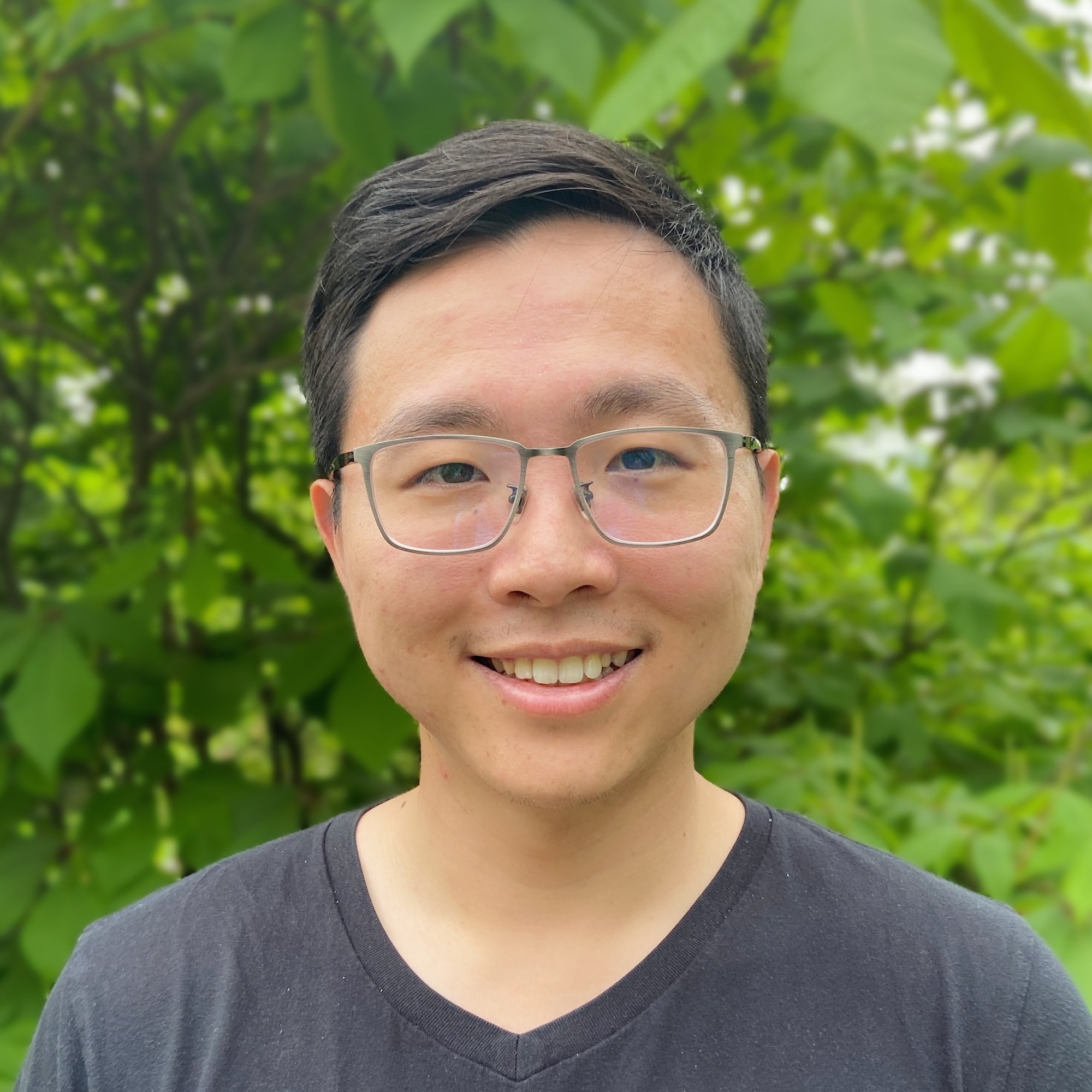}}]{Danfei Xu}
is an Assistant Professor in the School of Interactive Computing at Georgia Institute of Technology. He received his Ph.D. in Computer Science from Stanford University in 2021. His research is in machine learning methods for robotics, with a focus on manipulation planning and imitation learning. His research goal is to enable physical autonomy in everyday human environments with minimum expert intervention. Xu serves as an Associate Editor for International Journal on Robotics Research. He was named as a 2022 DARPA Riser. His work has been nominated for Best Paper at CoRL and IEEE RA-L.
\end{IEEEbiography}

\vspace{-0.25in}

\begin{IEEEbiography}[{\includegraphics[width=1in,height=1.25in,clip,keepaspectratio]{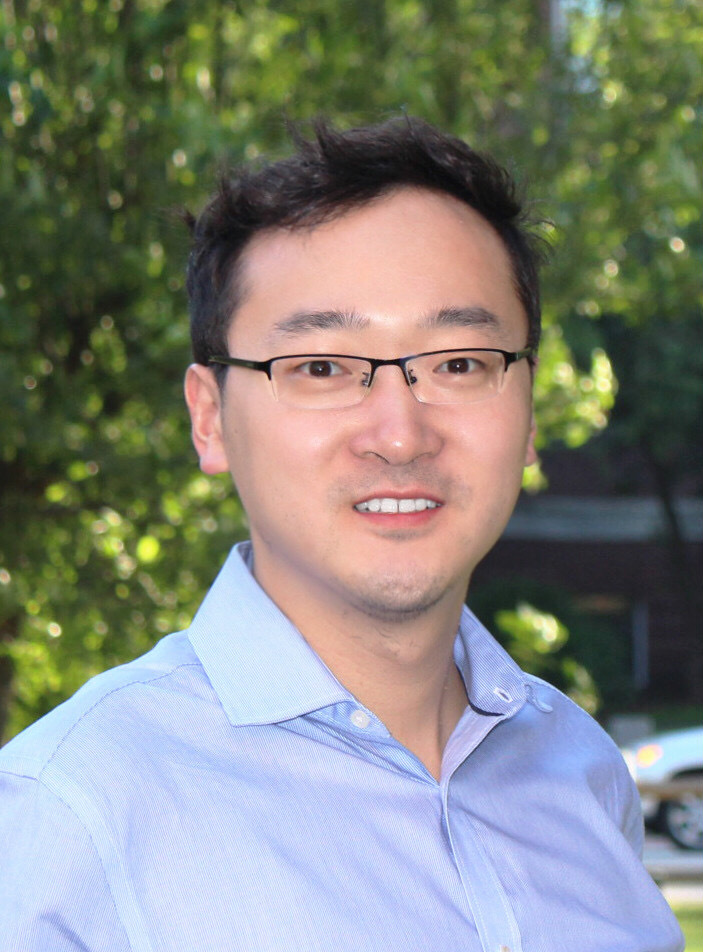}}]{Ye Zhao}
(Senior Member, IEEE) received the Ph.D. degree in Mechanical Engineering from The University of Texas at Austin, Austin, TX, USA, in 2016. He is currently an Assistant Professor with the George W. Woodruff School of Mechanical Engineering, Georgia Institute of Technology, Atlanta, GA, USA. He was a Postdoctoral Fellow with the John A. Paulson School of Engineering and Applied Sciences, Harvard University, Cambridge, MA, USA. His research interests include robust task and motion planning, contact-rich trajectory optimization, formal methods for legged locomotion and navigation. He serves as an Associate Editor of T-RO, TMECH, RA-L, and L-CSS. He received the George W. Woodruff School Faculty Research Award at Georgia Tech in 2023, NSF CAREER Award in 2022, and ONR YIP Award in 2023.
\end{IEEEbiography}

\vfill

\end{document}